\documentclass[]{elsarticle}

\usepackage{amsmath}

\usepackage{amssymb}
\usepackage{cases}

\usepackage{setspace}
\usepackage{breakcites}
\usepackage{graphicx} % more modern
\usepackage{subfig}
\usepackage{algorithm}
\usepackage{algorithmic}

\usepackage{hyperref}

\allowdisplaybreaks[1]

\newcommand{\reals}{\mathbb R}
\newcommand{\nats}{\mathbb N}
\newcommand{\E}{\mathbb E}
\newcommand{\sphere}{\mathbb{S}}
\def\bx{\boldsymbol{x}}
\def\by{\boldsymbol{y}}
\def\bY{\boldsymbol{Y}}
\def\bX{\boldsymbol{X}}
\def\bP{\boldsymbol{P}}
\def\bx{\boldsymbol{x}}
\def\bv{\boldsymbol{v}}
\def\bu{\boldsymbol{u}}
\def\bw{\boldsymbol{w}}

\def\bU{\boldsymbol{U}}

\def\cX{\mathcal{X}}
\def\cM{\mathcal{M}}
\def\cB{\mathcal{B}}

\def\cL{\mathcal{L}}
\def\cU{\mathcal{U}}
\def\wH{\widehat{H}}
\def\bzero{\boldsymbol{0}}
\newcommand{\latop}[2]{\genfrac{}{}{0pt}{}{#1}{#2}}
\def\supp{\mathrm{supp}}

\usepackage{accents}
\newlength{\dhatheight}

\DeclareMathOperator{\dist}{\mathrm{dist}}
\DeclareMathOperator{\argmin}{\mathrm{argmin}}

\DeclareMathOperator{\Sp}{\mathrm{Sp}}

\date{}

\newtheorem{theorem}{Theorem}
\newtheorem{proposition}{Proposition}
\newtheorem{lemma}{Lemma}

\newproof{proof}{Proof}

\begin{document}
%\large

\begin{frontmatter}

\title{Fast, Robust and Non-convex  Subspace Recovery}

\author{G. Lerman}
\author{T. Maunu}
\address{School of Mathematics, University of Minnesota}

\begin{abstract}
    This work presents a fast and non-convex algorithm for robust subspace recovery. The data sets considered include inliers drawn around a low-dimensional subspace of a higher dimensional ambient space, and a possibly large portion of outliers that do not lie nearby this subspace. The proposed algorithm, which we refer to as Fast Median Subspace (FMS), is designed to robustly determine the underlying subspace of such data sets, while having lower computational complexity than existing methods. We prove convergence of the FMS iterates to a stationary point. Further, under a special model of data, FMS converges to a point which is near to the global minimum with overwhelming probability. Under this model, we show that the iteration complexity is globally bounded and locally $r$-linear. The latter theorem holds for any fixed fraction of outliers (less than 1) and any fixed positive distance between the limit point and the global minimum. Numerical experiments on synthetic and real data demonstrate its competitive speed and accuracy.
\end{abstract}

\end{frontmatter}

\section{Introduction}

In the modern age, data is collected in increasingly higher dimensions and massive quantities. An important method for analyzing large, high-dimensional data involves modeling it by a low-dimensional subspace. By projecting the data on this subspace, one can significantly reduce the dimension of the data while capturing its most significant information. Classically, this is the problem of principal component analysis (PCA), which finds the  subspace of maximum variance. PCA is efficiently implemented for moderate-size data by using the singular value decomposition (SVD) of the data matrix. For larger data, recently developed random SVD methods have been proved to be stable, accurate and fast~\cite{HMT_siamrev_pca}.

Despite the impressive progress with effective algorithms for PCA, the underlying idea of PCA is completely useless when the data is corrupted. Among the many possible models for corrupted data sets, here we follow an ``inliers-outliers'' corruption model. More precisely, we assume that some of the data points (the inliers) are sampled around a $d$-dimensional subspace, whereas, the rest of them (the outliers) are sampled from a different (and possibly arbitrary) model. The problem of Robust Subspace Recovery (RSR) asks to robustly estimate the underlying low-dimensional subspace in the presence of outliers. We note that this problem is distinct from what is commonly referred to as ``robust PCA'', that is, recovering the low rank structure in a matrix with sparse element-wise corruptions (see e.g., the work of~\citet{candes_wright_robust_pca09}). Experience has dictated that robust PCA algorithms tend to perform poorly in the RSR regime, especially when the proportion of outliers is high. Much recent work has been devoted for developing numerically efficient solutions of the RSR problem. Current batch RSR formulations are at best comparable to full PCA (which computes all $D$ eigenvectors). That is, their complexity is of order $O(TND^2)$, where $T$ is the number of iterations till convergence, $N$ is the number of points and $D$ is the ambient dimension. We are unaware of sufficiently accurate RSR batch algorithms that scale at least like $O(TNDd)$, where $d$ is the dimension of the approximated subspace.
%(\cite{Li_Haupt_ACOS_2014} has very low complexity and can solve the RSR problem, however, does not achieve the accuracy of other state-of-the-art batch RSR algorithms).
%In particular, we are unaware of using effectively any of the recent randomized methods for PCA in the robust setting.

To address this void, we propose a novel non-convex algorithm for RSR: the Fast Median Subspace (FMS) algorithm. The computational cost of FMS is of order $O(TNDd)$, which not only depends linearly on $D$ (when $d$ is small), but empirically FMS seems to obtain the smallest $T$ and the highest accuracy among all other RSR algorithms (the Tyler M-estimator~\cite{tyler_dist_free87,Teng_log_rpca} has comparable accuracy in many cases, but its computational cost per iteration is significantly larger with moderate or high ambient dimensions). Theoretical guarantees under a model of corrupted data and empirical tests demonstrate the merit of the FMS algorithm.

\subsection{Previous Works}
%{\bf Review of Related Work:}
PCA is by now a classic and ubiquitous method in data analysis~\cite{Jolliffe02PCA}. Since it is obtained by the SVD of the data matrix, it enjoys a wealth of efficient numerical methods.  In the last decade, various random methods have been proposed for fast and accurate computation of the top singular vectors and values (see the review by~\citet{HMT_siamrev_pca}). For example,~\citet{Liberty+2007} demonstrated an order of $O(ND\log(d)+(N+D)d^2)$ randomized algorithm for $d$-approximation PCA; and~\citet{RST_random_pca} have combined random dimension reduction with the power method to obtain a PCA algorithm with $C N D d$ complexity (where $C$ is a small constant) and with significantly improved accuracy when the singular values decay sufficiently fast. The complexity of state-of-the-art  algorithms for online PCA~\cite{Arora+Allerton12,arora2013stochastic} is at best of order $O(TDd)$; however, in practice $T$ is often large and their accuracy is often not competitive. %Despite the impressive progress with effective algorithms for PCA, the underlying idea of PCA is completely useless when the data is corrupted. Here, we only discuss corruption by outliers (as opposed to element-wise corruption~\cite{candes_wright_robust_pca09}). That is, we assume that some of the data points, which we refer to as inliers, are sampled around a $d$-dimensional subspace, whereas, the rest of data points (the outliers) are sampled from a rather different (and possibly arbitrary) model.
%Some of the randomized methods can be performed with sufficient accuracy while streaming over the data with only few passes~\cite{HMT_siamrev_pca}.

While PCA is ubiquitous for subspace modeling without corruption, there is still not yet a clear choice for a best RSR algorithm. Many strategies for RSR have been established in the last three decades (see the review by~\citet{LMTZ2014} and some of the recent developments by~\citet{xu2013outlier, Xu2012,robust_mccoy, robust_pca_ZL, LMTZ2014, Teng_log_rpca, Feng:12, moitra_pca2012}, and~\citet{online_robust_pca13}). Most of the emphasis of the theoretical analysis of RSR algorithms has been on quantifying the largest percentage of outliers under which the studied algorithm can be sufficiently accurate~\cite{Xu2012, LMTZ2014, robust_pca_ZL, Teng_log_rpca, moitra_pca2012}. In particular,~\citet{moitra_pca2012} have shown that guaranteeing the success of an RSR algorithm with a fraction of outliers larger than $(D-d)/D$ for a broad range of instances is as hard as solving the small set expansion problem; they also showed that this fraction can be achieved in their setting; though it is possible to achieve a better fraction in special instances~\cite[p.~766]{robust_pca_ZL}. As opposed to the algorithms of~\citet{LMTZ2014, robust_pca_ZL, Teng_log_rpca}, and~\citet{moitra_pca2012}, other RSR algorithms may not be accurate with high percentage of outliers. Table 1 in~\citep{robust_pca_ZL} summarizes theoretical bounds for the percentage of inliers to outliers required for recovery. All of the algorithms in this table asymptotically depend on $d$ and $D$, where some also depend on the variances of inliers and outliers.

%Despite this impressive progress, there is not yet a sufficiently fast RSR algorithm that can handle a sufficiently large data, as modern implementations of PCA do, and is also sufficiently accurate and robust to outliers.
%
%When the underlying subspace is known, the accuracy of RSR algorithms is often quantified by the error between the computed subspace and the original one.
%The other RSR algorithms listed in~\cite{robust_pca_ZL} do not enjoy this property. For example, the HR-PCA algorithm of~\cite{Xu2010_highdimensional} may be competitive with a small amount of adversary outliers, but can fail with large percentages of outliers. Further,

%There are different models of data corruption~\cite{candes_wright_robust_pca09,Chandrasekaran_Sanghavi_Parrilo_Willsky_2009} and also different models of outliers within RSR algorithms. In general, the right model for outliers is application-dependent and we thus emphasize the numerical experimentation.
%and we thus consider here a real astrophysics application and test our algorithm against all possible competitors.

Many of the successful RSR algorithms involve minimizing an energy, which is robust to outliers. For example,~\citet{Xu2012,xu2013outlier,robust_mccoy, robust_pca_ZL, LMTZ2014}, and~\citet{online_robust_pca13} use convex relaxations of the same energy, which is later formulated in~\eqref{costf2} when $p=1$ and $\delta=0$. We believe that since FMS targets the original robust energy and not a convex relaxation of it, FMS achieves higher accuracy and possibly even faster convergence; however, its analysis is difficult due to the non-convexity. The Tyler M-estimator minimizes a possibly more robust energy and thus obtains competitive accuracy (empirically, our method is as accurate as Tyler's M-estimator). However, it cannot obtain sufficiently competitive speed since it requires full eigenvalue decomposition as well as initial dimensionality reduction by PCA onto a subspace whose dimension is of the order of the number of points. While many of these algorithms for RSR are not sufficiently fast, others are also not very well justified in theory. For example, HR-PCA~\citep{xu2013outlier} and DHR-PCA~\citep{Feng:12} quantify their recovery by the "expressed variance" (EV), but their actual bounds seem to be weak. This is evident in Theorem 2 of~\cite{Feng:12}, which gives asymptotic guarantees. Consider the case of 10\% outliers drawn from a standard Gaussian, and inliers drawn from a standard Gaussian restricted to a subspace. Then it can be shown that their lower bound for EV is 0.09; an EV of 1 amounts to exact recovery.

On the other hand, the procedures of~\citet{moitra_pca2012} do not involve robust energy minimization, but try to fit many different subspaces until success. They are not sufficiently fast and we are unfamiliar with truly competitive implementations of them. Online algorithms~\cite{MKF_workshop09,online_robust_pca13} for RSR suffer from the same problems of online PCA algorithms mentioned above. Namely, the number of iterations required can be quite large, and their accuracy is often not competitive.

%While some algorithms for RSR are not sufficiently accurate or fast, others are also not well motivated by theory. For example, the HR-PCA and DHR-PCA algorithms of~\citet{Xu2010_highdimensional} and~\cite{Feng:12} are designed to emphasize a small amount of adversary outliers. While the performances of DHR-PCA and HR-PCA were not competitive in our tests, we also found their theory to be quite weak. Indeed, they quantify their recovery by the "expressed variance" (EV), but their actual bounds do not hold much value. For example in Theorem 2 of~\cite{Feng:12}, which gives asymptotic guarantees, consider the case of 10\% outliers drawn from a standard Gaussian, and inliers drawn from a standard Gaussian restricted to a subspace. Then it can be shown that their lower bound for EV is 0.09, which is a very weak guarantee since an EV of 1 amounts to exact recovery.

An important algorithm to compare with is spherical PCA (SPCA). SPCA involves performing PCA on the data after it is centered and then projected to the unit sphere.~\citet{robust_stat_book2006} determined that SPCA was their method of choice when compared with various RSR algorithms~\cite{Maronna2005}. Further, the complexity of running SPCA on a data set is $O(NDd)$, which is faster than FMS by a multiplicative constant. Our tests indicate that while SPCA is faster, it does not achieve the competitive accuracy of FMS on subspace recovery problems in the numerical tests of~\S\ref{sec:numericalexp}. Similarly to SPCA, many energy-minimization based algorithms (in particular,~\cite{Xu2012,robust_mccoy, robust_pca_ZL, LMTZ2014, online_robust_pca13}) benefit from initial data normalization to the unit sphere (after robust centering). Indeed, while their underlying energies are robust to high percentages of some outliers, they may be sensitive to adversary outliers of very large magnitude.

It is also worth noting a couple recent works which scale to larger data than previous RSR algorithms. The work on Adaptive Compressive Outlier Sampling by~\citet{Li_Haupt_ACOS_2014}, can be viewed as a solution to the RSR problem with drastically reduced complexity that depends on how many rows and columns of the data are selected. However, it is not as effective at precisely identifying the underlying subspace as our method, which stems from the fact that it builds on Outlier Pursuit (OP)~\cite{Xu2012} (i.e. it is an approximate version of OP, which is not accurate enough). OP could not compete with the accuracy of other RSR algorithms in many of the regimes we test. Another recent algorithm with the potential to scale as well as FMS for RSR is the work on Grassmann Averages~\cite{hauberg2016grassmannave}, provided that the correct robust function $\mu_{rob}$ is chosen. However, Grassmann Averages lack any sort of theoretical justification, both for convergence and robustness.

A similar algorithm to FMS is explored in~\citet{WSL13}, which proposed a non-convex robust PCA algorithm. Although their algorithm is not suited for the RSR problem, it was still relevant for our work on FMS. First, we borrowed from~\citet{WSL13} an argument for the proof of convergence of the FMS iterates to a stationary point (it is one of several different arguments used in our proof). Second of all, the FMS algorithm might be viewed as a soft analog of the alternating least squares (ALS) procedure of~\cite{WSL13} (FMS divides by a power of the distance to a subspace and ALS divides by 1 or ``infinity''; FMS applies randomized SVD, whereas ALS applies alternating low-rank approximation).

Finally, there are many recent works on the analysis of non-convex algorithms and their surprising effectiveness on problems with structured data. Some examples include works by~\citet{SunQuWright_nonconvex_sphere_2015,sun2015nonconvex,zhang2015global,jain2014iterative,dauphin2014identifying,matrix_complete_descent_montanari09,keshavan2009matrix,boumal2016nonconvex}, and~\citet{bandeira2016low}. In particular, there has been related work on non-convex analysis related to low rank modeling (see the work of~\citet{matrix_complete_descent_montanari09,keshavan2009matrix,jain2013low,hardt2014understanding,netrapalli2014non,jain2014fast,jain2014iterative}, and~\citet{zhang2015global} for some examples). Our analysis of FMS presents yet another example where a non-convex algorithm is surprisingly accurate in low rank modeling despite potential issues of non-convex optimization, such as slow convergence or convergence to a non-optimal point. We also point the reader to the work of~\citet{Daubechies_iterativelyreweighted}, which aims at analyzing the convergence of an IRLS method when the energy is non-convex. Although their method is for a different problem, there is a strong similarity in the use of non-convex energies: in their case when $\tau < 1$ and in our case when $p<1$.

\subsection{This Work}

The FMS algorithm improves on existing methods due to its fast runtime and state-of-the-art accuracy. However, the underlying minimization of FMS is non-convex and thus difficult to analyze. This work contributes to non-convex analysis and direct optimization on the Grassmannian manifold $G(D,d)$ in the following ways:
\begin{enumerate}
    \item We prove convergence of the FMS iterates to a stationary point over $G(D,d)$.
    \item For two special models of data, we prove:
    \begin{itemize}
        \item This stationary point is sufficiently close to the global minimum with overwhelming probability;
        \item The convergence rate is globally bounded and locally $r$-linear.
    \end{itemize}
    These two models are:
    \begin{enumerate}
        \item Inliers are drawn from a spherically symmetric distribution restricted to a fixed subspace $L_1^*$ and outliers are drawn from a spherically symmetric distribution in the whole space;
        \item The subspace dimension is $d=1$ and outliers are either symmetrically distributed in the ambient space or lie on another subspace $L_2^*$ (where it is less probable to draw points from $L_2^*$ than $L_1^*$).
    \end{enumerate}
    \item For both models in 2, we guarantee approximate recovery for any percentage of outliers (less than 1); the theory of other RSR algorithms requires bounds on this percentage.
    \item Out of all other RSR algorithms, we provide the only guarantees for the model 2b.
\end{enumerate}

   %While minimizers of convex relaxations of the same energy enjoy richer mathematical theory for both robustness and convergence~\cite{Xu2012, robust_mccoy, robust_pca_ZL, LMTZ2014}, they may falter in some practical cases not satisfying their theoretical model.
In addition to the theory, we rely on careful numerical experimentation and believe that the results reported in this paper strongly indicate the merit of FMS. The FMS algorithm displays competitive speed and accuracy on synthetic data sets. Unlike other RSR algorithms, FMS also shows strong performance as a dimension reduction tool for clustering data since it scales well, as we demonstrate on the human activity recognition data in~\S\ref{exp:clust}.

\subsection{Structure of The Paper}
%{\bf Structure of The Paper:}
This paper begins by motivating and outlining our new algorithm in~\S\ref{rob_pca_sec}. Next,~\S\ref{sec:theory} establishes convergence of the FMS iterates to a stationary point, and further gives optimality and rate guarantees for FMS under a certain model of data. Experiments on synthetic and real data (of astrophysics, human activity, and face recognition) are done in~\S\ref{sec:numericalexp} to demonstrate the usefulness of our new approach. Lastly,~\S\ref{sec:conclusions} concludes this work.

\section{The FMS Algorithm}\label{rob_pca_sec}

This section presents the FMS algorithm. First,~\S\ref{rob_pca_sec_not} presents basic notation used throughout the paper. Then, in~\S\ref{rob_pca_sec_min} we describe its underlying minimization problem and its robustness. Next, in~\S\ref{rob_pca_sec_solve} we propose the FMS algorithm, while motivating it in a heuristic way. Finally,~\S\ref{rob_pca_sec_complexity} summarizes its complexity, and~\S\ref{sec:choiceparam} discusses the choices of parameters for the FMS algorithm.

\subsection{Notation}\label{rob_pca_sec_not}
%{\bf Notation:}
We assume a data set of $N$ points in $\reals^D$, $\cX =\{\boldsymbol{x}_i\}_{i=1}^N$. We seek an approximating $d$-dimensional subspace, $L$, or in short a $d$-subspace.
We denote by $G(D,d)$ the Grassmannian manifold of linear $d$-subspaces of $\reals^D$. For $L \in G(D,d)$ we denote by $\boldsymbol{P}_L$ the orthogonal projector onto $L$, which we view as an element of $\mathbb{R}^{D \times D}$. Let $\| \cdot \|$ denote the Euclidean norm on $\reals^D$.
For $\bx \in \reals^D$ and $L \in G(D,d)$ we denote by $\dist(\bx,L)$ the Euclidean distance of $\bx$ to $L$, that is, $\|\bx-\bP_L \bx\|$. For the distance between $L_1$, $L_2 \in G(D,d)$, which we denote by $\dist(L_1,L_2)$, we use here the square-root of the sum of the squared principal angles between $L_1$ and $L_2$.

\subsection{The Underlying Minimization Problem}
\label{rob_pca_sec_min}

Many approaches for RSR are motivated by the following minimization problem:
For the data set $\cX \subset \reals^D$, $0 < p < 2$ and $\delta>0$, find a $d$-subspace $L$ that minimizes among all such subspaces the energy
%\begin{align}
\begin{equation}
\begin{aligned}\label{costf2}
&F_{p,\delta}(L;\cX) = \!\!\!\!\!\!\!\!\!\!\!\!\sum_{\latop{1 \leq i \leq N}  {\dist^{2-p}(\bx_i,L)\geq p\delta}}
\!\!\!\!\!\!
\dist^p(\bx_i,L) + \!\!\!\!\!\!\!\!\!\!
\!\!\sum_{\latop{1 \leq i \leq N}{\dist^{2-p}(\bx_i,L)< p\delta}}
\!\!\!\!\!\!\!\!\!\! \left(\frac{\dist^2(\bx_i,L)}{2\delta} + (p\delta)^{p/(2-p)} - \frac{(p\delta)^{2/(2-p)}}{2\delta}\right).
\end{aligned}
\end{equation}
%\end{align}
Further, taking $\lim_{p \to 2} F_{p,\delta}$ results in the PCA energy. Thus, we let
\begin{equation}
\begin{aligned}\label{costfPCA}
F_{2,\delta}(L;\cX) = \sum_{1 \leq i \leq N}  \dist^2(\bx_i,L) .
\end{aligned}
\end{equation}
Setting $p=1$ in this energy results in a natural robust extension of PCA, since the solution to the minimization can be thought of as a geometric median subspace. Even approximate minimization of this energy is nontrivial, since it has been shown to be NP hard for $1 \leq p < 2$~\cite{Clarkson_Woodruff_RSA2015} (and assumed to be even harder for $0<p<1$). This minimization was suggested with $p=1$ and $\delta=0$ by
%\cite{osborne_watson85, spath_watson1987, Nyquist_l1_88, Bargiela_Hartley93},
\citet{osborne_watson85, spath_watson1987}, and~\citet{Nyquist_l1_88},
who also proposed algorithmic solutions when $d=D-1$. Later heuristic solutions were proposed for any $d<D$ by~\citet{Ding+06} and~\citet{MKF_workshop09}.

While the domain of this minimization is the set of all affine $d$-subspaces,
experience shows that initial robust centering by the geometric median and then minimization over
$G(D,d)$ is successful. We thus assume in our discussion that the data is centered (if not, we center it at the geometric median) and the domain of the minimization is $G(D,d)$. It is important to be aware that in the presence of a single outlier with sufficiently large magnitude, the minimizer of~\eqref{costf2} (or any of its convex relaxations) may fail to approximate the underlying subspace. Such a case (and its many variants) can be avoided by normalizing the centered data points according to their Euclidean norms so that they lie on the unit sphere (see more discussion in~\cite{robust_pca_ZL, LMTZ2014}).

On the surface, this is a rather natural robust minimization problem, which formally generalizes the notion of the geometric median to subspaces (see discussion in~\cite{robust_pca_ZL}).
However, this minimization is non-convex (since its domain, the Grassmannian, is non-convex). As was mentioned, convex relaxations of it when $p=1$ have been studied~\cite{Xu2012, robust_mccoy, robust_pca_ZL, LMTZ2014}. Nevertheless, for some real data their solutions are not satisfying (see~\S\ref{exp:realdata}).
Furthermore, it is possible that direct approaches to the non-convex minimization, especially with $p<1$, can yield even more robust solutions.
Finally, we are unaware of sufficiently fast implementations for such convex relaxations.
We thus suggest here to revisit the original minimization problem while aiming to obtain faster and more accurate algorithms for practical data.

Robustness of the abstract global minimizer of~\eqref{costf2} when $\delta=0$  was analyzed in~\cite{lp_recovery_part1_11, lp_recovery_part2_11} under special assumptions on the data.
In particular,~\citet{lp_recovery_part2_11} showed that for spherically symmetric outliers and spherically symmetric inliers within a $d$-subspace  (possibly with additional outliers within less significant $d$-subspaces) asymptotic exact recovery is possible even when the fraction of outliers approaches $100\%$. Similarly, near recovery is possible with small amount of noise.
Furthermore, in the noiseless case with outliers, the theory of~\citet{robust_pca_ZL} and~\citet{LMTZ2014} imply that  its theory directly extends to the abstract global minimizer of~\eqref{costf2} when $p=1$ and $\delta=0$ (see Remarks of~\S2.3 in~\cite{LMTZ2014} and Theorem 1 of~\cite{robust_pca_ZL}).

%However, the very special and uncommon case of many normalized outliers with significantly high correlation (i.e., small vector angles) is similar to the case of one arbitrarily large outlier that may fail the recovery by the minimizer of~\eqref{costf2}.

\subsection{Proposed Solution to the Non-convex Minimization}
\label{rob_pca_sec_solve}

We heuristically develop the FMS algorithm that iteratively computes subspaces $(L_k)_{k \in \nats} \subset G(D,d)$; a more rigorous treatment of the resulting sequence follows from the proof of Theorem~\ref{thm:conv} (presented later in~\S\ref{sec:thm:conv}).  Assume first that $\delta=0$. Since $\dist^p(\boldsymbol{x}_i,L) = \dist(\boldsymbol{x}_i,L)^2/\dist(\boldsymbol{x}_i,L)^{2-p}$, instead of minimizing \eqref{costf2}, we may try to minimize at iteration $k+1$ the function
\begin{equation}
\label{costf3} \sum\limits_{i=1}^{N} \dist(\boldsymbol{x}_i,L)^2/\dist(\boldsymbol{x}_i,L_k)^{2-p}.
\end{equation}
The minimizer of \eqref{costf3} is easily obtained by weighted PCA, and thus the whole procedure can be viewed as IRLS (iteratively re-weighted least squares).
However, since the weight $1/\dist(\boldsymbol{x}_i,L_k)^{2-p}$ may be undefined,
we assume that $\delta>0$ and modify the weight to be $1/\max(\dist(\boldsymbol{x}_i,L_k)^{2-p},p\delta)$ (an explanation for this regularized term follows from~\eqref{eq:argminH} which appears later in the proof of Theorem~\ref{thm:conv}).
To solve the weighted PCA problem, one first needs to weight the centered data points by
the latter term and then apply PCA (without centering) to compute $L_{k+1}$. The ability to directly apply
PCA, or equivalently SVD, to the scaled data matrix is numerically attractive, and we can apply any of the state-of-the-art suites for it.

Our procedure at iteration $k$ is outlined as follows. First, form the new weighted data points
\begin{equation}
\by_i = \bx_i/\max(\dist(\boldsymbol{x}_i,L_k)^{(2-p)/2},\sqrt{p\delta}), \  \  1 \leq i \leq N.
\end{equation}
Then, compute top $d$ right singular vectors of the data matrix $\boldsymbol{Y}$, whose columns are the weighted data points $\{\by_i\}_{i=1}^{N}$ (for SVD, we found the randomized method of~\citet{RST_random_pca} to be sufficiently fast without sacrificing accuracy). The subspace $L_{k+1}$ is then the span of these vectors. This procedure is iterated until $L_k$ sufficiently converges. We formally call this iterative procedure the Fast Median Subspace (FMS) and summarize it in Algorithm~\ref{algorithm:FMS} (for simplicity we use the notation $\epsilon$
for $\sqrt{p\delta}$ used above).
%\vspace{-.2cm}
\linespread{1}
\begin{algorithm} [H]
	\caption{Fast Median Subspace (FMS$_p$)}\label{algorithm:FMS}
	\begin{algorithmic}[1]
		\STATE {\bfseries Input:} $\boldsymbol{X} = [\bx_1,\dots,\bx_N]$: $D \times N$ centered data matrix, $d$: desired rank, $p$: robustness power ($0<p<2$; default: $p=1$), $n_m$: maximum number of iterations,  $\tau$, $\epsilon$: parameters (default: $10^{-10}$ for both)
		\STATE {\bfseries Output:}  $L$: $d$-subspace in $\mathbb{R}^D$
		\STATE $k \gets 1$
		\STATE $L_0 \equiv L_1 \gets$ PCA $d$-subspace in $\mathbb{R}^D$
		\WHILE{$k<n_m$ and $\dist(L_k,L_{k-1})>\tau$}
		\FOR{i=1:N}
		\STATE$\boldsymbol{y}_i \gets {\bx_i}/{\max(\dist(\bx_i,L_k)^{(2-p)/2},\epsilon)}$
		\ENDFOR
		\STATE $\bY \gets [\by_1,\dots,\by_N]$
		\STATE $[\boldsymbol{U},\boldsymbol{S},\boldsymbol{V}] \gets$ RandomizedPCA($\bY$,$d$)~\cite{RST_random_pca};
		\STATE $L_{k+1} \gets $ column space of $\bU$
		\STATE $k \gets k+1$
		\ENDWHILE
	\end{algorithmic}
\end{algorithm}
\linespread{1.5}
\vspace{-.2cm}

%The FMS algorithm is an example of can be viewed as IRLS (iteratively re-weighted least squares). Indeed, \eqref{costf4} is a minimization of the squares of
%$\dist(\boldsymbol{x}_i,L)$ with weighted coefficients (based on $\dist(\boldsymbol{x}_i,L_k)$). However, the minimization over the Grassmannian makes it
%a rather unconventional IRLS problem (as shown in our supplementary analysis). Furthermore, the way FMS is formulated in Algorithm~\ref{algorithm:FMS}
%hardly leaves any trace of the IRLS procedure.~\cite{LMTZ2014,robust_pca_ZL} propose IRLS procedures for the minimizer of a relaxed version of~\eqref{costf}
%when $p=1$, however, their computation is more complicated.~\cite{robust_pca_ZL} require matrix inversion at each iteration, which is also not stable, whereas
%\cite{LMTZ2014} require full SVD at each iteration.

In practice, we have found that the rate of convergence of the FMS algorithm is not affected by its particular use of RandomizedPCA~\cite{RST_random_pca}. In other words, if RandomizedPCA is replaced with the exact SVD, the convergence and accuracy are the same, but RandomizedPCA results in shorter runtime. Also, it seems advantageous to initialize $L_0$ with the result of RandomizedPCA (or SVD) on the full data set $\cX$, although initialization can also be done randomly. Random initialization is not recommended, though. For example, in a regime where outliers lie on another weaker subspace, starting too close to the outlier subspace can result in convergence to a local minimum rather than the global minimum. Empirically, with PCA initialization in these cases, FMS converges to the stronger subspace (i.e. it converges to the global minimum). Finally, we denote the FMS algorithm run with a parameter $p$ by FMS$_p$ for the remainder of the paper.

\subsection{Complexity}
\label{rob_pca_sec_complexity}

%The FMS algorithm is advantageous to other RSR batch methods due to its decreased complexity.
At each iteration, FMS$_p$ creates a scaled data matrix of centered data points, which takes $O(DNd)$ operations, although the scaling can be done in parallel. It then finds the top $d$ singular vectors of the scaled data matrix to update the subspace, which takes $O(DN d)$. Thus, the total complexity is $O(TDNd)$, where $T$ is the number of iterations. Empirically, we have noticed that $T$ can be treated as a relatively small constant. For example, in the special case of Theorem~\ref{thm:sublinconv}, $T \leq O\left(1/(p\delta \min(1,\eta^{3(p-1)}))\right)$ for an $\eta$-approximation to the limiting stationary point. This further reduces to $T = O(\log(1/\eta))$ if the iterates are sufficiently close to the limiting stationary point by Theorem~\ref{thm:linconv}. The storage of the FMS$_p$ algorithm involves the $N \times D$ data matrix $\bX$ and the weighted data matrix $\boldsymbol{Y}$ at each iteration. FMS$_p$ must also store the $D \times d$ bases for the subspaces $L_k$ and $L_{k+1}$.
Thus, the storage requirement for FMS$_p$ is $2D N + 2Dd$.

\subsection{Choice of Parameters $p$, $\delta$, and $d$}
\label{sec:choiceparam}

In the later experimental sections (see \S\ref{sec:numericalexp}), we compare FMS$_p$ run with $p=1$ and $p=0.1$. Although there is not always a difference, in some cases we see one of the two choices of $p$ performing better than the other. There also does not appear to be an advantage for using $1<p<2$. Currently, the theory seems to support $p=1$ for the best rates of recovery (see Theorem~\ref{thm:globconv}). Further, the theory seems to also indicate that smaller $p$ leads to less robustness to higher levels of noise (see Theorem~\ref{thm:globconv:stab}). Later experiments in \S\ref{sec:numericalexp} indicate that with small numbers of points, a choice of small $p<1$ can lead to convergence to a non-optimal point, while $p=1$ is still able to converge (see Figures~\ref{fig:phasetrans_fmsp1_1e5noise} and~\ref{fig:phasetrans_fmsp5_1e5noise}).

We believe that $p$ can be optimized for a specific data set, given some prior knowledge of it. In other words, $p$ can be chosen if the user designates a training data set where the truth is known. Due to the low complexity of the method, it is possible to efficiently run it over an array of values of $p$. Thus, with the specified training set, cross-validation can be used to select the proper value of $p$ for a given type of data, although this requires the user to have some prior knowledge of the data.

For choice of $\delta$, we have not noticed too much difference between different values, although there may be certain cases where it is necessary to be careful with the choice of $\delta$. In Theorem~\ref{thm:globconv}, we see rates of asymptotic recovery for the FMS$_p$ algorithm under a special model of data. This theory seems to point to taking $\delta$ as small as possible when $1 \leq p < 2$, but to take larger values of $\delta$ when $0<p<1$. Further, some experiments with smaller sample sizes indicate that using a parameter $\delta$ too small with $p<1$ can lead to convergence to a non-optimal point (i.e. a local minimum). Thus, while we advocate choosing $\delta$ as small as possible, some care must be taken when $p<1$ to ensure that $\delta$ is not chosen to be too small.

Finally, one may ask how to select the subspace dimension $d$ for FMS$_p$. Picking the correct subspace dimension $d$ is not well studied or justified in the literature. Heuristic strategies, such as the elbow method, can be used to guess what the best subspace dimension is; such strategies usually require a test over a range of possible values for $d$. On the other hand, in some domains there is prior knowledge of $d$. For example, in facial recognition type datasets, images of a person's face with constant pose under differing illumination conditions approximately lie on a $d=9$ dimensional subspace~\cite{Basri03}. In practice, we advocate either using the elbow method or domain knowledge to select the best value for $d$.

\section{Theoretical Justification}
\label{sec:theory}
Since FMS$_p$ proposes an iterative process it is rather important to analyze its convergence. In~\S\ref{sec:thm:conv}, we formulate the main convergence theorem, which establishes convergence of FMS$_p$ to a stationary point (or continuum of stationary points). Next,~\S\ref{sec:globconv} assumes the data is sampled from a certain distribution and proves that FMS$_p$ converges to a point near to the global minimum with overwhelming probability. In~\S\ref{sec:convrate}, it is further shown under this model that the rate of convergence is globally bounded and locally $r$-linear (again with overwhelming probability). The proofs of all theorems are left to~\S\ref{sec:proof}.

\subsection{General Convergence Theorem}\label{sec:thm:conv}

We establish convergence to a stationary point of the energy $F_{p,\delta}$ over $G(D,d)$:
\begin{theorem}[Convergence to a Stationary Point]\label{thm:conv}
	Let $(L_k)_{k \in \nats}$, be the sequence obtained by applying FMS$_p$ without stopping for the data set $\cX$ for a fixed $0 < p < 2$. Then, $(L_k)_{k \in \nats}$ converges to a stationary point $L^*$ of $F_{p,\delta}$ over $G(D,d)$, or the accumulation points of $(L_k)_{k \in \nats}$ form a continuum of stationary points.
\end{theorem}

The proof the theorem appears in~\S\ref{sec:conv:proof}. While there are no assumptions on $\cX$, it is important to discuss the implications of this Theorem. In~\S\ref{sec:continuum} we discuss the possibility that FMS$_p$ converges to a continuum of stationary points. Then,~\S\ref{sec:saddle} discusses the possibility of convergence to a saddle point, and~\S\ref{sec:multlocmin} discusses convergence to a local minimum.

\subsubsection{Convergence to a Continuum of Stationary Points}\label{sec:continuum}

Theorem~\ref{thm:conv} proves convergence of the FMS$_p$ iterates to a stationary point or a continuum of stationary points. Another way to think of this issue is that the continuum of stationary points is also a continuum of fixed points for the FMS$_p$ algorithm. It is desirable to know when the algorithm converges to a single point versus a continuum. However, while we cannot see how to rule out the continuum case, we also cannot construct an example of a discrete data set with a continuum of stationary points when the rank of the data set is less than the subspace dimension $d$. When the rank of the data is less than $d$, all subspaces containing the data set are essentially equivalent with respect to the data. We conjecture that we have actual convergence to a single stationary point of $G(D,d)$ for data sets which are full rank.

\subsubsection{Can $(L_k)_{k\in \nats}$ Converge to a Saddle Point?}\label{sec:saddle}

While we prove convergence to a stationary point of $F_{p,\delta}$ over $G(D,d)$, we are not able to say what kind of a stationary point we converge to. In theory we cannot rule out a saddle point, but we are unaware of an example of a saddle point which is also a fixed point of FMS$_p$. The following example describes a saddle point of $F_{p,\delta}$ which is not a fixed point of FMS$_p$ with probability 1. For this example, we assume that if the solution to PCA is not unique, then the PCA output is selected uniformly at random from the solution set.
Consider the data set in $\reals^3$
\begin{equation}
\cX = \left\{(1,0,0)^T, (0,1,0)^T\right\}.
\end{equation}
For the FMS$_p$ energy function $F_{p,\delta}$, the line defined by $\ell_{\text{sad}} = \Sp([1,1,0]^T)$ is a saddle point for the FMS$_p$ energy, since it is a minimum along the geodesic from $ \ell_{\max} = \Sp([0,0,1]^T)$ to $\ell_{\text{sad}}$, but a maximum along the geodesic from $\ell_{\min_1} = \Sp ([1,0,0]^T)$ to $\ell_{\min_2} =  \Sp([0,1,0]^T)$. However, $\ell_\text{sad}$ is not a fixed point of FMS$_p$. Suppose that $\ell_\text{sad}$ is selected as a candidate subspace by FMS$_p$. Then, the two data points $\bx$ and $\bx_2$ are equidistant from $\ell_\text{sad}$ and are scaled by the same amount. Then, when PCA is done to find the new subspace from the scaled data, all lines along the geodesic from $\ell_{\min_1}$ to $\ell_{\min_2}$ are solutions, and thus $\ell_\text{sad}$ is selected again with probability 0. After a new line is selected, the points will no longer be equidistant and one of the two points will dominate the next round of PCA. This gives convergence to either $\ell_{\min_1}$ or $\ell_{\min_2}$.

We also see examples of asymptotic saddle points in the model considered in~\S\ref{sec:globconv}. However, it can be shown with overwhelming probability that the finite sample counterparts to these asymptotic saddle points are not fixed points. Thus, in general, we are not concerned with saddle points that are also fixed points, since we have no proof that such a point exists.

\subsubsection{Convergence to Local Minima}
\label{sec:multlocmin}

Unlike PCA (i.e. when $p=2$), there can potentially be many local minima for the energy $F_{p,\delta}$ when $0<p<2$ . Such local minima can also be fixed points for the FMS$_p$ algorithm. Hence, we cannot guarantee that FMS$_p$ converges to an optimal stationary point in general, since it could converge to one such local minima. For example, some local minima are discussed by~\citet[Example 2]{lp_recovery_part1_11} when $\delta =0$. A modified argument can be used to show that local minima still exist when $\delta >0$. We observe that local minima generally occur when points concentrate around lower dimensional subspaces. Another simple example with many local minima is a symmetric case when $D=2$ and $d=1$. Suppose that points are symmetrically distributed on $\sphere^1$: for some even $N \in \nats$, the data set consists of points $\bx_i = (\cos(2 \pi i / N),\sin(2 \pi i / N))^T$, $i=1,\dots,N$. Then, the span of each pair of antipodal points will define a local minimum for the FMS$_p$ energy $F_{p,\delta}$. However, we note that in this case all local minima are also global minima.

\subsection{Convergence to the Global Minimum for a Special Model of Data}\label{sec:globconv}

In this section, we will show that under a certain model of probabilistic generation of the data, the FMS$_p$ algorithm nearly recovers an underlying subspace with overwhelming probability (w.o.p.). By w.o.p., we mean that the probability of recovery is bounded below by an expression of the form $1-C_1 e^{-C_2 N}$, where $C_1$ and $C_2$ are constants with respect to $N$ (but depend on all other parameters, such as $d$, $D$, $p$, and $\delta$). In \S\ref{sec:globconv:prelim} we lay out some necessary concepts for the statement of the theorems, and then in \S\ref{sec:globconv:thm} we state the theorem giving near recovery of the underlying subspace. Next,~\S\ref{sec:globconv:K2d1} extends to a special case of recovery with multiple subspaces.

\subsubsection{Preliminaries}\label{sec:globconv:prelim}

This section proves global convergence for a special model of data. We use a simple version of the most significant subspace model outlined in~\cite{lp_recovery_part1_11}, and much of the notation and concepts are borrowed from this paper. The general setting considers points distributed on the sphere $\sphere^{D-1}$. In~\S\ref{sec:globconv:thm} we consider the special case of one underlying subspace, rather than the more general setting of $K$ distinct underlying subspaces with one most significant. A more general theorem for $K>1$ has been hard to prove, although \S\ref{sec:globconv:K2d1} gives a theorem for approximate recovery in the case $K=2$ and $d=1$. However, we conjecture that a general theorem for $K>1$ holds with any subspace dimension $d$ due to empirical performance of the algorithm on data sets sampled from these distributions.

Let $L_1^*$ be the most significant subspace within our data set. We construct a mixture measure by combining $\mu_i$ for $i=0,...,K$, where $\mu_0$ is the uniform distribution on $\sphere^{D-1}$ and $\mu_i$ is the uniform distribution on $\sphere^{D-1}\cap L_i^*$. In the noisy case, we add an additive noise distribution $\nu_{i,\varepsilon}$ such that $\supp(\mu_i + \nu_{i,\varepsilon}) \subseteq \sphere^{D-1}$. We also require that the $p$th moment of $\nu_{i,\varepsilon}$ is smaller than $\varepsilon^p$ (where $p$ is the robustness parameter for FMS). Finally, we attach weights $\alpha_0\geq 0,\alpha_i > 0$ to the measures $\mu_i$ such that $\sum_{i=0}^K \alpha_i = 1$ and $\alpha_1 > \sum_{i=2}^K \alpha_i$. The mixture distribution is given by
\begin{equation}
\begin{aligned}
\mu_{\varepsilon} = \alpha_0 \mu_0 + \sum_{i=1}^K \alpha_i(\mu_i + \nu_{i,\varepsilon}).
\end{aligned}
\label{mixtmeasnoise}
\end{equation}
We first consider a noiseless version of $\mu_{\varepsilon}$, and then extend the result to the noisy case. The noiseless measure is written as
\begin{equation}
\mu = \alpha_0\mu_0 + \sum_{i=1}^K \alpha_i \mu_i.
\label{mixtmeas1}
\end{equation}
We assume data sampled independently and identically from $\mu$, so that points sampled from $\mu_0$ and $\mu_i$ for $i=2,...,K$ represent pure outliers, and points sampled from $\mu_1$ represent pure inliers. Everything we prove for the spherical model we can generalize to a spherically symmetric model, where outliers are spherically symmetric and symmetrically distributed on $K-1$ less significant subspaces, and inliers are symmetrically distributed on the most significant subspace $L_1^*$. Note that in practice, normalizing the latter distribution to the sphere $\sphere^{D-1}$ yields a mixture measure of the form~\eqref{mixtmeas1}.

\subsubsection{Global Convergence Theorem for $K=1$}\label{sec:globconv:thm}
We are now ready to state a global convergence theorem for FMS$_p$.

\begin{theorem}[Probabilistic Recovery of the Underlying Subspace]
	Let $\cX$ be sampled independently and identically from the mixture measure $\mu$ given in~\eqref{mixtmeas1} with $K=1$.  Then, for any $ 0 < \eta \leq \pi/6 $ and $0<p\leq 1$, the FMS$_p$ algorithm converges to an $\eta$-neighborhood of the underlying subspace $L_1^*$ w.o.p.~at least
	\begin{equation}\label{eq:wopFMS0p1}
	1 - C_1 e^{-C_2 N  (p\delta)^{2(1-p)/(2-p)} \min\left( \left(\frac{\pi}{6}\right)^{2(p-1)},\frac{\eta^2}{(p\delta)^2}\right) }.
	\end{equation}
	For $1<p<2$, the FMS$_p$ algorithm converges to an $\eta$-neighborhood of the underlying subspace $L_1^*$ w.o.p.~at least
	\begin{equation}\label{eq:wopFMS1p2}
	1 - C_1' e^{-C_2' N \min\left(\eta^{2(p-1)},\frac{\eta^2}{(p\delta)^2}\right) }.
	\end{equation}
    For comparison, using the same techniques to analyze PCA ($p=2$), PCA outputs a subspace in an $\eta$-neighborhood of $L_1^*$ w.o.p.~at least
	\begin{equation}\label{eq:woppca}
	1 - C_1'' e^{-C_2'' N \eta^2}.
	\end{equation}
	Here, $C_2$, $C_2'$, and $C_2''$ have no dependence on $N$, $\eta$, $p$, or $\delta$, but may depend on $D$ and $d$.
	\label{thm:globconv}
\end{theorem}

The proof of this theorem is given in \S\ref{sec:globconv:proof}. This theorem gives a probabilistic near recovery result for the FMS$_p$ algorithm and PCA. We note that the result given for PCA is comparable to the asymptotic result of~\citet[Proposition 2.1]{Vershynin_sample_cov11}, albeit by a different argument. Our result is more restricted, though, since the result of~\citet[Proposition 2.1]{Vershynin_sample_cov11} applies for any $\eta > 0$. Also, our estimates for the PCA constants (see below) are not ideal (again see [58]); however, this is not an issue since we are more interested in contrasting the dependence of these probabilities on $\eta$. There is no restriction on $\alpha_0$ and $\alpha_1$ in~\eqref{mixtmeas1}, although the probability of recovery depends on the fractions. Bounds for the constants $C_2$, $C_2'$ and $C_2''$ can be seen later in~\eqref{eq:probnostatfin},~\eqref{eq:probnostatfin2}, and~\eqref{eq:probnostatfinpca} respectively. Worst case estimates of $C_1$, $C_1'$ and $C_1''$ are given later in Proposition~\ref{prop:coveringfms}, where we examine their dependence on $d$, $D$, $\eta$, $p$, and $\delta$. 

This theorem shows the benefit of using FMS$_p$ over PCA. For the following
discussion, we follow our default choice of the algorithm and assume that $\delta$ is on the order of machine precision. This means that we only need to consider the first term within the minimum function in~\eqref{eq:wopFMS0p1} (since $\eta$ cannot be lower than machine precision). Examining dependence on $\eta$ for the probability bounds, the exponent in the PCA formulation is $O(\eta^2 N)$, the FMS$_1$ exponent is $O(N)$, and for $p<1$ the FMS$_p$ exponent is $O((p\delta)^{2(1-p)/(2-p)}N)$. Altogether, this means that FMS$_p$ is expected to have much more precise recovery for vastly smaller sample sizes. We also advocate choosing $p=1$ when running the FMS$_p$ algorithm for this reason: when $\delta$ is chosen to be very small in this way, the $O((p\delta)^{2(1-p)/(2-p)}N)$ exponent for $p<1$ leads to a much worse bound than the $O(N)$ exponent for $p=1$. Another consequence of this theorem is that we generally advocate for larger values of $\delta$ for smaller values of $p$ (although we do not have optimal expressions for this choice of $\delta$). For demonstrations of the phase transitions exhibited by the probability of recovery, see Figures~\ref{fig:phasetrans_pca_0noise},~\ref{fig:phasetrans_fmsp1_0noise}, and~\ref{fig:phasetrans_fmsp5_0noise}. Again, we emphasize the difference between the bounds on the probabilities of $\eta$-recovery for PCA, FMS$_1$, and FMS$_{0.5}$: their bounds are $1-C_1''e^{-C_2''\eta^2 N}$, $1-C_1e^{-C_2 O(1) N}$, and $1-C_1e^{-C_2 O(\delta^{2/3}) N}$ respectively.

The theoretical result of Theorem~\ref{thm:globconv} extends to the noisy mixture measure~\eqref{mixtmeasnoise} as well.
\begin{theorem}
	Let $\cX$ be sampled independently and identically from the noisy mixture measure $\mu_\varepsilon$ given in~\eqref{mixtmeasnoise}.  Then for any $ 0 < \eta \leq \pi/6 $, if $0<p\leq 1$ and
	\begin{equation}
	\varepsilon < \left( \frac{1}{4} \frac{2}{\pi d^{5/2}} \min\left(\left(\frac{\pi}{6}\right)^{p-1} \eta,\frac{\eta^2}{p\delta} \right) \right)^{1/p},
	\end{equation}
	the FMS$_p$ algorithm converges to an $\eta$-neighborhood of $L_1^*$ w.o.p.~stated in~\eqref{eq:wopFMS0p1}. If $1 < p < 2$ and
	\begin{equation}
	\varepsilon < \left( \frac{1}{4} \frac{2}{\pi d^{5/2}} \min\left( \eta^{p},\frac{\eta^2}{p\delta} \right) \right)^{1/p},
	\end{equation}
    the FMS$_p$ algorithm converges to an $\eta$-neighborhood of $L_1^*$ w.o.p.~stated in~\eqref{eq:wopFMS1p2}.

    For comparison, using the same techniques to analyze PCA, if
	\begin{equation}
	\varepsilon < \left( \frac{1}{4} \frac{2}{\pi d^2}  \eta^2  \right)^{1/2},
	\end{equation}
	PCA outputs a subspace in an $\eta$-neighborhood of $L_1^*$ w.o.p.~stated in~\eqref{eq:woppca}.
	\label{thm:globconv:stab}
\end{theorem}
The proof of Theorem~\ref{thm:globconv:stab} is given in~\S\ref{sec:globconv:proof:stab}. Among choices of $p$, choosing larger values of $p$ seems to give the most robustness to noise. This theorem indicates that PCA has the best stability to noise, although these estimates are not ideal. This result stands in contrast to the result of~\cite{Coudron_Lerman2012}, which shows a higher robustness to noise for a convex relaxation of $F_{1,\delta}$. Less robustness to noise for FMS$_p$ may be attributable to non-convexity, but we cannot make any definitive statement on this fact. In the future, we plan to follow~\citet{Coudron_Lerman2012} and establish the stronger robustness to noise of FMS$_p$ at least when $p=1$. For demonstrations of the phase transitions exhibited by the probability of recovery for this noisy model, see Figures~\ref{fig:phasetrans_pca_1e5noise},~\ref{fig:phasetrans_fmsp1_1e5noise}, and~\ref{fig:phasetrans_fmsp5_1e5noise}.

\subsubsection{Global Convergence Theorem for $K=2$ and $d=1$}
\label{sec:globconv:K2d1}

Another important setting of the most significant subspace model where PCA does not recover the underlying subspace is when $K>1$. We have found it hard to prove anything in general for the case $K>1$ because it is hard to characterize the derivative of $F_{p,\delta}$ in general. However, we are able to prove near recovery for the case $K=2$ and $d=1$.
\begin{theorem}[Probabilistic Recovery for $K=2$ and $d=1$]\label{thm:K2d1}
	Let $\cX$ be sampled independently and identically from the mixture measure $\mu$ in~\eqref{mixtmeas1} with $K=2$ and $d=1$. Then, for any $ 0 < \eta \leq \pi/6 $ and $0<p \leq 1$, the FMS$_p$ algorithm with PCA initialization converges to a point in $\overline{B(L_1^*,\max(\eta,\arcsin((p\delta)^{1/(2-p)})))}$ w.o.p.
\end{theorem}
The proof of Theorem~\ref{thm:K2d1} is given in~\S\ref{sec:K2d1:proof}. It has proven too hard to derive bounds or closed form expressions for the constants in the probability bound, and so we do not present them here. Further, a similar stability result as that in Theorem~\ref{thm:globconv:stab} holds for Theorem~\ref{thm:K2d1}, however we do not display it here. We also note that this Theorem only holds for $0<p\leq 1$. This case is particularly important because the guarantees of other algorithms, such as those of~\citet{moitra_pca2012} and~\citet{robust_pca_ZL}, break down in this setting. Further, although it is very specific, it is an important example for us because FMS$_p$ is still able to recover the correct subspace in the presence of another potential local minimum $L_2^*$. Finally, this is a clear example where PCA cannot recover the most significant subspace asymptotically while FMS$_p$ can.

\subsection{Rate of Convergence for FMS$_p$ Under~\eqref{mixtmeas1}}
\label{sec:convrate}

For this section, we define $L^*$ to be a stationary limit point of the FMS$_p$ algorithm. We begin with a probabilistic global rate of convergence bound for the FMS$_p$ algorithm under~\eqref{mixtmeas1} when $K=1$ or $K=2$ and $d=1$. Under these models, Theorem~\ref{thm:globconv} and Theorem~\ref{thm:K2d1} show that $L^*$ is near to $L_1^*$ (the underlying subspace) w.o.p. The proof of Theorem~\ref{thm:sublinconv} is given in~\S\ref{sec:proof:sublinconv}.
\begin{theorem}[Probabilistic Global Convergence Bound]\label{thm:sublinconv}
	Suppose that $\cX$ is sampled i.i.d.~from the mixture measure $\mu$ in~\eqref{mixtmeas1} with $K=1$. Then, for $0<p\leq 1$, the number of iterations $T$ such that $\dist(L_T,L_1^*) < \eta$ is at worst 
	\begin{equation}
		T=O\left(\frac{1}{\min\left( \left(\frac{\pi}{6}\right) ^{2(p-1)} p\delta ,\frac{\eta^2}{(p\delta)}\right)}\right) \ (\text{w.o.p.}).
	\end{equation}
	In contrast, for $1<p < 2$, the number of iterations is at worst
	\begin{equation}
		T=O\left(\frac{1}{\min\left(\eta^{2(p-1)} p\delta,\frac{\eta^2}{(p\delta)}\right)}\right) \ (\text{w.o.p.}).
	\end{equation}
	 For $\mu$ with $K=2$ and $d=1$ and $0 < p \leq 1$, the number of iterations $T$ such that $ \arcsin((p\delta)^{1/(2-p)}) \leq \dist(L_T,L_1^*) < \eta$ is at worst
	 \begin{equation}
		 T=O\left(\frac{1}{(\alpha_1 - \alpha_2)^2 p\delta}\right) \ (\text{w.o.p.}).
	 \end{equation}
\end{theorem}

Beyond this, Theorem~\ref{thm:linconv} yields local $r$-linear convergence w.o.p.~for the FMS$_p$ algorithm under~\eqref{mixtmeas1} when $K=1$ or when $K=2$ and $d=1$. The proof of Theorem~\ref{thm:linconv} is given in~\S\ref{sec:proof:linconv}.
\begin{theorem}[Probabilistic Local Linear Convergence]\label{thm:linconv}
	Suppose that $\cX$ is sampled i.i.d.~from the mixture measure $\mu$ in~\eqref{mixtmeas1} with $K=1$ (or $K=2$, $d=1$, $\alpha_1 > (2-p) \alpha_2$, and $\dist(L_1^*,L_2^*) > 2 \arcsin(p\delta^{1/(2-p)})$). Then, w.o.p., there exists an index $\kappa$ such that $(L_k)_{k>\kappa}$ converges $r$-linearly to its limit point $L^*$.
\end{theorem}
The bound on the rate of this $r$-linear convergence can be seen in the proof of Theorem~\ref{thm:linconv}: specifically see~\eqref{eq:lambdakbound} and~\eqref{eq:linearconv}. Theorems~\ref{thm:sublinconv} and~\ref{thm:linconv} can be combined to give a bound on the overall iteration complexity of FMS$_p$. In general, given a choice of $p \leq 1$, the number of iterations required to converge is bounded by $O\left(1/(p\delta \min(1,\eta^{3(p-1)}))\right)$ (or $O\left(1/(p\delta(\alpha_1 - \alpha_2)^4)\right)$ when $K=2$ and $d=1$). However, once the iterates are sufficiently close to the limit point, the iteration complexity becomes $O(\log(1/\eta))$. Figure~\ref{fig:linconvergence_verification} verifies that on a data set sampled from~\eqref{mixtmeas1} with $K=1$, the convergence of FMS$_1$ and FMS$_{0.5}$ is $r$-linear.

%%%%%%%%%%%%%%%%%%%%%%%%%%%%%%%%%%%%%%%%%%%%%%%%%%%%%%%%%%%%%%%%%%%%%%%%%%%%%%%%%%%%%%%%%%%%%%%%%%%%%%%%%%%%%%%

\section{Numerical Experiments}\label{sec:numericalexp}

In this section, we illustrate how the FMS$_p$ algorithm performs on various synthetic and real data sets in a MATLAB test environment (except for \S\ref{sec:scalability}, which was run in Python). It was most interesting for us to test FMS$_p$ with the value of $p=1$, in order to compare it with various convex relaxations of its energy in this case. FMS$_1$ denotes a case where the algorithm run with a value of $p=1$. In certain cases, we have not noticed a difference in the performance by choosing lower values of $p$ and will make clear when this is the case. In places where we noted such a difference we report them with $p=0.1$ and let FMS$_{0.1}$ denote the case of $p=0.1$. We set the parameter $\epsilon$ to be $10^{-10}$.

The algorithms we compare with are the Tyler M-estimator~\cite{Teng_log_rpca}, Median K-flats (MKF)~\cite{MKF_workshop09}, Reaper~\cite{LMTZ2014}, R1-PCA~\cite{Ding+06}, GMS~\cite{robust_pca_ZL}, and Robust Online Mirror Descent (R-MD)~\cite{online_robust_pca13}. We also tried a few other algorithms, in particular, HR-PCA and DHR-PCA~\cite{xu2013outlier,Feng:12}, LLD~\cite{robust_mccoy}, and Outlier-Pursuit~\cite{Xu2012},  but they were not as competitive; we thus do not report their results. For example, both HR-PCA and DHR-PCA were slower and surprisingly worse than PCA in many of our tests. Even though MKF and R-MD were not competitive either in many cases, it was important for us to compare with online algorithms. The comparison with R1-PCA was also important since its aims to directly minimize~\eqref{costf2} when $p=1$ and $\delta=0$.
In addition, we compared with principal component pursuit (PCP)~\cite{candes_wright_robust_pca09,Lin_Chen_Ma_2010}, which aims to solve the robust PCA problem. The code chosen for this comparison was the Accelerated Proximal Gradient with partial SVD~\cite{rpca_code09} obtained from \url{http://perception.csl.illinois.edu/matrix-rank/sample_code.html}, although similar results were given by the ALM codes~\cite{Lin_Chen_Ma_2010}. We emphasize that Robust PCA methods are designed for the regime where there are sparse, element-wise corruptions of the data matrix, rather than the wholly corrupted data points, which we consider in this paper. We have noticed that robust PCA algorithms based on this model exhibit quite poor performance compared to algorithms tailored for RSR when data points are wholly corrupted.

Experiments by~\cite{LMTZ2014} demonstrate the advantage of scaling each centered data point by its norm, i.e., by ``spherizing'' each data point (equivalently, projecting onto the unit sphere). In cases where we examine the effect of ''spherizing'' the data, we denote an algorithm run on a spherized data set with the prefix S. For example, as a baseline in many of the experiments we compare with SPCA, performed by using the RandomizedPCA~\cite{RST_random_pca} algorithm to find the top $d$ singular vectors of the spherized data. In the same vein, we sometimes also compare with SFMS$_p$, which is FMS$_p$ run on spherized data. Spherizing seems to reduce noise and produce a better subspace in some cases: we will display results from SFMS$_p$ and SPCA when this is the case (but omit them when there is no difference).

Tyler M-estimator is used with regularization parameter $\epsilon = 10^{-10}$. Median K-Flats passes over the data many times to find a single subspace of dimension $d$, with a step size of 0.01 and maximum number of iterations 10000. The Reaper algorithm is run with the regularization parameter $\delta = 10^{-20}$. R1-PCA uses stopping parameter $10^{-5}$ and is capped at 1000 iterations. R-MD passes over the data 10 times and uses step size $1/\sqrt{k}$ at iteration $k$. The PCP parameter was set as $\lambda = 1/\sqrt{\max(D,N)}$.
%We center all data points by the geometric median.

\subsection{Synthetic Experiments}\label{exp:synth}

%\begin{figure}[ht!]
%\centering
%\vspace{-.7cm}
%\subfloat[\label{errplot}]{\includegraphics[trim=100 240 100 250,clip,width=0.5\textwidth]{errplot.pdf}}
%\subfloat[\label{errplottimes}]{\includegraphics[trim=100 240 100 250,clip,width=0.5\textwidth]{errplottimes.pdf}} \\
%\vspace{-.5cm}
%\subfloat[\label{timeplot}]{\includegraphics[trim=100 240 100 250,clip,width=0.5\textwidth]{timeplot.pdf}}
%\subfloat[\label{varplot}]{\includegraphics[trim=100 240 100 250,clip,width=0.5\textwidth]{errscale.pdf}}
%\vspace{-.3cm}
%\caption{
%Demonstration of recovery error and runtime on artificial instances for FMS$_p$ and other competitive RSR algorithms.
%Fig.~\ref{errplot} and Fig.~\ref{errplottimes} demonstrate the accuracy and runtime versus percentage of outliers (bounded by the Hardt \& Moitra's upper bound for RSR; see~\cite{moitra_pca2012}). In both measures, FMS$_p$ shows state-of-the-art performance. Fig.~\ref{timeplot} demonstrates runtime versus ambient dimension while the subspace dimension remains fixed at $d=5$. The runtime of the FMS$_p$ is superior to existing methods. Fig.~\ref{varplot} demonstrates recovery error versus the ratio of the variance between outliers and inliers.}
%%\vspace{-.2cm}
%\end{figure}

A series of synthetic tests are run to determine how the FMS$_p$ compares to other state-of-the-art algorithms.
In all of these examples the data is sampled according to variants of the needle-haystack model of~\cite{LMTZ2014}. More precisely, inliers are sampled from a Gaussian distribution within a random linear $d$-subspace in $\mathbb{R}^D$, and outliers are sampled from a Gaussian distribution within the ambient space. Noise is also added to all points. In all of these experiments, the fraction of outliers is restricted by Hardt \& Moitra's upper bound for RSR, that is, $(D-d)/D$~\cite{moitra_pca2012}.

\begin{figure}[h!]
	{\centering
		\subfloat[\label{errplot}]{\includegraphics[trim=60 200 40 210,width=.5\textwidth]{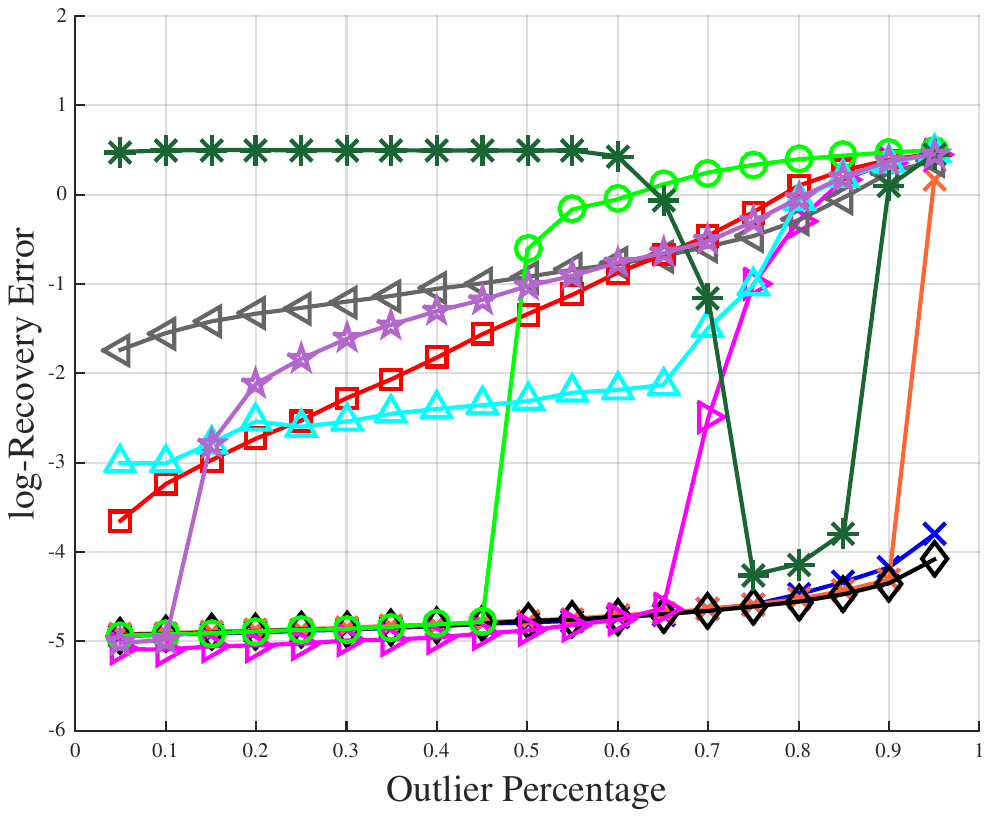}}
		\subfloat[\label{errplottimes}]{\includegraphics[trim=60 200 40 210,width=.5\textwidth]{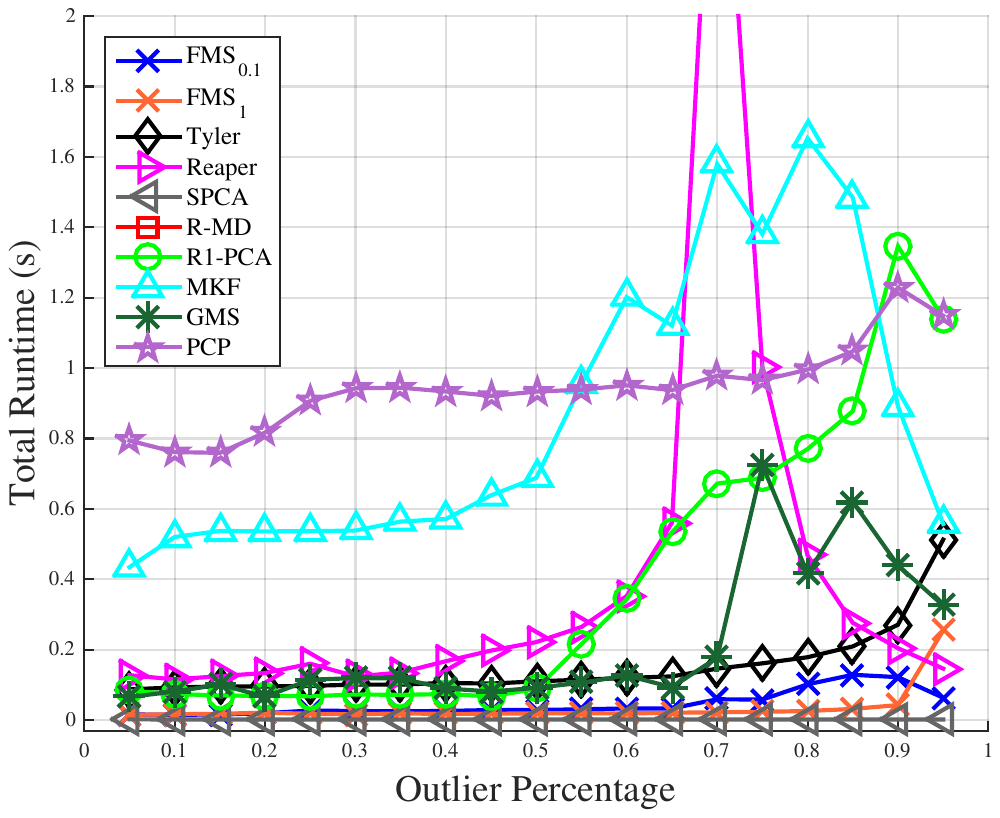}}}
	%\vspace{-.3cm}
	\caption{Plots demonstrating the accuracy and total runtime of some subspace recovery algorithms versus the percentage of outliers in the data set (bounded by the upper bound of~\cite{moitra_pca2012} for RSR). FMS$_1$ and FMS$_{0.1}$ obtain both competitive time and accuracy, with the exception of FMS$_1$ for $95\%$ outliers.}\label{errplotfig}
	%\vspace{-.1cm}
\end{figure}

The first experiment demonstrates the effect of the percentage of outliers on the total runtime and error for subspace recovery.
Let $\mathbf{\Sigma}_{in}$ denote the orthogonal projector onto the randomly selected subspace, %scaled by $1/d$,
and let $\mathbf{\Sigma}_{out}$ denote the identity transformation on $\mathbb{R}^D$.
In this experiment, inliers are drawn from the distribution $\mathcal{N}(0,\mathbf{\Sigma}_{in}/d)$ and outliers from the distribution $\mathcal{N}(0,\mathbf{\Sigma}_{out}/D)$. Scaling by $1/d$ and $1/D$ respectively ensures that both samples have comparable magnitudes. Error is measured by %finding the output subspace given by each algorithm and
calculating the distance between the found and ground truth subspaces.
The ambient dimension is fixed at $D=100$, the subspace dimension is $d=5$, and the total number of points is fixed at $N=200$. Every data point is also perturbed by added noise drawn from $\mathcal{N}(0,10^{-6}\mathbf{\Sigma}_{out})$.

Figure~\ref{errplotfig} displays results for recovery error and total runtime versus the percentage of outliers in a data set. At each percentage value, the experiment is repeated on 20 randomly generated data sets and the results are then averaged. We note that the runtime of R-MD is too large to be displayed on the graph of total runtime. FMS$_{0.1}$ and Tyler M-estimator have the best accuracy on this data, while FMS$_1$ only demonstrates problems at high ends of outlier percentages. Out of the robust methods, FMS$_{1}$ and FMS$_{0.1}$ are the fastest (excluding the high end of outlier percentage). It is interesting to note in these figures that GMS fails for lower percentages of outliers;~\citet{robust_pca_ZL} acknowledge that to be safe, GMS needs at least $1.5(D-d)$ outliers are needed to ensure recovery. The authors advocate either initial dimensionality reduction or the addition of synthetic outliers to increase the chances of finding the correct subspace. In our tests, initial dimensionality reduction was still not competitive, but the addition of synthetic outliers results in precise recovery (although we do not show this to illustrate the deficiency of GMS in low percentages of outliers).

\begin{figure}[h!]
	\centering
	\subfloat[\label{timeplot}]{\includegraphics[trim=60 200 40 210,width=.5\textwidth]{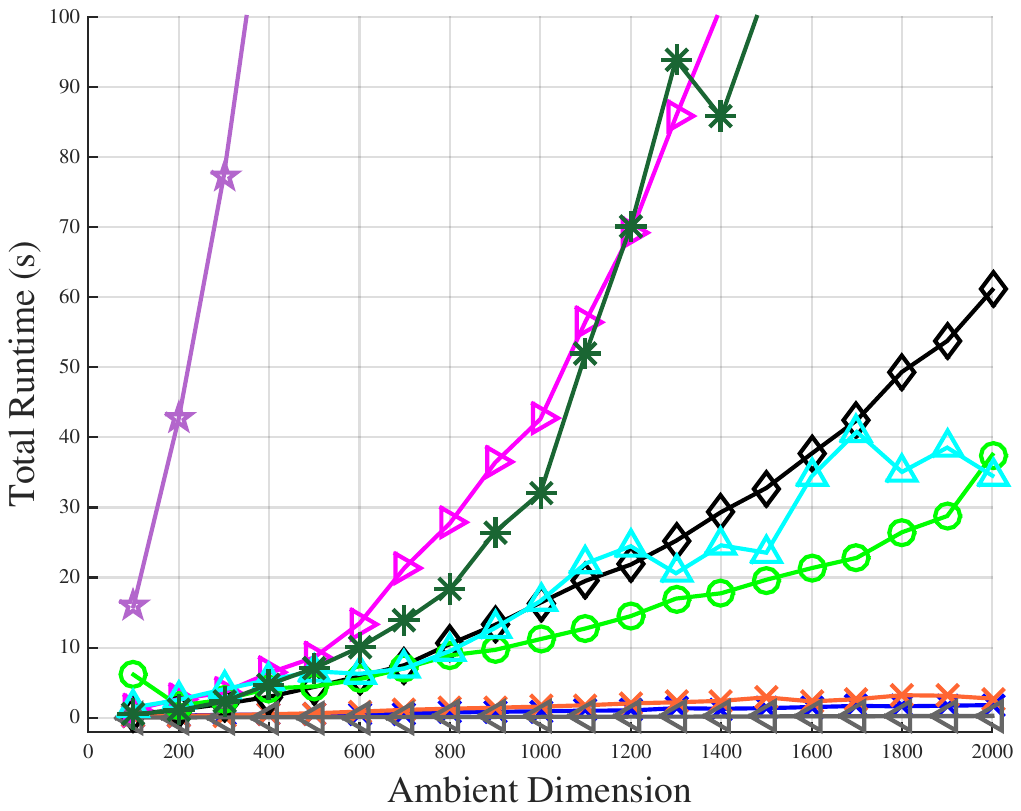}}
	\subfloat[\label{timeploterr}]{\includegraphics[trim=60 200 40 210,width=.5\textwidth]{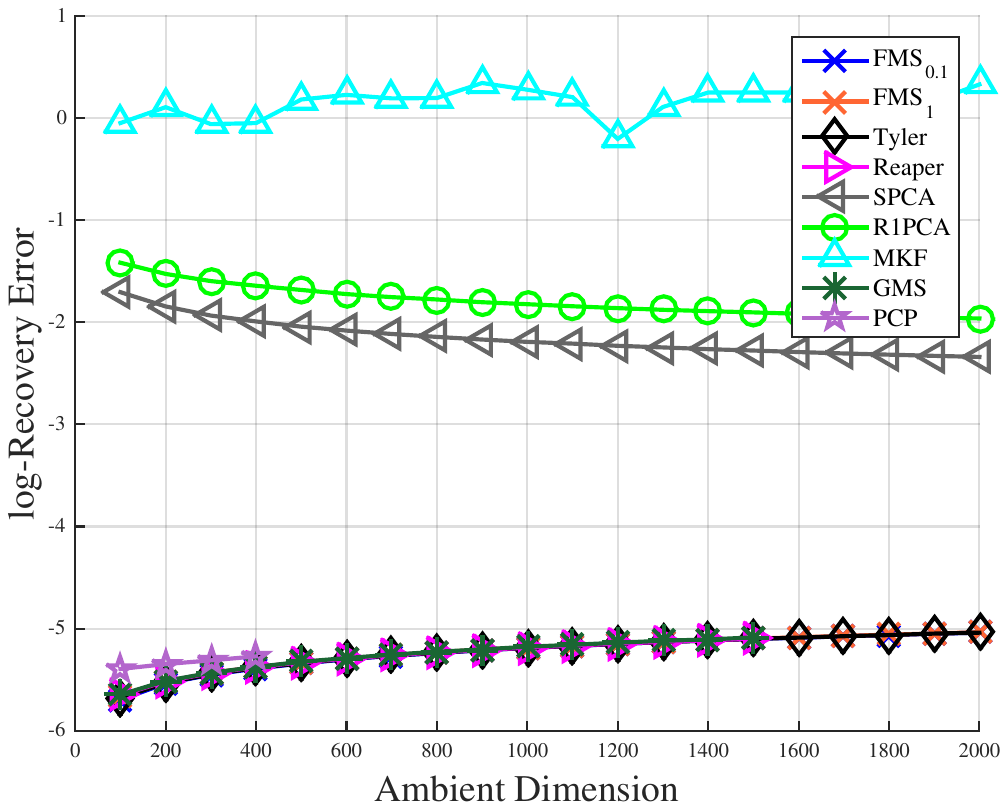}}
	\caption{Demonstration of accuracy and total runtime for various subspace recovery algorithms. The left figure displays how total runtime varies versus the ambient dimension. The right figure shows the corresponding recovery error for each ambient dimension. The runtime experiments were cut off when the algorithm exceeded 100 seconds. The runtimes of FMS$_{0.1}$  and FMS$_1$ are superior to existing methods. FMS$_{0.1}$, FMS$_1$, Tyler, Reaper, and GMS all achieve competitive accuracy on these data sets (PCP also does for low ambient dimension, but we were unable to run in higher ambient dimension due to poor computational complexity).}\label{timeplotfig}
\end{figure}

A second experiment is displayed in Figure~\ref{timeplotfig}, where we demonstrate the total runtime superiority of FMS$_p$ versus other RSR algorithms. In the Figure~\ref{timeplot}, the runtime is plotted as a function of ambient dimension for different algorithms. In Figure~\ref{timeplot}, the corresponding errors for these runtimes are given. Here we fix the total number of points at $N=6000$ with 3000 inliers and 3000 outliers. The subspace dimension is fixed at $d=5$, and the variance model for the sampled points is as before. Again, all points also have noise drawn from $\mathcal{N}(0,10^{-6}\mathbf{\Sigma}_{out})$. The ambient dimension is varied from 100 to 2000, and for each method runtime is cut off at 100 seconds. The plotted runtime is averaged over 20 randomly generated data sets. Robust Online Mirror Descent is not shown in Figure~\ref{timeplotfig} due to the very large runtime required for higher dimensions. For the data sets tested here, FMS$_1$, Reaper, GMS, PCP, and Tyler M-estimator all precisely found the subspace for each ambient dimension. Among these, we note that FMS$_1$ has the best runtime at higher ambient dimension due to its lower complexity, while algorithms like Reaper, GMS, PCP, and Tyler do not scale nearly as well. We note that the PCP algorithm is especially slow on such data, requiring a large runtime for even relatively low dimensional data. Runtimes were calculated on a computer with 8 GB RAM and an Intel Core i5-2450M 2.50 GHz CPU. We remark that not only is FMS$_p$ faster than Tyler M-estimator, it also does not require initial dimensionality reduction, which is required for Tyler M-estimator when applied for subspace recovery~\cite{Teng_log_rpca}.

\begin{figure}[h!]
	\centering
	\includegraphics[trim=100 240 100 250,clip,width=.55\textwidth]{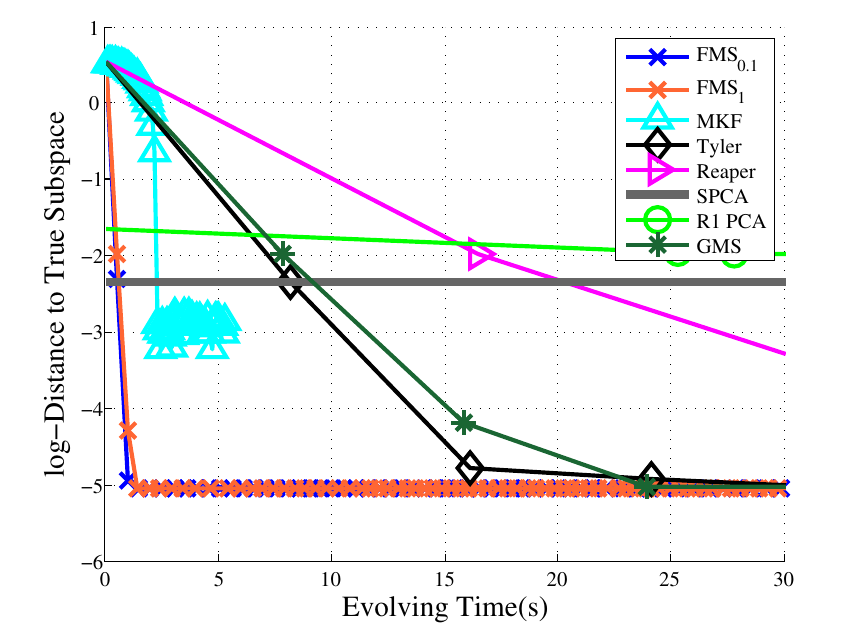}
	\caption{Demonstration of error as time evolves for subspace recovery algorithms. Marks appear per iterations for all algorithms but MKF (1 mark per 100 iterations) and PCA (no marks since there is no iteration). FMS$_1$ and FMS$_{0.1}$ demonstrate the fastest convergence to an accurate subspace among all existing methods.}\label{convergence_per_time}
\end{figure}

In Figure~\ref{convergence_per_time}, we demonstrate accuracy achieved by different RSR algorithms as a function of  evolving time. The data set has $N=6000$ points consisting of 3000 inliers and 3000 outliers, with ambient dimension $D=2000$, subspace dimension $d=5$, and added noise drawn from $\mathcal{N}(0,10^{-6}\mathbf{\Sigma}_{out})$. Each mark on the graph corresponds to the accuracy achieved by the given algorithm after a certain amount of runtime has passed. Clearly, FMS$_p$ has the fastest convergence of the existing RSR algorithms, and achieves competitive accuracy in a matter of seconds. PCP is not shown here due to the large amount of time required to complete even one iteration.

\begin{figure}[h!]
	\centering
	\includegraphics[trim=70 240 80 240,clip,width=.55\textwidth]{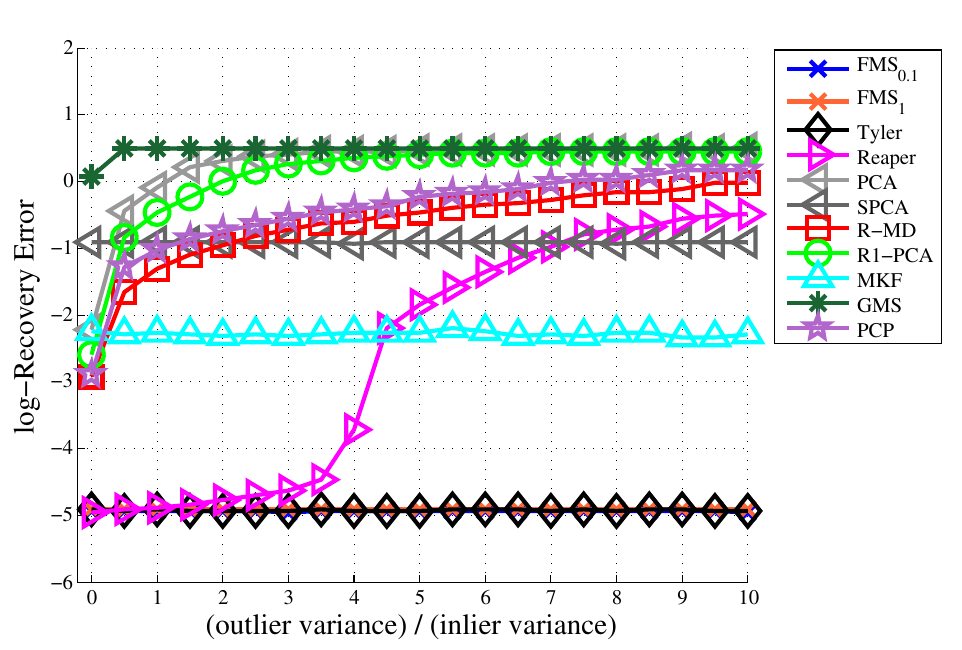}
	\caption{Demonstration of recovery error versus the scale ratio of the variance between inliers and outliers. FMS$_{0.1}$, FMS$_1$, and Tyler are the only competitive algorithms (with identical output).}\label{varplot}
\end{figure}

A final test on synthetic data displays the accuracy when the scale of the variance is different between the inliers and outliers. For this experiment, inliers are still drawn from the distribution $\mathcal{N}(0,\mathbf{\Sigma}_{in}/d)$. The outliers are drawn from $\mathcal{N}(0,\lambda\mathbf{\Sigma}_{out}/D)$, where $\lambda$ is a scaling parameter used to change the variance. All points have noise drawn from $\mathcal{N}(0,10^{-6}\mathbf{\Sigma}_{out})$. The plot in Figure~\ref{varplot} displays the resulting error from various algorithms as the scaling parameter is changed. All points are the average error over 20 randomly generated data sets. FMS$_1$ and Tyler M-estimator both have perfect performance across scale. FMS$_{0.1}$ is not displayed due to identical performance with FMS$_1$. Again, GMS fails here due to too few outliers: the addition of synthetic outliers leads to better results in this figure (although it is not as competitive as Tyler M-Estimator and FMS$_p$).

The takeaway from these tests should be that the FMS$_p$ algorithm offers state-of-the-art accuracy for the synthetic data model while having a complexity that leads to better runtimes in high dimensions.

To conclude this section, we will display some plots verifying the convergence properties of FMS$_p$ to make sure they align with the theory in~\S\ref{sec:theory}. We begin by displaying the phase transition of probabilistic recovery exhibited by FMS$_p$ and PCA under the models~\eqref{mixtmeas1} and~\eqref{mixtmeasnoise} with $K=1$. These figures will validate Theorems~\ref{thm:globconv} and~\ref{thm:globconv:stab}, which states that both FMS$_p$ and PCA have asymptotic recovery of the underlying subspace, but the rate of FMS$_p$ is much better than that of PCA (i.e. FMS$_p$ requires smaller sample sizes for accurate recovery). We set $\alpha_0 = \alpha_1 = 1/2$, and sample sizes were varied. For each sample size, 100 data sets were generated, and the recovery error was calculated as the distance between the found subspace and the underlying subspace $L_1^*$. In the plots, the value at each $\log_{10}(\eta)$ and sample size $N$ is the percentage of times that the recovery error was less than or equal to $\eta$.

\begin{figure}[H]
	\centering
	\includegraphics[trim=40 195 80 215,clip,width=.7\textwidth]{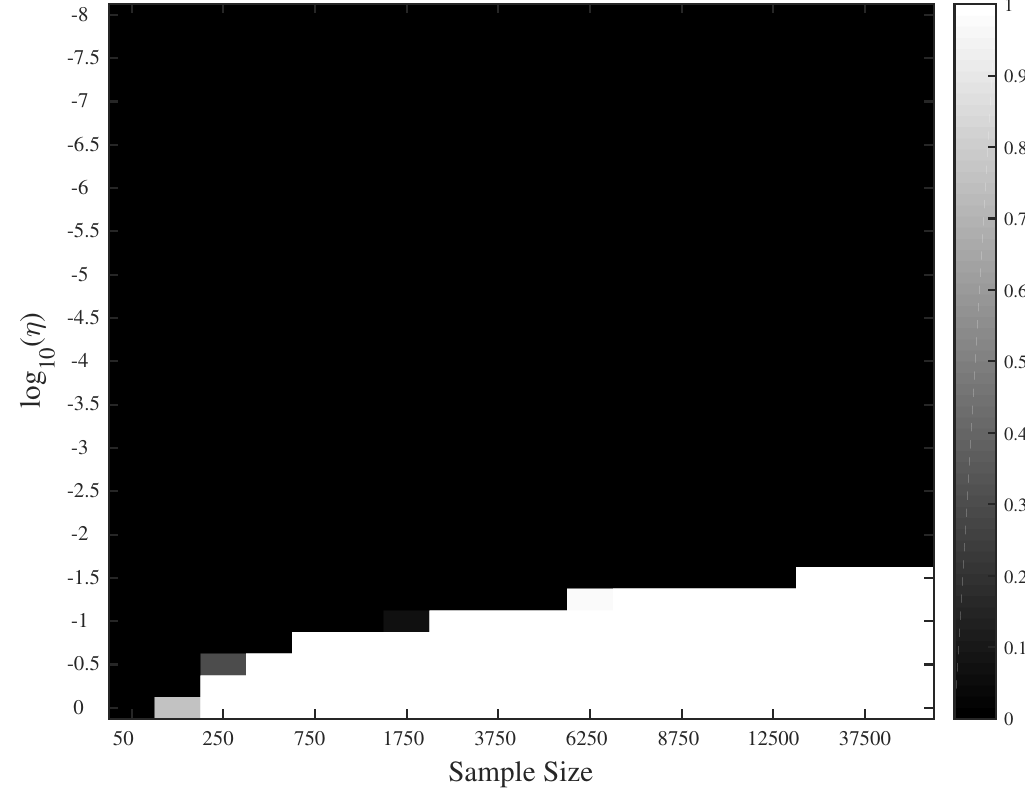}
    \caption{The percentage of times an $\eta$-accurate or better solution was given by PCA with varying sample sizes. The ratio of inliers to outliers here is 1:1, and the data is i.i.d.~sampled from~\eqref{mixtmeas1} with $K=1$, $D=100$, $d=10$.}\label{fig:phasetrans_pca_0noise}
\end{figure}

\begin{figure}[H]
	\centering
	\includegraphics[trim=40 195 80 215,clip,width=.7\textwidth]{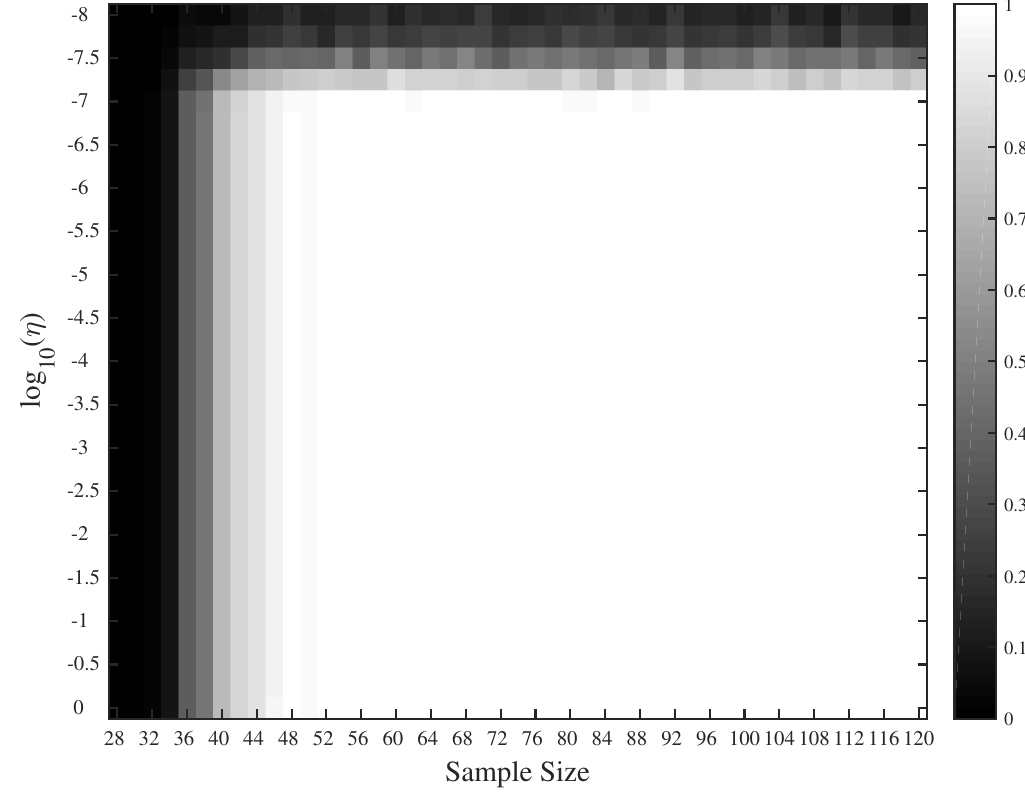}
	\caption{The percentage of times an $\eta$-accurate or better solution was given by FMS$_1$ with varying sample sizes. The ratio of inliers to outliers here is 1:1, and 100 data sets are i.i.d.~sampled from~\eqref{mixtmeas1} with $K=1$, $D=100$, $d=10$.}\label{fig:phasetrans_fmsp1_0noise}
\end{figure}

\begin{figure}[H]
	\centering
	\includegraphics[trim=40 195 80 215,clip,width=.7\textwidth]{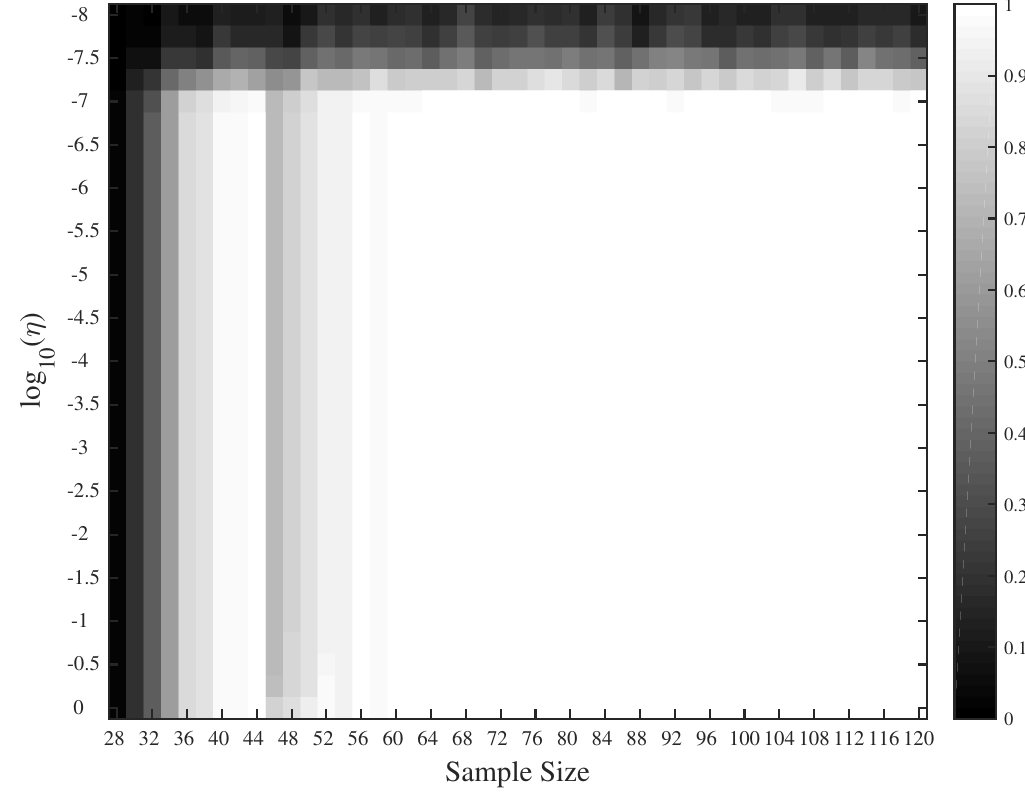}
    \caption{The percentage of times an $\eta$-accurate or better solution was given by FMS$_{0.5}$ with varying sample sizes. The ratio of inliers to outliers here is 1:1, and 100 data sets are is i.i.d.~sampled from~\eqref{mixtmeas1} with $K=1$, $D=100$, $d=10$.}\label{fig:phasetrans_fmsp5_0noise}
\end{figure}

The noiseless case is displayed in Figures~\ref{fig:phasetrans_pca_0noise},~\ref{fig:phasetrans_fmsp1_0noise}, and~\ref{fig:phasetrans_fmsp5_0noise}. Within these figures, FMS$_p$ shows a clear advantage over PCA for the asymptotic rate of recovery for the underlying subspace. Even for very small numbers of points ($N\approx 40$), FMS$_p$  for $p=1$ or $p=0.5$ can approximate the underlying subspace to a precision of $10^{-7}$. On the other hand, PCA can only approximate the subspace to a precision of $10^{-1.5}$ for sample sizes as large as $N=50000$. We do note that FMS$_{0.5}$ does seem to have some trouble around a sample size of $N=48$, which indicates convergence to a non-optimal solution. This fits with earlier theory that indicates this possibility (see Theorem~\ref{thm:globconv} and discussion). Although this may be mitigated with a larger choice of $\delta$, some precision may be lost with larger values of $\delta$.

\begin{figure}[H]
	\centering
	\includegraphics[trim=40 195 80 215,clip,width=.7\textwidth]{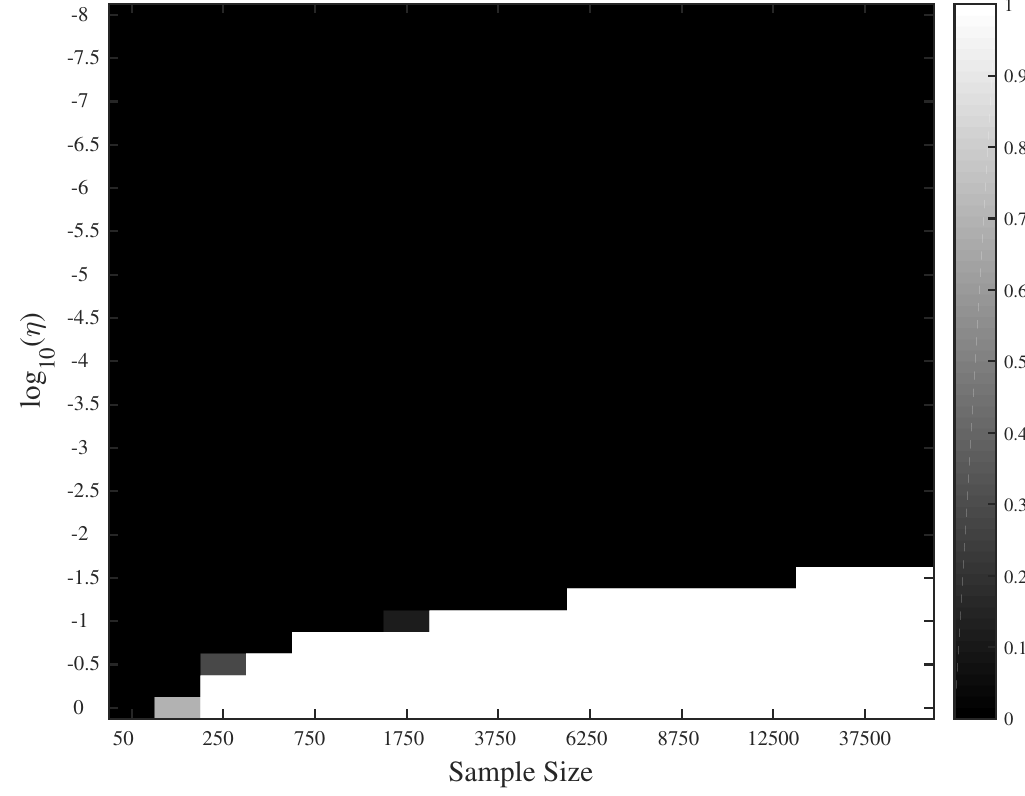}
    \caption{The percentage of times an $\eta$-accurate or better solution was given by PCA with varying sample sizes. The ratio of inliers to outliers here is 1:1, and 100 data sets are i.i.d.~sampled from~\eqref{mixtmeasnoise} with $K=1$, $D=100$, $d=10$, and added Gaussian noise of directional variance $10^{-5}$ (which is projected to $\sphere^{D-1}$).}\label{fig:phasetrans_pca_1e5noise}
\end{figure}

\begin{figure}[H]
	\centering
	\includegraphics[trim=40 195 80 215,clip,width=.7\textwidth]{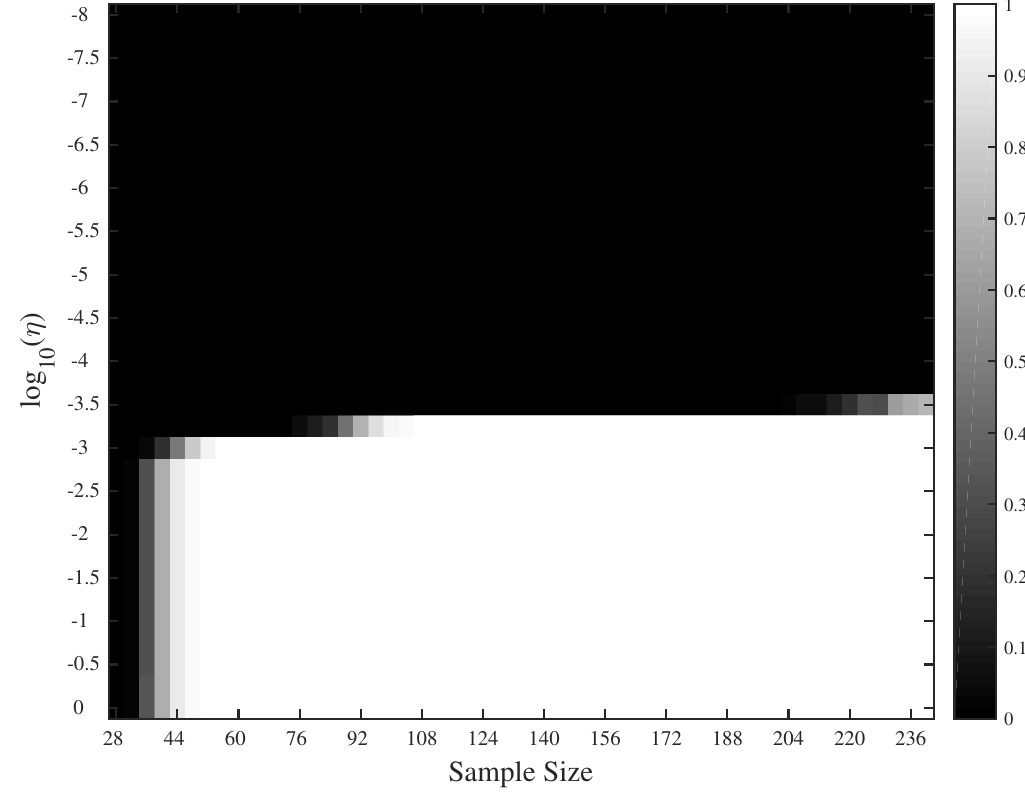}
    \caption{The percentage of times an $\eta$-accurate or better solution was given by FMS$_1$ with varying sample sizes. The ratio of inliers to outliers here is 1:1, and 100 data sets are i.i.d.~sampled from~\eqref{mixtmeasnoise} with $K=1$, $D=100$, $d=10$, and added Gaussian noise of directional variance $10^{-5}$ (which is projected to $\sphere^{D-1}$).}\label{fig:phasetrans_fmsp1_1e5noise}
\end{figure}

\begin{figure}[H]
	\centering
	\includegraphics[trim=40 195 80 215,clip,width=.7\textwidth]{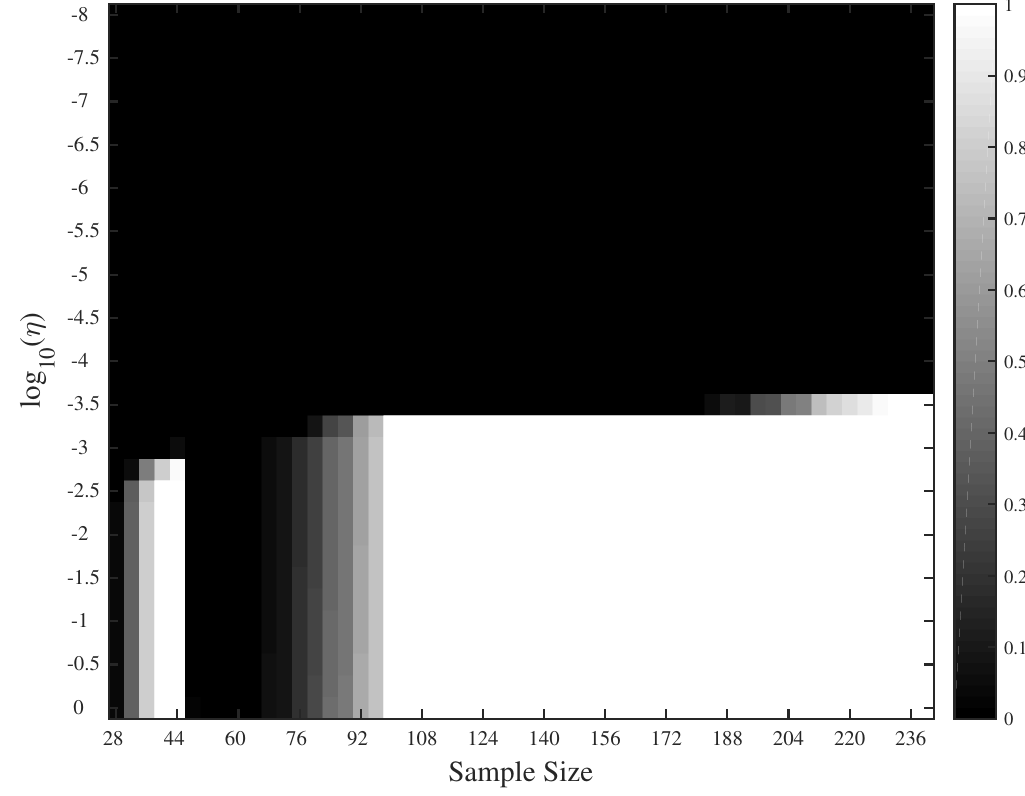}
    \caption{The percentage of times an $\eta$-accurate or better solution was given by FMS$_{0.5}$ with varying sample sizes. The ratio of inliers to outliers here is 1:1, and 100 data sets are i.i.d.~sampled from~\eqref{mixtmeasnoise} with $K=1$, $D=100$, $d=10$, and added Gaussian noise of directional variance $10^{-5}$ (which is projected to $\sphere^{D-1}$).}\label{fig:phasetrans_fmsp5_1e5noise}
\end{figure}

The noisy case is displayed in Figures~\ref{fig:phasetrans_pca_1e5noise},~\ref{fig:phasetrans_fmsp1_1e5noise}, and~\ref{fig:phasetrans_fmsp5_1e5noise}. Within these figures, we notice that the rate of recovery for PCA does not change from the noiseless case. Between the two FMS$_p$ plots, $p=0.5$ seems to have issues with small numbers of points. The issues occur around $N=48$, which is where we saw slight issues in the noiseless case also. The convergence of FMS$_p$ to a non-optimal solution for $p<1$ fits in with Theorem~\ref{thm:globconv} and Theorem~\ref{thm:globconv:stab}. Again, this may be alleviated by choosing larger values of $\delta$, but solutions may not be as precise. When comparing $p=1$ versus $p=0.5$ for larger $N$, it appears that the rate of asymptotic recovery may be better for smaller $p$ (see $N \approx 204$ in Figures~\ref{fig:phasetrans_fmsp1_1e5noise} and~\ref{fig:phasetrans_fmsp5_1e5noise}).


\begin{figure}[H]
	\centering
    \includegraphics[trim=40 180 60 210,clip,width=.7\textwidth]{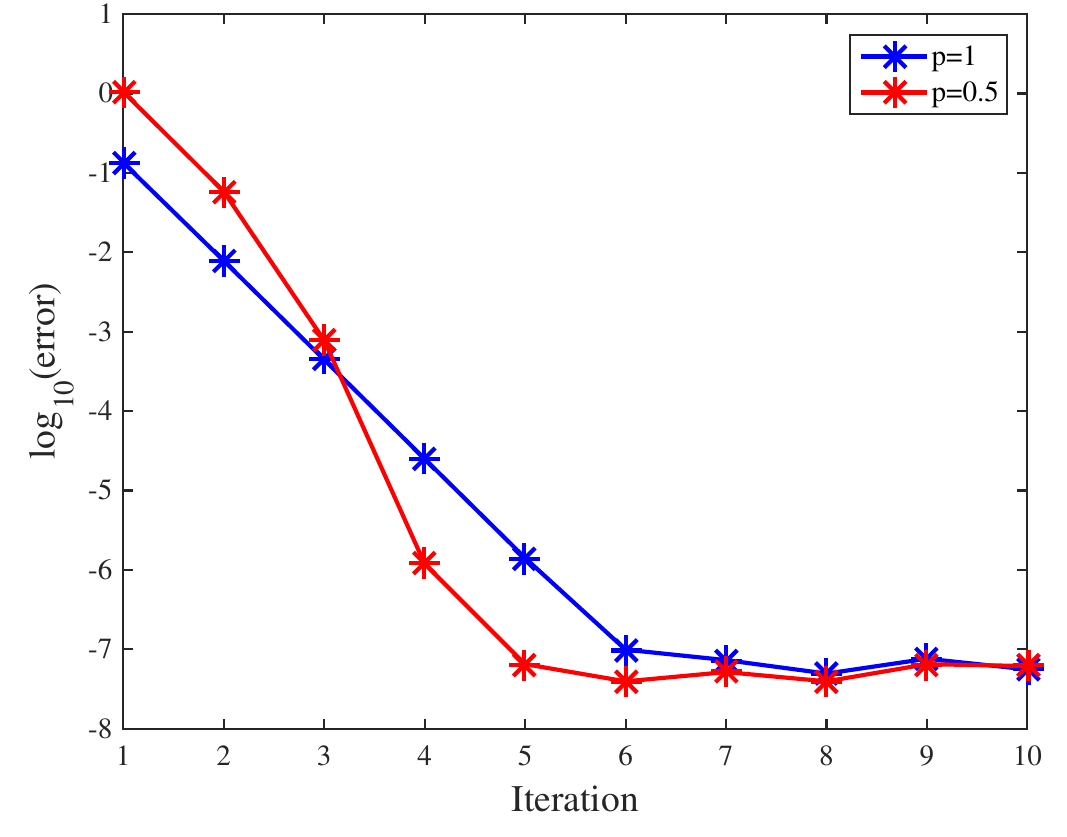}
    \caption{Verification that the FMS$_1$ and FMS$_{0.5}$ algorithms converge $r$-linearly to their limit point under~\eqref{mixtmeas1} when $K=1$. In both cases, the $r$-linear convergence appears to be global, not just local.}\label{fig:linconvergence_verification}
\end{figure}

Finally, we verify that the convergence of FMS$_1$ and FMS$_{0.5}$ is at least locally linear under~\eqref{mixtmeas1} with $K=1$. Figure~\ref{fig:linconvergence_verification} displays $\log_{10}(\dist(L_k,L_1^*))$ versus iteration count $k$. In both cases $p=1$ and $p=0.5$, the convergence to $L_1^*$ is $r$-linear.

\subsection{Real Data Experiments}\label{exp:realdata}

One of the real strengths of PCA in reducing dimensionality comes from its denoising effect. Projection to a subspace by PCA has long been a popular preprocessing step for classification and clustering (see e.g.~\cite{vempala_kannan_spectr_alg,hopcroft_kannan_foundations}) due to this denoising effect. In some cases, RSR and robust PCA algorithms seem to demonstrate higher resistance to noise in data than PCA. The first two experiments displayed in~\S\ref{exp:astro} and~\S\ref{exp:clust} show the viability of FMS$_p$ for denoising. We finish in~\S\ref{exp:faces} with a stylized experiment on real data with explicit outliers to demonstrate the accuracy of FMS$_p$, and then~\S\ref{sec:scalability} demonstrates the ability of FMS$_p$ to scale to truly massive data.

\subsubsection{Eigenspectra Calculation from Astrophysics}\label{exp:astro}
The first experiment demonstrating the usefulness of the FMS$_p$ algorithm on real data that comes from astronomy. The goal of this experiment is to robustly locate eigenspectra in a large set of galaxy spectrum data. The eigenspectra found can be used in the classification of galaxies within the complete data set, since a the galaxy spectra can be decomposed by projection onto the span of the eigenspectra.~\citet{Budavari_astro_09} provide criteria for determining what makes a resulting eigenspectra good. The key attribute of good eigenspectra is that they should not be noisy themselves. This would in turn introduce noise into the decomposition of individual galaxy spectra using the eigenspectra, which in turn leads to inaccurate classification using the reduced spectra. In this experiment, we judge the RSR algorithms on how noisy the eigenspectra they find are.

A data set is taken from the Sixth Data Release of the  Sloan Digital Sky Survey~\cite{2008ApJS_175_297A}. A total of 83686 spectra were taken from databases using code from~\cite{2008asvo_proc_79D}. Spectral reduction was performed to account for resampling, restframe shifting, and PCA gap correction~\cite{2004AJ_128_585Y}. The resulting data consisted of 83686 data points in dimension 3841. To use RSR on this data set, we follow the example of previous work done with RSR for finding eigenspectra~\cite{Budavari_astro_09}. The data is first centered by subtracting the mean spectra from all values. FMS$_{0.1}$, FMS$_1$, RandomizedPCA~\cite{RST_random_pca}, and the Tyler M-estimator are then applied to the data to find the top eigenspectra of the data set. Additionally, we spherize the data and run PCA and FMS$_1$ (SPCA and SFMS$_1$) to see whether it changes the resulting eigenspectra (SFMS$_{0.1}$ is not shown due to similarity with the results of SFMS$_1$).  Other methods are not shown because they either do not do better than standard PCA, or because the methods take too long to be feasibly run due to the large size of the data set.

\begin{figure}[ht]
	%\vspace{-.5cm}
	\centering
	\includegraphics[trim=10 170 20 180,clip,width=.9\textwidth]{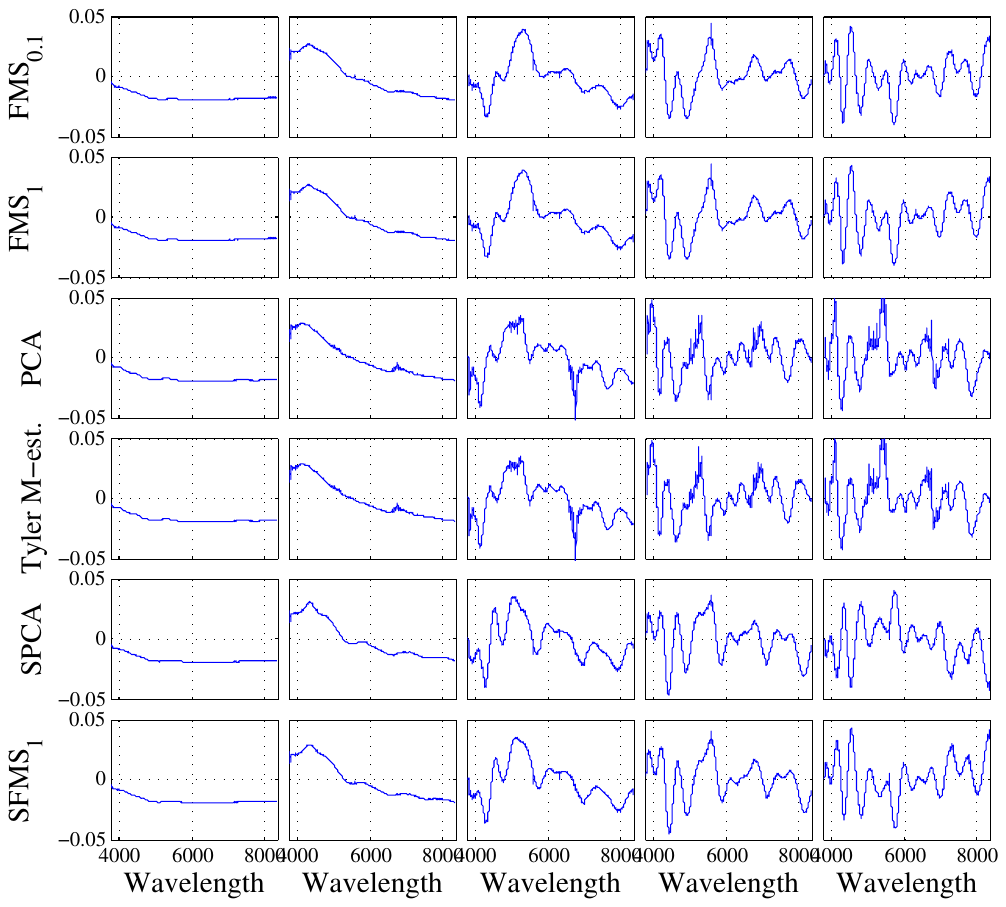}
	%\vspace{-.6cm}
	\caption{Flux vs.~wavelength (\AA) for the top 5 eigenspectra found by FMS$_p$, randomized PCA, and Tyler M-estimator. FMS$_{0.1}$, FMS$_1$, SFMS$_1$, and SPCA find eigenspectra that are not noisy.} \label{spectrafig}
	\vspace{-.1cm}
\end{figure}

Figure~\ref{spectrafig}, shows the results from running FMS$_{0.1}$, FMS$_1$, SFMS$_1$, Tyler M-estimator, PCA, and SPCA on the data. %Randomized PCA is just standard PCA using a randomized singular value decomposition for speed.
As we can see, parts of the eigenspectra in standard PCA are quite noisy, especially in the third, fourth, and fifth eigenspectra. Tyler M-estimator, although it converges in 3 iterations, makes no improvement on the eigenspectra found from standard PCA. The robustness of the FMS$_{0.1}$ and FMS$_1$ algorithms allows them to find eigenspectra that are not noisy while not sacrificing too much speed. We note here that SPCA also shows qualitatively good results, which are comparable to FMS$_p$. However, FMS$_p$, SFMS$_p$, and SPCA all have qualitatively different looking results, and this suggests that more comprehensive testing should go into seeing which method produces the best eigenspectra.

\subsubsection{Clustering Data}\label{exp:clust}

FMS$_p$ can also be used as a preprocessing step for clustering. In the previous section we examined projection to the robust subspace as a denoising technique, and in this section we demonstrate the gains denoising gives when preprocessing a data set by PCA or FMS$_p$ for $k$-means. Assume that we are given a data set and desire to partition it into $k$ clusters. If one decides to use PCA or FMS$_p$ to reduce the dimensionality of the data set, some thought must be given to what dimension of subspace to project to. The literature suggests that there is no good rule of thumb for choosing the subspace dimension $d$ without a clear model for generating the data (see e.g.~\cite{pca_based_bioinf}). In the following experiments, we show results over a range of possible values for $d$.

\begin{figure}[ht!]
	\centering
	\includegraphics[trim=100 240 100 250,clip,width=.7\textwidth]{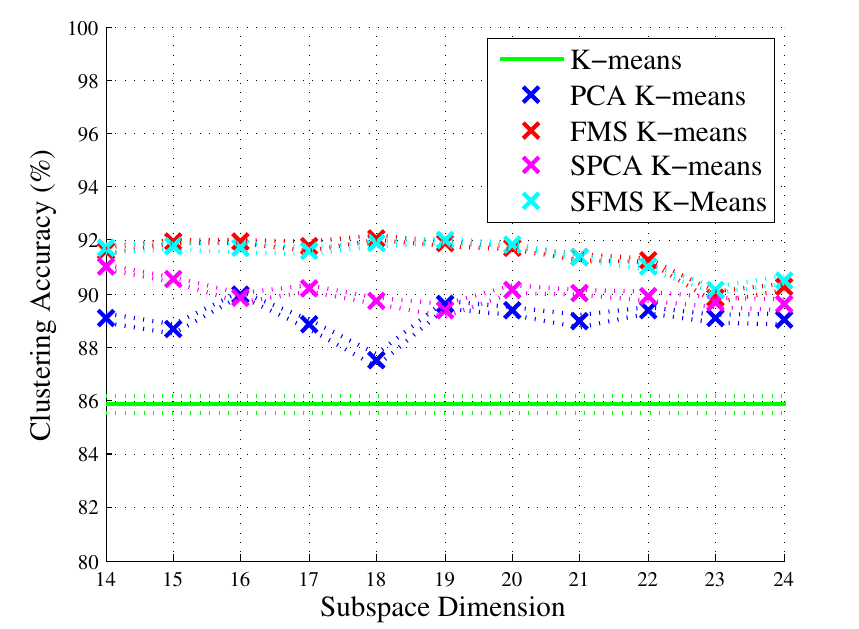}
	\caption{Clustering accuracy results for the Daily Sports and Activities data set. For this set, $N=9120$, $D=5625$, and the number of clusters is $k=19$. The results are averaged over 20 runs. Cluster accuracy is calculated as the correct number of pairwise relations between points over the total number of pairwise relations.}\label{UCIDSA}
\end{figure}

The data is taken from the "Daily Sports and Activities" data set available at \url{https://archive.ics.uci.edu/ml/data sets/Daily+and+Sports+Activities}~\cite{daily_sports_activities}, and the "Human Activity Recognition Using Smartphones" data set at \url{https://archive.ics.uci.edu/ml/data sets/Human+Activity+Recognition+Using+Smartphones}~\cite{human_act_recog_smartphone}. The "Daily Sports and Activities" data set consists of sensor data taken over a 5 second period while the subject performs a certain action. Together, there are 19 different actions, and we would like to cluster the points according to action. In total, there are 9120 data points in dimension 5625. We compare three techniques for classifying the activities: $k$-means, PCA projection$+$$k$-means, and FMS projection$+$$k$-means. By FMS here we mean FMS$_1$, since observed results were similar for FMS$_1$ and FMS$_{0.1}$. For the projection methods, we find a low dimensional subspace and project the data to that subspace before running $k$-means. For $k$-means, we use the built in MATLAB method with default parameters, which initializes using $k$ points of the data set. Clustering accuracy is measured by the number of correct pairwise relations (true positive and true negative) between points over the total number of pairwise relations. This accuracy measure is also known as the Rand index~\cite[Chapter 16]{manning2008introduction}.
%\begin{equation}
%    RI = \frac{TP + TN}{TP + TN + FP + FN}.
%    \label{}
%\end{equation}
Results are averaged over 20 runs. We display the resulting experiment in Figure~\ref{UCIDSA}, where the clustering accuracy and approximate 95\% confidence intervals (dotted lines) are given for the three methods. For this experiment, FMS is the clear choice of denoising technique for this data.

\begin{figure}[ht!]
	\centering
	\includegraphics[trim=100 240 100 250,clip,width=.7\textwidth]{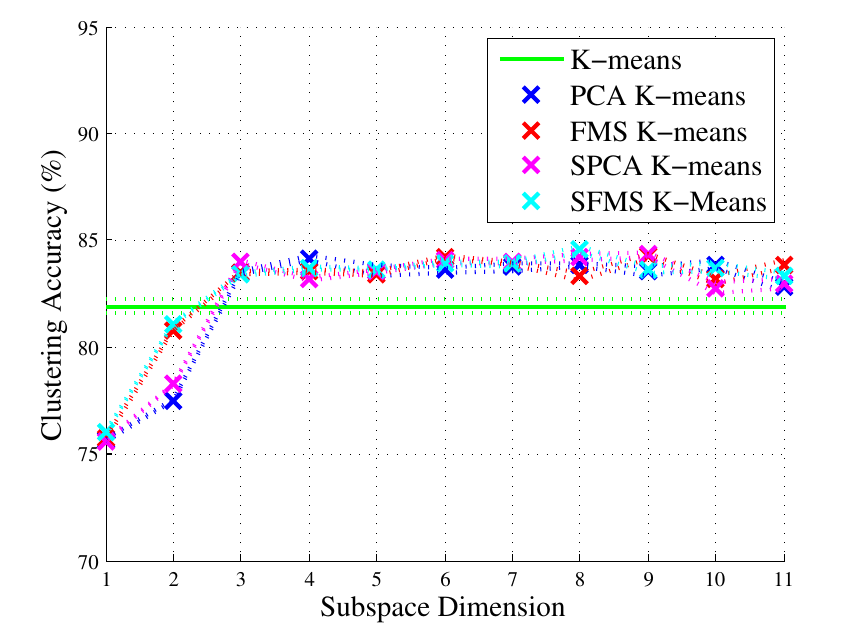}
	\caption{Clustering accuracy results for the Human Activity recognition data set. For this set, $N=7352$, $D=561$ and the number of clusters is $k=6$. The results are averaged over 20 runs. Cluster accuracy is calculated as the correct number of pairwise relations between points over the total number of pairwise relations.}\label{UCIHAR}
\end{figure}

Our second clustering data set is the training set from the "Human Activity Recognition Using Smartphones" data set, which consists of 7352 points in dimension 561. Each point consists of sensor outputs taken in a 2.56 second window from the accelerometer and gyroscope of a Samsung Galaxy S II. There are six different activities performed by each subject, and we would like to classify the data by activity. Results of the test on this data set is displayed in Figure~\ref{UCIHAR}, where again the clustering accuracy and approximate 95\% confidence intervals (dotted lines) are given for the three methods. In this experiment, denoising by PCA, SPCA, FMS, and SFMS all give comparable results.

For each of these data sets, it was not feasible to run other RSR or robust PCA algorithms. Due to the large dimension and number of points, runtimes would be very large and we ran into memory issues trying to run them in MATLAB on a personal machine. In the future, it would be ideal to run more in depth experiments to determine how other algorithms perform at such dimensionality reduction tasks. However, we believe that these experiments demonstrate a selling point for FMS$_p$. The data was able to be processed in MATLAB on a personal machine in a matter of minutes, and we are unaware of any other robust methods able to do this. It is worth noting, though, that PCA can be run in a matter of seconds and may be more efficient in some cases. While the first case shows better performance of FMS$_p$, the second case shows an example where FMS$_p$ and PCA both improve results to the same degree.

\subsubsection{Stylized Application: Faces in a Crowd}\label{exp:faces}

\begin{figure}[h!]
	\centering
	\vspace{-.7cm}
	\begin{minipage}{0.5\textwidth}
		\subfloat[\label{faces20x20}]{\includegraphics[trim=100 100 70 100,angle=90,clip,width=\textwidth]{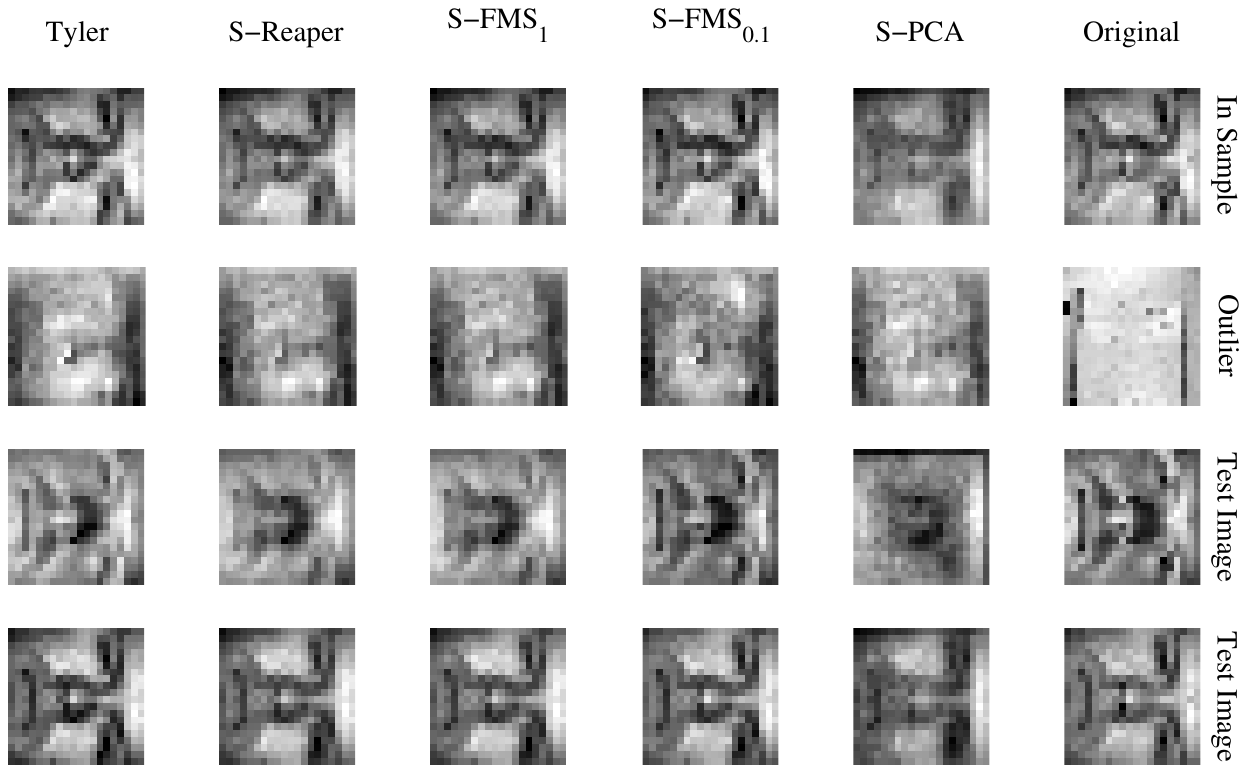}} \vfill
		\subfloat[\label{facedist}]{\includegraphics[trim=100 240 100 250,clip,width=\textwidth]{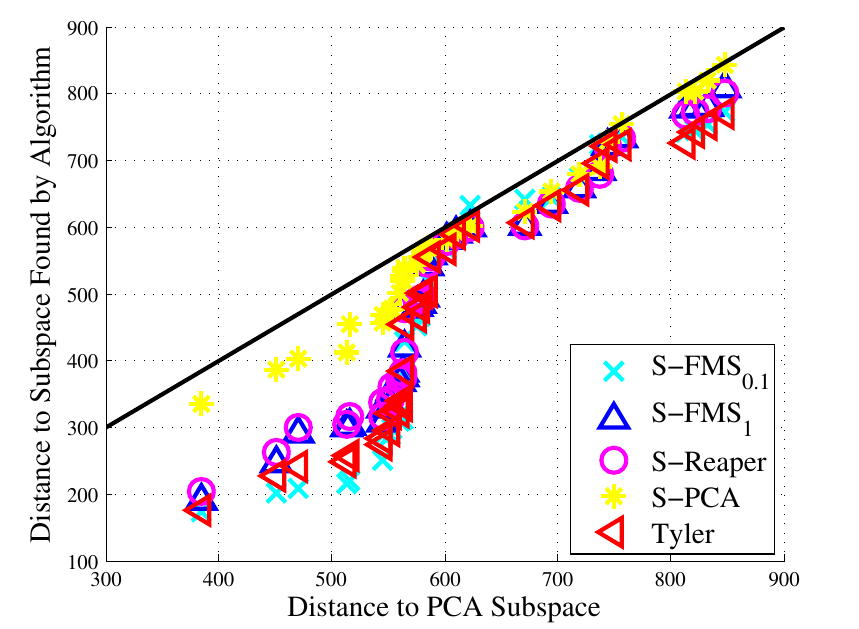}}
	\end{minipage}\hfill
	\begin{minipage}{0.5\textwidth}
		\subfloat[\label{faces30x30}]{\includegraphics[trim=100 100 70 100,angle=90,clip,width=\textwidth]{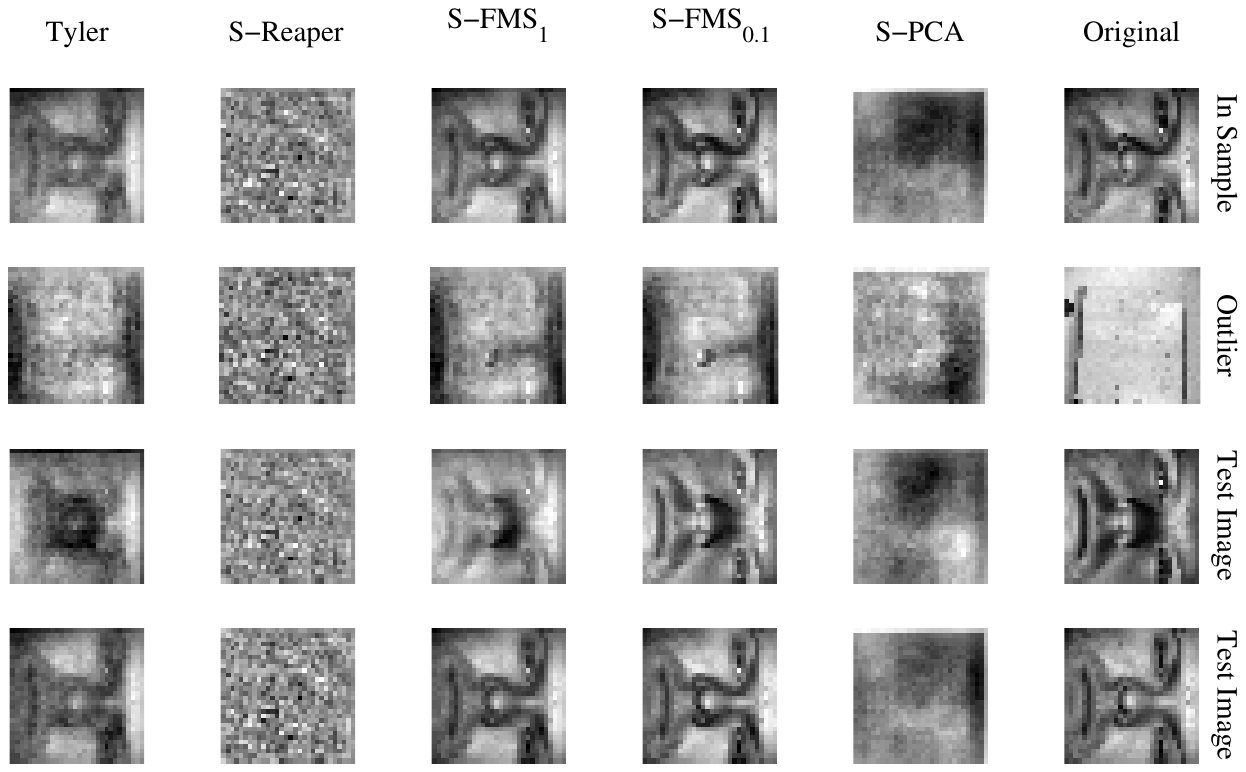}} \vfill
		\subfloat[\label{facedist2}]{\includegraphics[trim=100 240 100 250,clip,width=\textwidth]{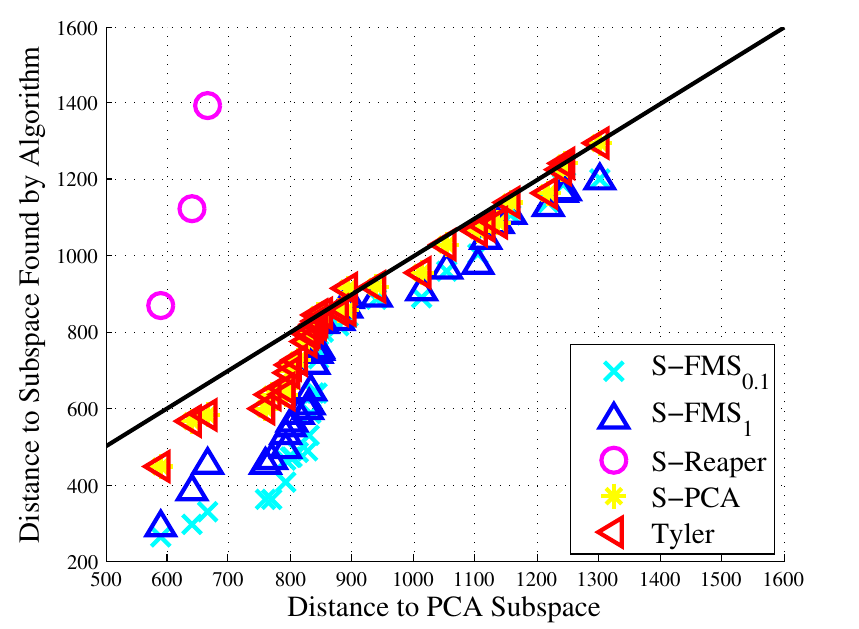}}
	\end{minipage}
	\caption{The faces in a crowd experiment for pictures downsampled to $20 \times 20$ and $30 \times 30$ dimensional pictures. Fig.~\ref{faces20x20} and Fig.~\ref{facedist} correspond to the experiment run on pictures downsampled to $20 \times 20$, and Fig.~\ref{faces30x30} and Fig.~\ref{facedist2} correspond to the experiment run on $30 \times 30$ dimensional pictures.  On the top, we show projections of the pictures onto the subspaces found by each method. On the bottom, we show the ordered distances to RSR subspace against the distance to the PCA subspace. Lower distances to the robust subspace signify a greater degree of accuracy in locating the 9-dimensional subspace in the set with outliers.} \label{faces}
	\vspace{-.3cm}
\end{figure}

The next experiment we run on real data is a stylized example from image processing. The experiment shown here is the 'Faces in a Crowd' experiment outlined by~\cite{LMTZ2014}. This experiment is motivated by the fact that images of an individual's face with fixed pose under varying lighting conditions should fall on a subspace of dimension at most 9~\cite{Basri03}. We draw a data set of 64 cropped face images from the Extended Yale Face Database~\cite{KCLee05}. 32 of these face images are sampled to be the inliers of the data set, and 400 outlier images are selected from the "BACKGROUND\_Google" folder of the Caltech 101 database. %\cite{cal101}.0

On this data, spherized algorithms tend to do better than running on the non-spherized data. Thus, we only report results of algorithms which apply such initial spherizing (after centering by the geometric median) and denote them with additional ``S-''. We remark that Tyler M-estimator implicitly spherizes the data. We fit a 9 dimensional subspace to the data set using  SPCA, SFMS$_1$, SFMS$_{0.1}$, Tyler M-estimator, and S-Reaper. Pictures are downsampled to $20 \times 20$ and $30 \times 30$ in our two tests to show performance of the algorithms on images of different dimensions.

%\begin{figure}[t!]
%\centering
%\includegraphics[trim=200 130 170 120,angle=90,clip,width=.5\textwidth]{faces20x20.pdf}
%\vspace{-.5cm}
%\caption{Projections of face images to the subspaces found in the faces in a crowd experiment on $20 \times 20$ dimensional pictures.} \label{faces20x20}
%\vspace{-.5cm}
%\end{figure}
%\begin{figure}[t!]
%\centering
%\includegraphics[trim=100 240 100 250,clip,width=.5\textwidth]{facesplot20x20_2.pdf}
%\vspace{-.3cm}
%\caption{Ordered distances to RSR subspace against the distance to the PCA subspace on the faces in a crowd data set with $20 \times 20$ dimensional pictures. Lower distances to the robust subspace signify a greater degree of accuracy in locating the 9-dimensional subspace in the set with outliers. SFMS$_{0.1}$ and Tyler appear to have the most competitive performance.} \label{facedist}
%\vspace{-.3cm}
%\end{figure}

Figure~\ref{faces20x20} demonstrates the accuracy of the found subspaces in pictures of dimension $20 \times 20$. For each subspace model, we project 4 images onto the subspace: one face from the inliers, one outlier point, and two out-of-sample faces. A better subspace should not distort the original image of the faces, and it is evident that the robust algorithms S-Reaper, Tyler, and both versions of FMS$_p$ appear to work well on this data. However, the first test image appears to be better for FMS$_{0.1}$. Another comparison of the performance of these algorithms is given in Figure~\ref{facedist}. This graph displays the ordered distances for the 32 out-of-sample faces to the robust subspaces and against their ordered distances to the PCA subspace. The $i$th point for each algorithm corresponds to the $i$th closest distance to the robust subspace against the $i$th closest distance to the PCA subspace. In general, the closer faces are to the robust subspace the better. The algorithms all appear to offer robust approximations of the underlying subspace, but SFMS$_{0.1}$ seems to have a slight edge in the lower region.

Figures~\ref{faces30x30} and~\ref{facedist2} demonstrate the same experiment, but on faces of dimension $30 \times 30$. First we note that the S-Reaper algorithm cannot locate the robust subspace in this higher dimension. SPCA and Tyler M-estimator also struggle in this scheme. However, SFMS$_{1}$ and SFMS$_{0.1}$ do quite well at finding the face subspace. In fact, looking at Figure~\ref{facedist2}, SFMS$_1$ and SFMS$_{0.1}$ outperform other algorithms by a significant degree.

%\begin{figure}[t!]
%\centering
%\includegraphics[trim=100 180 100 180,clip,width=.5\textwidth]{faces30x30.pdf}
%\vspace{-.5cm}
%\caption{Projections of face images to the subspaces found in the faces in a crowd experiment on $30 \times 30$ dimensional pictures. Out of the tested algorithms, FMS$_1$ appears to have the best performance.}\label{faces30x30}
%\vspace{-.2cm}
%\end{figure}

%\begin{figure}[ht!]
%\centering
%\includegraphics[trim=100 240 100 250,clip,width=.5\textwidth]{facesplot30x30_2.pdf}
%\vspace{-.5cm}
%\caption{Ordered distances to the subspace recovered by the algorithm against the distances to the PCA subspace on the faces in a crowd data set with $30 \times 30$ dimensional pictures. Lower distances to the robust subspace signify a greater degree of accuracy in locating the 9-dimensional subspace in the set with outliers. Most of the points from S-Reaper are too large for the scale of this graph. FMS$_1$ clearly has the best performance on this data set.} \label{facedist2}
%\vspace{-.5cm}
%\end{figure}

\subsubsection{FMS$_p$ Scales to Massive Data}
\label{sec:scalability}

To demonstrate the ability of FMS$_p$ to scale to truly large data sets, we follow the example of~\cite{hauberg2016grassmannave} and run the FMS$_p$ algorithm on a portion of the Star Wars Episode IV movie. Using $p=1$, FMS$_p$ is run on 30 minutes of Star Wars Episode IV to find a 20 dimensional robust subspace. Each point in this data set is a $720 \times 304$ RGB image, which results in a data matrix of size $54000 \times 656640$. Altogether, it took $\approx 130$ GB to store this matrix as single precision in memory. This experiment ran on two 1 TB nodes with 32 Intel Sandy Bridge processors each. FMS$_p$ was implemented in Python with \texttt{numpy} and the randomized TruncatedSVD in \texttt{sklearn}. With this set-up, the run took a total of 33 hours to complete. While there is no good choice of error metric to evaluate the found subspace here, we note that the average peak signal to noise ratio for FMS$_p$ was slightly better than that of plain PCA (20.23 vs. 20.13). However, we emphasize that the point of this experiment is to demonstrate that it is possible to run on data sets of this size: to our knowledge no other truly accurate RSR algorithm is able to do this.
%\begin{table}[h]
%	\centering
%	\begin{tabular}{l|c}
%		& PSNR$_{ave}$ \\ \hline
%		PCA	&  20.13 \\
%		FMS$_p$	&  \textbf{20.23}
%	\end{tabular} \vspace{.1cm}
%	\caption{Average PSNR on every 10th frame in the first 30 minutes of Star Wars Episode IV.}
%	\label{tab:psnr}
%\end{table}

\section{Proof of Theorems}
\label{sec:proof}

The following sections prove the Theorems presented in this paper.

\subsection{Proof of Theorem~\ref{thm:conv}}\label{sec:conv:proof}

The proof of Theorem~\ref{thm:conv} is given in the following sections. First, in~\S\ref{sec:conv:proof:weak}, we prove monotonicity and consequently convergence of $(F_{p,\delta}(L_k;\cX))_{k \in \nats}$. Next, in~\S\ref{sec:conv:proof:fixed}, we prove that the iterates $L_k$ converge to a fixed point. Finally, in~\S\ref{sec:conv:proof:stationary}, we prove that such a fixed point is necessarily a stationary point.

\subsubsection{Monotonicity and Convergence of $(F_{p,\delta}(L_k;\cX))_{k \in \nats}$}\label{sec:conv:proof:weak}
We begin with a proposition demonstrating monotonicity and consequently weak convergence of the FMS$_p$ algorithm.
\begin{proposition}\label{prop:converge:weak}
	For a fixed data set $\cX$, let $(L_k)_{k \in \nats}$ be the sequence obtained by applying FMS$_p$ without stopping and let $F_{p,\delta}$ be the function expressed in \eqref{costf2}. Then $(F_{p,\delta}(L_k;\cX))_{k \in \nats}$ is non-increasing and converges in $\reals$.
\end{proposition}

\begin{proof}[Proof of Proposition~\ref{prop:converge:weak}]
	%{\em Proof of Proposition~\ref{prop:converge:weak}.}
	For this analysis, it is useful to define a majorizing function $H_{p,\delta}$ for our cost function $F_{p,\delta}$ by:
	\begin{align}\label{majfunc:prop}
	H_{p,\delta}(L,L_0;\cX) &= \sum_{\latop{1 \leq i \leq N}{\dist^{2-p}(\bx_i,L_0) \geq p\delta}}
	\left(\frac{p}{2} \frac{\dist(\boldsymbol{x}_i,L)^2}{\dist(\boldsymbol{x}_i,L_0)^{2-p}}
	+ \left(1-\frac{p}{2}\right)\dist(\boldsymbol{x}_i,L_0)^{p} \right) +\\ \nonumber
	&\sum_{\latop{1 \leq i \leq N}{\dist^{2-p}(\bx_i,L_0) < p\delta}}
	\left(\frac{\dist^2(\bx_i,L)}{2\delta}  + (p\delta)^{p/(2-p)} - \frac{(p\delta)^{2/(2-p)}}{2\delta}\right).
	\end{align}
	
	This function is said to majorize $F_{p,\delta}$ since it satisfies the following two properties
	\begin{equation} \label{prop:hyp1}
	F_{p,\delta}(L;\cX) \leq H_{p,\delta}(L,L_0;\cX) \ \forall \  L, L_0 \in G(D,d),
	\end{equation}
	\begin{equation} \label{prop:hyp2}
	F_{p,\delta}(L_0;\cX) = H_{p,\delta}(L_0,L_0;\cX) \ \forall \ L_0 \in G(D,d).
	\end{equation}
	
	We prove these two relations in the following. The relation in \eqref{prop:hyp2} can be simply shown by evaluating $H_{p,\delta}(L_0,L_0;\cX)$ to find that $H_{p,\delta}(L_0,L_0;\cX) = F_{p,\delta}(L_0;\cX)$. For the relation in \eqref{prop:hyp1}, we will examine $H_{p,\delta}$ and $F_{p,\delta}$ term by term. Let
	\begin{align}\label{termwisecost}
	H_{p,\delta}(L,L_0;\bx_i) &=
	\begin{cases}
	\frac{p}{2} \frac{\dist(\boldsymbol{x}_i,L)^2}{\dist(\boldsymbol{x}_i,L_0)^{2-p}}
	+
	\left(1-\frac{p}{2}\right)\dist(\boldsymbol{x}_i,L_0)^{p} , & \text{if $\dist^{2-p}(\bx_i,L_0) \geq p\delta$;} \\
	\frac{\dist^2(\bx_i,L)}{2\delta}  + (p\delta)^{p/(2-p)} -
	\frac{(p\delta)^{2/(2-p)}}{2\delta}, & \text{if $\dist^{2-p}(\bx_i,L_0))
		< p\delta$,}
	\end{cases} \\
	F_{p,\delta}(L;\bx_i) &=
	\begin{cases}
	\dist^p(\bx_i,L), & \text{if $\dist^{2-p}(\bx_i,L) \geq p\delta$;} \\
	\frac{\dist^2(\bx_i,L)}{2\delta} + (p\delta)^{p/(2-p)} -
	\frac{(p\delta)^{2/(2-p)}}{2\delta}, & \text{if $\dist^{2-p}(\bx_i,L) <
		p\delta$.}
	\end{cases}
	\end{align}
	We will show that $H_{p,\delta}(L,L_0;\bx_i) \geq F_{p,\delta}(L;\bx_i) \ \forall \ \bx_i \in \cX, \ \forall \ L,L_0 \in G(D,d)$. We first choose an arbitrary $\bx_i \in \cX$. With $\beta_0$ a fixed constant, it is helpful to define two auxiliary functions $f:[0,\infty) \to \reals$ and $h:[0,\infty) \to \reals$ by
	\begin{equation}\label{convexf}
	f(z) = \frac{p \beta_0^{p-2}}{2}z^{2/p} - z + \left(1-\frac{p}{2}\right) \beta_0^p,
	\end{equation}
	\begin{align}\label{parabolah}
	h(z) = \frac{p \beta_0^{p-2}}{2}z^{2} - \frac{z^2}{2\delta} + (1-\frac{p}{2})
	\beta_0^p - \left( (p\delta)^{p/(2-p)} - \frac{(p\delta)^{2/(2-p)}}{2\delta}
	\right).
	\end{align}
	With these functions in hand, the proof of $H_{p,\delta}(L,L_0;\bx_i) \geq F_{p,\delta}(L;\bx_i)$ follows by looking at cases. It is helpful to note that $f$ is convex for $0<p<2$,
	$f(\beta_0^p) = 0$, and $f'(\beta_0^p) = 0$.
	
	First, suppose $\dist^{2-p}(\bx_i,L_0) < p\delta.$ In this case, if $\dist^{2-p}(\bx_i,L) < p\delta$, then $H_{p,\delta}(L,L_0;\bx_i) = F_{p,\delta}(L;\bx_i)$. On the other hand, if $\dist^{2-p}(\bx_i,L) \geq p\delta$, consider the function $f$ given in \eqref{convexf}. Taking $\beta_0 = (p\delta)^{1/(2-p)}$ and $\beta = \dist(\bx_i,L)$, proving that $H_{p,\delta}(L,L_0;\bx_i) \geq F_{p,\delta}(L;\bx_i)$ is then equivalent to showing that $f(\beta^p) \geq 0$, which follows from the convexity of $f$ and the fact that $f(\beta_0^p) = 0$ and $f'(\beta_0^p) = 0$.

	Next, suppose that $\dist^{2-p}(\bx_i,L_0) \geq p\delta$. Again, let $\beta = \dist(\bx_i,L)$ and now let $\beta_0 = \dist(\bx_i,L_0)$. We must further break the problem down into two more subcases. $H_{p,\delta}(L,L_0;\bx_i) \geq F_{p,\delta}(L;\bx_i)$ becomes equivalent to showing that the following two inequalities hold:
	\begin{align}\label{betacases}
	&\frac{p \beta_0^{p-2}}{2}\beta^2 + (1-\frac{p}{2}) \beta_0^p \geq \beta^p,
	& \text{if $\beta^{2-p} \geq p \delta$};\\
	&\frac{p \beta_0^{p-2}}{2}\beta^2 + (1-\frac{p}{2}) \beta_0^p \geq
	\frac{\beta^2}{2\delta}  + (p\delta)^{p/(2-p)} -
	\frac{(p\delta)^{2/(2-p)}}{2\delta}, & \text{if $\beta^{2-p} < p\delta$}.
	\end{align}
    If $\beta^{2-p} \geq p\delta$, then we consider the case in \eqref{betacases}. In this case, using \eqref{convexf}, we again have that $f(z) \geq 0$ for all $z \in [0,\infty)$, and thus $f(\beta^{p}) \geq 0$ and $H_{p,\delta}(L,L_0;\bx_i) \geq F_{p,\delta}(L;\bx_i)$ follows. If $\beta^{2-p} < p\delta$, we must consider the case in \eqref{betacases}. Recall the definition of $h$ given in \eqref{parabolah}. The function $h(z)$ is a parabola with vertex at $z=0$ that opens down since $\beta_0 \geq p\delta$. Thus the minimum of $h(z)$ in the interval $\{z:z^{2-p} < p\delta, z\geq 0 \}$ is at $z^* = (p\delta)^{1/(2-p)}$. It suffices to show that $h(z^*)\geq 0$. That is,
	\begin{align}\label{minh}
	\frac{p \beta_0^{p-2}}{2}(p\delta)^{2/(2-p)} - \frac{(p\delta)^{2/(2-p)}}{2\delta} + (1-\frac{p}{2}) \beta_0^p - \left( (p\delta)^{p/(2-p)} - \frac{(p\delta)^{2/(2-p)}}{2\delta} \right)\\ \nonumber
	=\frac{p \beta_0^{p-2}}{2}(p\delta)^{2/(2-p)}+ (1-\frac{p}{2}) \beta_0^p - (p\delta)^{p/(2-p)}   \geq 0.
	\end{align}
	We notice that the inequality \eqref{minh} is equivalent to $f((p\delta)^{p/(2-p)})\geq 0$, which follows from $f(z)\geq 0$ for all $z \in [0,\infty)$.
	
	From the previous analysis, we are able to conclude that for any $\bx_i \in \cX$
	\begin{align}
	H_{p,\delta}(L,L_0;\bx_i) \geq F_{p,\delta}(L;\bx_i) \ \forall \ L,L_0 \in G(D,d).
	\end{align}
	From this we finally obtain the relation in \eqref{prop:hyp1} as follows:
	\begin{align}
	H_{p,\delta}(L,L_0;\cX) = \sum\limits_{i=1}^{N} H_{p,\delta}(L,L_0;\bx_i) \geq \sum\limits_{i=1}^{N}F_{p,\delta}(L;\bx_i) =F_{p,\delta}(L;\cX).
	\end{align}
	
	Now consider $H_{p,\delta}(L,L_k;\cX)$ as a function of $L$. We will show that the minimization of $H_{p,\delta}(L,L_k;\cX)$ over all $L \in G(D,d)$ is simply a least squares minimization that can be solved by PCA on the data set $\cX$ scaled by \\ $\max(\dist(\boldsymbol{x}_i,L_k)^{(2-p)/2},\sqrt{p\delta})^{-1}$. Suppose that we want to use PCA to calculate a subspace $L_{k+1}$ from the data set $\bY = \{\bx_i / \max(\dist(\boldsymbol{x}_i,L_k)^{(2-p)/2}, \sqrt{p\delta}) \}_{i=1}^N = \{\by_i\}_{i=1}^N$. By the definition of PCA, $L_{k+1}$ is given by
	\begin{align} \label{eq:argminH}
	&L_{k+1} = \underset{L}{\argmin} \sum\limits_{i=1}^{N} \dist(\by_i,L)^2 = \underset{L}{\argmin} \sum\limits_{i=1}^{N} \left\|\by_i - \bP_{L}\by_i\right\|^2 \\
	\nonumber &= \underset{L}{\argmin} \sum\limits_{i=1}^{N} \left\|\frac{\bx_i}{\max(\dist(\boldsymbol{x}_i,L_k)^{(2-p)/2},\sqrt{p\delta})} - \bP_{L}\frac{\bx_i}{\max(\dist(\boldsymbol{x}_i,L_k)^{(2-p)/2},\sqrt{p\delta})}\right\|^2\\
	\nonumber &= \underset{L}{\argmin} \sum\limits_{i=1}^{N} \frac{1}{\max(\dist(\boldsymbol{x}_i,L_k)^{2-p},p\delta)} \|\bx_i - \bP_{L}\bx_i\|^2 \\
	\nonumber &= \underset{L}{\argmin} \sum_{\latop{1 \leq i \leq N}{\dist^{2-p}(\bx_i,L_k)\geq p\delta}} \frac{\dist(\bx_i,L)^2}{\dist(\bx_i,L_k)^{2-p}}
	+ \sum_{\latop{1 \leq i \leq N}{\dist^{2-p}(\bx_i,L_k)< p\delta}} \frac{\dist(\bx_i,L)^2}{p\delta}\\
	\nonumber &= \underset{L}{\argmin} \sum_{\latop{1 \leq i \leq N}{\dist^{2-p}(\bx_i,L_k)\geq p\delta}} \frac{p}{2}\frac{\dist(\bx_i,L)^2}{\dist(\bx_i,L_k)^{2-p}}
	+ \sum_{\latop{1 \leq i \leq N}{\dist^{2-p}(\bx_i,L_k)< p\delta}} \frac{\dist(\bx_i,L)^2}{2\delta}
	\\
	\nonumber &= \underset{L}{\argmin} \sum_{\latop{1 \leq i \leq N}{\dist^{2-p}(\bx_i,L_k)\geq p\delta}} \left(\frac{p}{2}\frac{\dist(\bx_i,L)^2}{\dist(\bx_i,L_k)^{2-p}} +\left(1-\frac{p}{2}\right) \dist(\bx_i,L_k)^{p}\right)
	+  \\
	\nonumber & \sum_{\latop{1 \leq i \leq N}{\dist^{2-p}(\bx_i,L_k)< p\delta}}
	\left(\frac{\dist(\bx_i,L)^2}{2\delta} + (p\delta)^{p/(2-p)} - \frac{(p\delta)^{2/(2-p)}}{2\delta}\right) \\
	\nonumber &= \underset{L}{\argmin}\ H_{p,\delta}(L,L_k;\cX).
	\end{align}
	Thus, this definition of $H_{p,\delta}$ allows us to write the iterates $L_k$ of the FMS$_p$ algorithm as
	\begin{align} \label{iterates}
	L_{k+1} = \underset{L}{\argmin} \ H_{p,\delta}(L,L_k;\cX), \ k \in \nats.
	\end{align}
	The proof of the proposition is completed by noting that \eqref{iterates} and \eqref{prop:hyp1} imply that
	\begin{align}\label{}
	F_{p,\delta}(L_{k+1};\cX) \leq H_{p,\delta}(L_{k+1},L_k;\cX) \leq H_{p,\delta}(L_k,L_k;\cX) = F_{p,\delta}(L_k;\cX).\end{align}
	Thus the sequence $(F_{p,\delta}(L_k;\cX))_{k \in \nats}$ forms a non-increasing sequence that is bounded below by $0$, and so it must converge to a point in $\reals$.
	\qed
\end{proof}

\subsubsection{Convergence of $(L_k)_{k \in \nats}$ to a Fixed Point}\label{sec:conv:proof:fixed}

The next step is to show convergence of the iterates $L_k$ to a fixed point over $G(D,d)$. Before we continue, it is useful to remind ourselves of some general results on algorithms and point-to-set maps. Our discussion closely follows the discussion given by~\citet{tight_frames},~\citet{WSL13}, and~\citet{luenberger_ye_linear_nonlinear_prog}.

First, given two general spaces $\mathcal{U},\mathcal{V}$, a point-to-set map $\mathcal{F}$ is a function $\mathcal{F}:\mathcal{U} \to \mathcal{P}(\mathcal{V})$. A point-to-set map $\mathcal{F}$ is closed at $\widehat{\bx}$ if for a sequence $\{\bx_k\} \subset \mathcal{U}$ that converges to $\widehat{\bx}$, any sequence $\{\by_k\} \subset \mathcal{V}$ such that $\by_k \in \mathcal{F}(\bx_k)$ and $\by_k \to \widehat{\by}$ gives $\widehat{\by} \in \mathcal{F}(\widehat{\bx})$. For this discussion, it suffices to consider point-to-set maps which take a general space $\mathcal{U}$ to $\mathcal{P}(\mathcal{U})$. A point $\bx$ of the map $\mathcal{F}:\mathcal{U} \to \mathcal{P}(\mathcal{U})$ is a fixed point if $\{\bx\} = \mathcal{F}(\bx)$, and $\bx$ is a generalized fixed point if $\bx \in \mathcal{F}(\bx)$. Given a point-to-set map $\mathcal{F}$, an associated {\em iterative algorithm} $\mathcal{F}^*$ generates a sequence of points by $\mathcal{F}^*(\bx_k) = \bx_{k+1} \in \mathcal{F}(\bx_k)$. If we have a cost function $F: \mathcal{U} \to [0,\infty)$, the algorithm is said to be monotonic with respect to $F$ if $\by \in \mathcal{F}(\bx)$ implies that $F(\by) \leq F(\bx)$, and strictly monotonic if the equality only holds when $\by = \bx$. With this in mind, we are able to use two useful theorems on the convergence of monotonic algorithms.

\begin{theorem}[\citet{Zangwill69}]\label{thm_mono} Let $\mathcal{F}:\mathcal{U} \to \mathcal{P}(\mathcal{U})$ be a point-to-set map with an associated algorithm $\mathcal{F}^*:\mathcal{U} \to \mathcal{U}$ that is monotonic with respect to $F$. Suppose also that given an initial point $\bx_0$, $\mathcal{F}^*$ generates a sequence $\{\bx_k\}$ which lies in a compact set. Then, the sequence has at least one accumulation point $\widehat{\bx}$ and $F(\widehat{\bx})=\lim F(\bx_k)$. Moreover if $\mathcal{F}$ is closed at $\widehat{\bx}$ then $\bx$ is a generalized fixed point of $\mathcal{F}$.
\end{theorem}

\begin{theorem}[\citet{Meyer76}]\label{thm_strictmono} Let $\mathcal{F}:\mathcal{U} \to \mathcal{P}(\mathcal{U})$ be a point-to-set map with an associated algorithm $\mathcal{F}^*:\mathcal{U} \to \mathcal{U}$ that is strictly monotonic with respect to $F$, which generates a sequence $\{\bx_k\}$ that lies in a compact set. If $\mathcal{F}$ is closed at an accumulation point $\widehat{\bx}$, then $\widehat{\bx}$ is a fixed point of $\mathcal{F}$. If $\mathcal{U}$ is a metric space with metric $d(\cdot,\cdot)$, then $d(\bx_{k+1},\bx_k) \to 0$. It follows then that $\{\bx_k\}$ converges to $\widehat{\bx}$ or that the accumulation points of $\{\bx_k\}$ form a continuum.
\end{theorem}

Using the results of these two theorems, we are able to prove the following proposition, which establishes convergence of FMS$_p$ to a fixed point.
\begin{proposition}\label{prop:fixedpoint}
	The sequence $(L_k)_{k=1}^\infty$ generated by the FMS$_p$ algorithm converges to a fixed point or a continuum of fixed points in $G(D,d)$.
\end{proposition}
\begin{proof}[Proof of Proposition~\ref{prop:fixedpoint}]
	For the given data set $\{\bx_i\}_{i=1}^N$, let $L_k \in G(D,d)$ be a sequence of iterates obtained by applying FMS$_p$. We can define an equivalence relation on $G(D,d)$ which declares subspaces as equivalent if they yield the same FMS$_p$ iteration: $L_1 \sim L_2 \iff H_{p,\delta}(L,L_1;\cX) = H_{p,\delta}(L,L_2;\cX) \ \forall \ L \in G(D,d)$. Specifically, for a given data set $\cX = \{\bx_i\}_{i=1}^N$, this equivalence relation $\sim$ can be defined on $G(D,d)$ by $L_1 \sim L_2$ if, for all $1\leq i \leq N$, either $\dist(\bx_i,L_1) = \dist(\bx_i,L_2)$ or $\dist(\bx_i,L_1)< p\delta$ and $\dist(\bx_i,L_2)< p\delta$. The corresponding quotient space is then defined as
	\begin{equation}\label{Gequiv}
	\widetilde{G} = G(D,d)/\sim.
	\end{equation}
	For each $L_k$ let $\widetilde{L}_k$ denote its equivalence class in $\widetilde{G}$.
	
	If $L_{k+1} \not \sim L_k$, there are three cases to consider.  In each case, we demonstrate strict monotonicity of the sequence $\widetilde{L}_k$ with respect to $F_{p,\delta}$. In the proof of Proposition~\ref{prop:converge:weak}, we showed that the termwise inequality $F_{p,\delta}(\bx_i,L_{k+1}) \leq H_{p,\delta}(\bx_i,L_{k+1},L_k)$ held. Thus, for the strict inequality $F_{p,\delta}(L_{k+1};\cX)<F_{p,\delta}(L_k;\cX)$, it suffices to show that there is a strict inequality for one term $F_{p,\delta}(L_{k+1};\bx_i) < H_{p,\delta}(L_{k+1},L_k;\bx_i)$.
	
	First suppose that for some index $j$, $\dist^{2-p}(\bx_j,L_k) < p \delta$ and $\dist^{2-p}(\bx_j,L_{k+1}) \geq p\delta$. We recall the function $f$ defined in \eqref{convexf}, while this time letting $\beta_0 = \dist(\bx_j,L_{k})$ and $\beta = \dist(\bx_j,L_{k+1})$. We see that $f(\beta^p) = 0$ only if $\beta^p=\beta_0^p$, which in this case means that $F_{p,\delta}(L_{k+1};\bx_j) = H_{p,\delta}(L_{k+1},L_k;\bx_j)$ only if $\dist(\bx_j,L_{k+1}) = \dist(\bx_j,L_k)$. Thus, since $\dist(\bx_j,L_{k+1}) \neq \dist(\bx_j,L_k)$, we must have that $F_{p,\delta}(L_{k+1};\bx_j) < H_{p,\delta}(L_{k+1},L_k;\bx_j)$ and $F_{p,\delta}(L_{k+1};\cX)<F_{p,\delta}(L_k;\cX)$.
	
	Next, suppose that for some index $j$, $\dist^{2-p}(\bx_j,L_k) \geq p \delta$, and $\dist(\bx_j,L_{k+1}) \neq \dist(\bx_j,L_k)$. If $\dist^{2-p}(\bx_j,L_{k+1})< p \delta$, then note that the function defined in~\eqref{parabolah} is a parabola that opens downwards, with infimum on the interval $\{z:z^{2-p}<p\delta,z \geq 0\}$ at $z^* = (p\delta)^{1/(2-p)}$. From the previous proof, $h(z) \geq 0$ on this interval, and $h(z) = 0$ can only be zero at the infimum $z^*$. Noting that
	\begin{align}
	h(z^*) &= \frac{p \beta_0^{p-2}}{2}(p\delta)^{2/(2-p)}+ (1-\frac{p}{2}) \beta_0^p - (p\delta)^{p/(2-p)}=f((p\delta)^{p/(2-p)}),
	\end{align}
	we find that $h(z^*)$ is zero only if $\beta_0^p = (p\delta)^{p/(2-p)}$, or $\beta_0^{2-p} = p\delta$. But, taking $\beta = z^* = (p\delta)^{1/(2-p)}$, this corresponds to the case that $\beta^{2-p} = p\delta = \beta_0^{2-p}$. Thus, $h(z)>0$ when $\beta \neq \beta_0$, which then gives that $F_{p,\delta}(L_{k+1};\bx_j) < H_{p,\delta}(L_{k+1},L_k;\bx_j)$ and $F_{p,\delta}(L_{k+1};\cX)<F_{p,\delta}(L_k;\cX)$. On the other hand, if $\dist^{2-p}(\bx_j,L_{k+1})\geq p \delta$, then the function $f$ given by \eqref{convexf} satisfies $f(z) = 0$ if and only if $z=\beta_0^p$. In other words, $f(\beta^p) = 0$ only if $\beta = \beta_0$. Since $\beta \neq \beta_0$, we must have that $f(\beta^p)>0$ and thus $F_{p,\delta}(L_{k+1};\cX) < H_{p,\delta}(L_{k+1},L_k;\cX) \leq F_{p,\delta}(L_k;\cX)$.
	
    Now suppose that $L_{k+1} \sim L_k$. It is apparent that $H_{p,\delta}(L,L_{k+1};\cX) = H_{p,\delta}(L,L_k;\cX)$ and \eqref{eq:argminH} imply that $L_{k+1} = \argmin_{L \in G(D,d)} H_{p,\delta}(L,L_{k+1};\cX)$. This then implies that $L_{k+1}$ is a fixed point. In the case that there are more than one solution to the minimization $H_{p,\delta}(L,L_{k+1};\cX)$, this motivates an additional stopping condition for FMS$_p$ by: ``Stop if consecutive iterates belong to the same equivalence class''. In practice, we find that checking this condition is not needed, and therefore do not include it in Algorithm~\ref{algorithm:FMS}.
	
	We have found that if $L_{k+1} \not \sim L_k$, then $F_{p,\delta}(L_{k+1};\cX) < F_{p,\delta}(L_k;\cX)$, and if $L_{k+1} \sim L_k$, then $L_{k+1}$ is a fixed point. That is, the sequence of iterates generated over $G(D,d)$ is either strictly monotonic or FMS$_p$ converges to a fixed point in a finite number of iterations.
	
    Let us finally consider the case that FMS$_p$ does not converge to a fixed point in a finite number of iterations. Since $H_{p,\delta}(L,L_0;\cX)$ is continuous as a function of $L$, the infimal map $M: G(D,d) \to G(D,d)$ given by $M(L_0) = \argmin_{L \in G(D,d)} H_{p,\delta}(L,L_0;\cX)$ is closed~\cite{Dantzig67}. Therefore, by Theorem~\ref{thm_mono}, the sequence generated by FMS$_p$ over $G(D,d)$ has an accumulation point $L^*$. Strict monotonicity over $G(D,d)$ implies that this accumulation point is in fact a fixed point by Theorem~\ref{thm_strictmono}. The Grassmannian $G(D,d)$ is a compact metric space, and so following the result of Theorem~\ref{thm_strictmono}, we get that $\dist(L_{k+1},L_{k}) \to 0$. Consequently, FMS$_p$ generates a sequence $L_k$ over $G(D,d)$ which converges to fixed point or a continuum of fixed points.
	\qed
\end{proof}
%We have found that if $L_{k+1} \not \sim L_k$, then $F_{p,\delta}(L_{k+1}) < F_{p,\delta}(L_k)$. That is, the sequence of iterates generated over $\widetilde{G}$ is strictly monotonic. Further, since $H_{p,\delta}(L,L_0)$ is continuous as a function of $L$, the infimal map $M: G(D,d) \to G(D,d)$ given by $M(L_0) = \argmin_L H_{p,\delta}(L,L_0)$ is closed~\cite{Dantzig67}. Therefore, by Theorem~\ref{thm_mono}, the sequence generated by FMS$_p$ over $\widetilde{G}$ has an accumulation point $\widetilde{L}^*$. Strict monotonicity over $\widetilde{G}$ implies that this accumulation point is in fact a fixed point by Theorem~\ref{thm_strictmono}. $G(D,d)$ is also a metric space, whose metric $\dist(L_1,L_2)$ then descends to a metric on the quotient space $\widetilde{G}$ given by $\widetilde{\dist}(\widetilde{L}_1,\widetilde{L}_2) = \inf\{\dist(L_1,L_2):L_1 \in \widetilde{L}_1,L_2 \in \widetilde{L}_2\}$. Following the result of Theorem~\ref{thm_mono}, we get that $\dist(\widetilde{L}_1,\widetilde{L}_2) \to 0$. Consequently, FMS$_p$ generates a sequence $\widetilde{L}_k$ over $\widetilde{G}$ which converges to fixed point or a continuum of fixed points.
%\qed

\subsubsection{The Fixed Point $L^*$ is a Stationary Point}\label{sec:conv:proof:stationary}
We finish the proof of Theorem~\ref{thm:conv} by showing that any FMS$_p$ fixed point $L^*$ is a stationary point of the cost function $F_{p,\delta}$. Since $L^*$ is a fixed point, we know that $L^* = \argmin_{L \in G(D,d)} H_{p,\delta}(L,L^*;\cX)$. Let $L_1 \in B(L^*,1)$ be arbitrary, and parametrize a geodesic between $L^*$ and $L_1$ by $L(t)$, where $L(0) = L^*$ and $L(1) = L_1$. The facts that $L^*$ is a fixed point and that the derivative $\frac{d}{dt}H_{p,\delta}(L(t),L^*;\cX)$ exists give
\begin{equation}
\frac{d}{dt}H_{p,\delta}(L(t),L^*;\cX) \Bigg{|}_{t=0} = 0.
\end{equation}
Examining $F_{p,\delta}$ and $H_{p,\delta}$ termwise, it is readily apparent that
\begin{equation}
\frac{d}{dt}F_{p,\delta}(L(t);\bx_i) \Bigg{|}_{t=0} =  \frac{d}{dt}H_{p,\delta}(L(t),L^*;\bx_i) \Bigg{|}_{t=0}.
\end{equation}
Thus, we conclude that
\begin{align}
\frac{d}{dt}F_{p,\delta}(L(t);\cX) \Bigg{|}_{t=0} &= \sum_{i=1}^N \frac{d}{dt}F_{p,\delta}(L(t);\bx_i) \Bigg{|}_{t=0} = \sum_{i=1}^N \frac{d}{dt}H_{p,\delta}(L(t),L^*;\bx_i) \Bigg{|}_{t=0}\\
\nonumber &= \frac{d}{dt}H_{p,\delta}(L(t),L^*;\cX) \Bigg{|}_{t=0} = 0.
\end{align}
Since the point $L_1$ was arbitrary, $L^*$ must be a stationary point of $F_{p,\delta}$ over $G(D,d)$. This concludes the proof of Theorem~\ref{thm:conv}.
\qed

\subsection{Proof of Theorems~\ref{thm:globconv} and~\ref{thm:globconv:stab}}\label{sec:globconv:proof}

The proof of Theorems~\ref{thm:globconv} and~\ref{thm:globconv:stab} proceed in the following sections. First, we prove some preliminary lemmas in \S\ref{sec:globconv:proof:prelim}. Next, in \S\ref{sec:globconv:proof:deriv} we prove a proposition that gives probabilistic estimates on where the stationary points of $F_{p,\delta}$ occur when $\cX$ is sampled from~\eqref{mixtmeas1} when $K=1$. Then, we finish the proof of Theorem~\ref{thm:globconv} in \S\ref{sec:globconv:proof:fin}. Next, \S\ref{sec:globconv:proof:stab} gives the proof of Theorem~\ref{thm:globconv:stab}. Finally,~\S\ref{sec:globconv:proof:parameters} gives bounds on some constants used to prove Theorems~\ref{thm:globconv} and~\ref{thm:globconv:stab}

\subsubsection{The Limiting Stationary Behavior of $F_{p,\delta}$}
\label{sec:globconv:proof:prelim}

We begin with some notation, and then proceed to prove two lemmas. For a fixed subspace $\dot{L}$, we can parametrize a geodesic on the Grassmannian by $L(t):[0,1] \to G(D,d)$ from $\dot{L}$ to $\widehat{L} \in B(\dot{L},1)$ by
\begin{equation}
L(t)  = \Sp( \{ \cos(t\theta_j) \bv_j + \sin(t\theta_j) \bu_j \}_{j=1}^d),
\label{eq:grassgeodesic}
\end{equation}
where $\{\theta_j\}_{j=1}^d$ are the principal angles between $\dot{L}$ and $\hat{L}$, $\{\bv_j\}_{j=1}^d$ is a basis for $\dot{L}$, and $\{\bu_j\}_{j=1}^d$ is a complementary orthogonal system for $\widehat{L}$. For a more detailed discussion on the construction of this geodesic, see~\cite[\S3.2.1]{lp_recovery_part1_11}. For all arguments in this paper, we assume that the interaction dimension between $L(0)$ and $L(1)$ is greater or equal to one, which means that $\theta_1 > 0$. When $\theta_1=0$, the geodesic is trivial since $L(t) = L(0) = L(1)$ and consequently the proof becomes trivial.

We also consider an asymptotic limit of the cost $F_{p,\delta}$ given in~\eqref{costf2} when the data set $\cX$ is sampled i.i.d.~from the mixture measure~\eqref{mixtmeas1} with $K=1$. For a given subspace $L$, let $\mathcal{U}_{L,p,\delta} \subset \sphere^{D-1}$ denote the set of points
\begin{equation}\label{eq:ULpdelta}
\mathcal{U}_{L,p,\delta} = \{\bx \in S^{D-1}: \dist^{2-p}(\bx,L) < p\delta\}.
\end{equation}
Then we define the asymptotic cost function $F_{p,\delta}^*$ for $0<p<2$ by
\begin{align}
F_{p,\delta}^*(L;\mu) &= \int_{\sphere^{D-1} \setminus \mathcal{U}_{L,p,\delta}} \dist^p(\bx,L) d\mu(\bx) + \\ \nonumber &\int_{\mathcal{U}_{L,p,\delta}} \frac{\dist^2(\bx,L)}{2\delta} + (p\delta)^{p/(2-p)} - \frac{(p\delta)^{2/(2-p)}}{2\delta} d\mu(\bx).
\label{}
\end{align}
It is readily apparent that $F_{p,\delta}(L;\cX)/N \overset{\text{a.s.}}{\to} F_{p,\delta}^*(L;\mu)$. On the other hand, the PCA energy $F_{2,\delta}(L;\cX) / N$ converges almost surely to its asymptotic cost
\begin{equation}
F_{2,\delta}^*(L;\mu) = \int_{\sphere^{D-1} } \dist^2(\bx,L) d\mu(\bx).
\label{}
\end{equation}

Lemma~\ref{lemma:deriv} gives formulas for the directional derivatives of $F_{p,\delta}$ and $F_{p,\delta}^*$ along the geodesic $L(t)$ given in~\eqref{eq:grassgeodesic}.
\begin{lemma}
	The derivatives of $F_{p,\delta}^*$ and $F_{p,\delta}$ for $0 < p < 2$  have the following forms:
	\begin{align}
	\frac{d}{dt} F_{p,\delta}^*(L(t);\mu) \Big|_{t=0} = \int_{\sphere^{D-1}} -p \frac{\sum_{j=1}^d \theta_j (\bv_j \cdot \bx)(\bu_j \cdot \bx)}{\max(\dist^{2-p}(\bx,L(0)),p\delta)} d\mu,
	\label{eq:asymderiv}
	\end{align}
	\begin{equation}
	\frac{d}{dt} F_{p,\delta}(L(t);\cX)\Big|_{t=0} = \sum_{i=1}^N -p \frac{\sum_{j=1}^d \theta_j (\bv_j \cdot \bx_i)(\bu_j \cdot \bx_i)}{\max(\dist^{2-p}(\bx_i,L(0)),p\delta)}.
	\label{eq:discrderiv}
	\end{equation}
	For $p=2$, $F_{2,\delta}^*$ and $F_{2,\delta}$ have the forms
	\begin{align}
	\frac{d}{dt} F_{2,\delta}^*(L(t);\mu) \Big|_{t=0} = \int_{\sphere^{D-1}} -2 \sum_{j=1}^d \theta_j (\bv_j \cdot \bx)(\bu_j \cdot \bx) d\mu,
	\label{eq:asymderivpca}
	\end{align}
	\begin{equation}
	\frac{d}{dt} F_{2,\delta}(L(t);\cX)\Big|_{t=0} = \sum_{i=1}^N -2 \sum_{j=1}^d \theta_j (\bv_j \cdot \bx_i)(\bu_j \cdot \bx_i).
	\label{eq:discrderivpca}
	\end{equation}
	\label{lemma:deriv}
\end{lemma}
\begin{proof}[Proof of Lemma~\ref{lemma:deriv}]
	%{\em Proof of Lemma~\ref{lemma:deriv}.}
	The proof of this lemma borrows from the derivations done in \S 3.2.2 of~\cite{lp_recovery_part1_11}. For a given geodesic $L(t)$, the directional derivative of the distance function is (provided $\bx \not \in L(0)$):
	\begin{equation}
	\frac{d}{dt} \dist(\bx,L(t)) \Big|_{t=0} = - \frac{\sum_{j=1}^d \theta_j (\bv_j \cdot \bx)(\bu_j \cdot \bx)}{\dist(\bx,L(0))}.
	\label{}
	\end{equation}
	In the regularized cost, when $\dist(\bx,L(0)) < \delta$, we instead calculate
	\begin{equation}
	\frac{d}{dt}\left( \frac{\dist^2(\bx,L(t))}{2\delta} +  (p\delta)^{p/(2-p)} - \frac{(p\delta)^{2/(2-p)}}{2\delta} \right)\Big|_{t=0} = - \frac{\sum_{j=1}^d \theta_j (\bv_j \cdot \bx)(\bu_j \cdot \bx)}{\delta}.
	\label{eq:regtermderiv}
	\end{equation}
	Thus, we can derive the derivative expressions for the cost functions $F_{p,\delta},F_{p,\delta}^*$ as
	\begin{align}
	\frac{d}{dt} F_{p,\delta}^*(L(t);\mu) \Big|_{t=0} &= \frac{d}{dt}  \left( \int_{\sphere^{D-1} \setminus \mathcal{U}_{L(t),\delta}} \dist^p(\bx,L(t)) d\mu \right) \Bigg|_{t=0}\\ \nonumber
	&\frac{d}{dt}\left( \int_{\mathcal{U}_{L(t),\delta}} \frac{\dist^2(\bx,L(t))}{2\delta}  +  (p\delta)^{p/(2-p)} - \frac{(p\delta)^{2/(2-p)}}{2\delta} d\mu \right) \Bigg|_{t=0} \\ \nonumber
	&= \int_{\sphere^{D-1}} -p \frac{\sum_{j=1}^d \theta_j (\bv_j \cdot \bx)(\bu_j \cdot \bx)}{\max(\dist^{2-p}(\bx_i,L(0)),p\delta)} d\mu,
	\end{align}
	\begin{equation}
	\frac{d}{dt} F_{p,\delta}(L(t);\cX)\Big|_{t=0} = \sum_{i=1}^N -p \frac{\sum_{j=1}^d \theta_j (\bv_j \cdot \bx)(\bu_j \cdot \bx)}{\max(\dist^{2-p}(\bx_i,L(0)),p\delta)}.
	\label{}
	\end{equation}
	Finally, when $p=2$, we can directly apply the derivative formula of $\dist^2(\bx,L(t))$ with respect to $t$ (seen in~\eqref{eq:regtermderiv}) to find the derivatives of $F_{2,\delta}^*$ and $F_{2,\delta}$ have the forms~\eqref{eq:asymderivpca} and~\eqref{eq:discrderivpca}.
	\qed
\end{proof}

Let $Z_{F_{p,\delta}^*}$ denote the set of stationary points of the energy $F_{p,\delta}^*$. These are precisely points on $G(D,d)$ at which all geodesic directional derivatives are zero. For the noiseless mixture measure~\eqref{mixtmeas1} with $K=1$, this set is precisely
\begin{equation} \label{eq:asymderivzeros}
Z_{F_{p,\delta}^*} = \{L \in G(D,d) : L = \Sp(\bv_1,...,\bv_d), \bv_j \in L_1^* \text{ or } \bv_j \in L_1^{*\perp},\ j=1,...,d \}.
\end{equation}
This is proved in Lemma~\ref{lemma:statpoint} below. We notice that in the set $Z_{F_{p,\delta}^*}$, $L_1^*$ is the global minimum, $L \subset L_1^{*\perp}$ are the global maxima, and any $L \in Z_{F_{p,\delta}^*}$ that contains basis vectors in both $L_1^*$ and $L_1^{*\perp}$ is a saddle point.

\begin{lemma}
	When $K=1$ in~\eqref{mixtmeas1}, for all $0 < p \leq 2$, the stationary points of $F_{p,\delta}^*(L;\mu)$ are $Z_{F_{p,\delta}^*}$ defined in~\eqref{eq:asymderivzeros}.
	\label{lemma:statpoint}
\end{lemma}

\begin{proof}[Proof of Lemma~\ref{lemma:statpoint}]
	%{\em Proof of Lemma~\ref{lemma:statpoint}.}
    We first show that $L_1^*$ and $L \subset L_1^{*\perp}$ are stationary points. The cost $F_{p,\delta}^*(L;\mu_0)$ is constant with respect to $L$. When $0<p<2$, the application of this observation in~\eqref{eq:asymderiv} leads to the simplification
	\begin{align}
	\label{eq:derivsimp}
	\frac{d}{dt} F_{p,\delta}^*(L(t);\mu)\big|_{t=0} &= \int_{\sphere^{D-1}} -p \frac{\sum_{j=1}^d \theta_j (\bv_j \cdot \bx)(\bu_j \cdot \bx)}{\max(\dist^{2-p}(\bx,L(0)),p\delta)} d\mu \\ \nonumber
	&= \int_{L_1^*}  -p \frac{\sum_{j=1}^d \theta_j (\bv_j \cdot \bx)(\bu_j \cdot \bx)}{\max(\dist^{2-p}(\bx,L(0)),p\delta)} \alpha_1 d \mu_1.
	\end{align}
	A similar result holds for $p=2$. For $L(0) = L_1^*$ and $L(1) \in B(L_1^*,1)$, $\bu_j \in L_1^{*\perp}$ for all $j$. Thus, the expression~\eqref{eq:derivsimp} is 0 by the orthogonality of $\bu_j$ and $L_1^*$. On the other hand, if $L(0) \subseteq L_1^{*\perp}$, then $\bv_j$ is orthogonal to $L_1^*$ for all $j$, and the expression~\eqref{eq:derivsimp} is 0. The same argument can be used to show that $\frac{d}{dt} F_{2,\delta}^*(L(t);\mu)\big|_{t=0}$ is zero in these cases.
	
	The next case to consider is a subspace which has basis vectors in both $L_1^*$ and $L_1^{*\perp}$. Let $L = \Sp( \bv_1,...,\bv_k,\bv_{k+1},...,\bv_d)$, where $\bv_1,...,\bv_k \in L_1^*$ and $\bv_{k+1},...,\bv_d \in L_1^{*\perp}$. Again, we first restrict to the case $0<p<2$. The derivative formula~\eqref{eq:derivsimp} of $F_{p,\delta}^*$ can be rewritten  as
	\begin{equation}\label{eq:asymFMSderiv2}
	\frac{d}{dt} F_{p,\delta}^*(L(t);\mu)\big|_{t=0} = -p \sum_{j=1}^k \theta_j \int_{L_1^*} \frac{ (\bv_j \cdot \bx)(\bu_j \cdot \bx)}{\max(\dist^{2-p}(\bx,L(0)),p\delta)} \alpha_1 d \mu_1.
	\end{equation}
	All the terms in the sum corresponding to $\bv_{k+1},...,\bv_d$ are zero due to orthogonality with $L_1^*$. Consider a single integral corresponding to an index $1 \leq l \leq k$ within the sum over $j$. For the $l^{th}$ term, if $\bu_l \in L_1^*$ or $\bu_l \in L_1^{*\perp}$, then the integral is 0 by symmetry or orthogonality respectively. On the other hand, if $\bu_l$ lies in between $L_1^*$ and $L_1^{*\perp}$, we can write $\bu_l = c_* \bw_l + c_\perp \bw_l^\perp$, where $\bw_l \in L_1^*$ and $\bw_l^\perp \in L_1^{*\perp}$. We notice that necessarily $\bw_l$ is orthogonal to $\bv_j$ for $1 \leq j \leq k$. Thus, the $l^{th}$ term in the derivative of $F_{p,\delta}^*$ reduces to
	\begin{equation}
	\int_{L_1^*} \frac{ (\bv_l \cdot \bx)(\bu_l \cdot \bx)}{\max(\dist^{2-p}(\bx,L(0)),p\delta)} \alpha_1 d \mu_1 = \alpha_1 \int_{L_1^*} \frac{ c_* (\bv_l \cdot \bx)(\bw_l \cdot \bx) }{\max\left(\left( \sum_{j=1}^k (\bu_j \cdot \bx)^2 \right)^{2-p},p\delta\right)}  d \mu_1.
	\end{equation}
	The last integral is zero  by symmetry of the integral over $L_1^*$. Again, the same logic can be applied to the case $p=2$.
	
    It remains to show that the right hand side of~\eqref{eq:asymderivzeros} contains all stationary points of $F_{p,\delta}^*$. Consider $L(0) \not \in Z_{F_{p,\delta}^*}$ . The derivative of $F_{p,\delta}^*$ is negative in the direction of $L_1^*$ and positive in the direction of $L(1) \in L_1^{*\perp}$ due to the positive measure on $L_1^*$. This can be seen from the following logic. First, assume $L(1) = L_1^*$, and we can assume that $0<\theta_j < \pi /2$ for $j=1,\dots,k$ for some $k$. Then we can write the integral in~\eqref{eq:derivsimp} as
    \begin{align}\label{eq:asymderivsaddle}
	\frac{d}{dt} F_{p,\delta}^*(L(t);\mu)\big|_{t=0} =  -\alpha_1 \sum_{j=1}^k \theta_j \int_{L_1^*}  p \frac{(\bv_j \cdot \bx)(\bu_j \cdot \bx)}{\max(\dist^{2-p}(\bx,L(0)),p\delta)} d \mu_1.
	\end{align}
    Notice that in the inner integral, since $L_1^* \cap \Sp(\bv_j,\bu_j)$ has angle less than $\pi/2$ to both $\bv_j$ and $\bu_j$, $(\bv_j \cdot \bx)$ and $(\bu_j \cdot \bx)$ have the same sign for all $\bx \in L_1^*$. Thus, the integral inside the sum in the right hand side of~\eqref{eq:asymderivsaddle} is strictly positive and the overall derivative is negative. A similar argument can be used to show that the derivative is positive from $L(0)$ in the direction of $L_1^{*\perp}$. Again, the same argument can be used for $F_{2,\delta}^*$.
	\qed
\end{proof}

%%%%%%%%%%%%%%%%%%%%%%%%%%%%%%%%%%%%%%%%%%%%%%%%%%%%%%%%%%%%%%%%%%%%%%%%%%%%%%%%%%%%%%%%%%%%%%%%%%%%%%%%%%%%%%%%

\subsubsection{The Non-Stationarity of $F_{p,\delta}$ in a Large Region}
\label{sec:globconv:proof:deriv}

We will continue the proof of Theorem~\ref{thm:globconv} with a proposition. For a given $0<\eta\leq\pi/6$, we define the set
\begin{equation}\label{eq:subspaceeta}
\cL_\eta = G(D,d) \setminus B\left(Z_{F_{p,\delta}^*},\eta\right).
\end{equation}
In other words, $\mathcal{L}_\eta$ is the set of all subspaces in $G(D,d)$ which cannot be spanned by vectors in $B(L_1^*,\eta) \cup B(L_1^{*\perp},\eta)$.
\begin{proposition}
	There exists a function $R_p : \mathcal{L}_{\eta} \to (0,\infty)$ such that for any $\dot{L} \in \mathcal{L}_{\eta}$, $B(\dot{L},R_p(\dot{L}))$ contains no stationary points of $F_{p,\delta}$ w.o.p.
	\label{prop:derivative}
\end{proposition}

\begin{proof}[Proof of Proposition~\ref{prop:derivative}]
	%{\em Proof of Proposition~\ref{prop:derivative}.}
	
	In the first part of the proof, we will show that the derivative of $F_{p,\delta}$ is bounded away from zero on $\mathcal{L}_\eta$ w.o.p.~For each $\dot{L} \in \mathcal{L}_{\eta}$, let $\dot{L}(t)$ be the geodesic from $\dot{L}$ towards $L_1^*$. In~\eqref{eq:asymderiv} and~\eqref{eq:asymderivpca}, the derivative along $\dot{L}(t)$ has a dependence on $\theta_j$, which are the principal angles between $\dot{L}(0)$ and $\dot{L}(1)$. When these two subspaces are close together, these angles are small. This means that the directional derivative is made smaller from the fact that the geodesic is shorter. To fix this issue, suppose we want to take the directional derivative between $\dot{L}$ and $L_1^*$, which have principal angles $\theta_j$. We remind ourselves of the geodesic defined with $\dot{L}(0) = \dot{L}$ and $\dot{L}(1) = L_1^*$ given in~\eqref{eq:grassgeodesic}. This geodesic can be reparametrized as
	\begin{equation}
        \dot{L}(t)  = \Sp\left( \left\{ \cos\left(t\theta_j \frac{\pi}{2 \theta_1}\right) \bv_i + \sin\left(t\theta_j \frac{\pi}{2 \theta_1}\right) \bu_i \right\}_{i=1}^d\right).
	\label{eq:grassgeodesic2}
	\end{equation}
	The reparametrization~\eqref{eq:grassgeodesic2} now has $\dot{L}(0) = \dot{L}$ and $\dot{L}( 2 \theta_1 / \pi) = L_1^*$. We have in effect lengthened the geodesic so that the maximum principal angle is $\pi/2$; now $\dot{L}(1)$ is a subspace that has principal angles $\left(\theta_j \pi/2 \theta_1\right)$ with $\dot{L}(0)$. We will refer to this as the extended geodesic between $\dot{L}$ and $L_1^*$. This geodesic maintains the property that it is still a geodesic on $G(D,d)$ from $\dot{L}$ to $L_1^*$, only now the dependence on $\theta_j$ is lessened in the derivatives~\eqref{eq:asymderiv},~\eqref{eq:discrderiv},~\eqref{eq:asymderivpca}, and~\eqref{eq:discrderivpca}.
	%All such extended geodesics of the form~\eqref{eq:grassgeodesic2} have length between $\pi/2$ and $\sqrt{d} \pi/2$.
		
	From here we fix a point $\dot{L} \in \cL_\eta$, and let $\dot{L}(t)$ be the extended geodesic between $\dot{L}$ and $L_1^*$. Define a function $C_\eta$ on $\mathcal{L}_\eta$ which is the magnitude of the directional derivative of $F_{p,\delta}^*$ at each point $\dot{L} \in \mathcal{L}_\eta$, using the extended geodesic parametrization towards $L_1^*$. By compactness of $\mathcal{L}_\eta$, $C_\eta$ has a nonzero lower bound which we can use to define $C_\eta^*$:
	\begin{equation}
	\min_{\dot{L}(0) \in \cL_\eta} \left|\frac{d}{dt}F_{p,\delta}^*(\dot{L}(t);\mu)\Big|_{t=0}\right| = \min_{\dot{L} \in \mathcal{L}_\eta} C_\eta(\dot{L},p) \geq C_{\eta}^*(p) > 0.
	\label{eq:derivlowerbd}
	\end{equation}
	For all subspaces in $\mathcal{L}_{\eta} $, the magnitude of the derivative of $F_{p,\delta}^*$ in the direction of $L_1^*$ is greater than $C_\eta^*(p)$ (using the extended geodesic parametrization).
	
	Using Lemma~\ref{lemma:deriv}, for $0 < p < 2$ we can write
	\begin{align}\label{eq:asymderivbound}
	\left| \frac{d}{dt} F_{p,\delta}^*(\dot{L}(t);\mu)\Big|_{t=0}\right| &= \left| \int_{L_1^*} -p \frac{\sum_{j=1}^d \left(\theta_j \pi/2 \theta_1\right) (\bv_j \cdot \bx)(\bu_j \cdot \bx)}{\max(\dist^{2-p}(\bx,\dot{L}(0)),p\delta)} \alpha_1 d\mu_1 \right| \\ \nonumber
	&= \left| E_{\alpha_1\mu_1} \left( - p \frac{\sum_{j=1}^d
 \left(\theta_j \pi/2 \theta_1\right) (\bv_j \cdot \bx)(\bu_j \cdot \bx)}{\max(\dist^{2-p}(\bx,\dot{L}),p\delta)}\right)\right| \\ \nonumber
	&\geq C_\eta^*(p).
	\end{align}
	When $p=2$, the expression~\eqref{eq:asymderivbound} becomes
	\begin{align}\label{eq:asymderivboundpca}
	\left| \frac{d}{dt} F_{2,\delta}^*(\dot{L}(t);\mu)\Big|_{t=0}\right| &= \left| \int_{L_1^*} -2 \sum_{j=1}^d \left(\theta_j \pi/2 \theta_1\right) (\bv_j \cdot \bx)(\bu_j \cdot \bx) \alpha_1 d\mu_1 \right| \\ \nonumber
	&= \left| E_{\alpha_1\mu_1} \left( - 2 \sum_{j=1}^d \left(\theta_j \pi/2 \theta_1\right) (\bv_j \cdot \bx)(\bu_j \cdot \bx) \right)\right| \\ \nonumber
	&\geq C_\eta^*(2).
	\end{align}
	For $\bx \in \sphere^{D-1}$ and $0 < p < 2$, let $J_p(\bx)$ be the random variable
	\begin{equation}
	J_p(\bx) = -p \frac{\sum_{j=1}^d \left(\theta_j \pi/2 \theta_1\right) (\bv_j \cdot \bx)(\bu_j \cdot \bx)}{\max(\dist^{2-p}(\bx,\dot{L}),p\delta)}.
	\label{eq:Jpx}
	\end{equation}
	For $p=2$, the random variable $J_2(\bx)$ is defined as
	\begin{equation}
	J_2(\bx) = -2 \sum_{j=1}^d \left(\theta_j \pi/2 \theta_1\right) (\bv_j \cdot \bx)(\bu_j \cdot \bx).
	\label{eq:J2x}
	\end{equation}
	
	We plan to use Hoeffding's inequality on the random variable $J_p(\bx)$ to get the overwhelming probability bounds in the theorem. For $0<p<2$,~\eqref{eq:Jpx} is absolutely bounded for $\bx \in S^{D-1}$
	\begin{align}\label{eq:Jbound1}
	&|J_p(\bx)| \leq p \left|\frac{\sum_{j=1}^d \left(\theta_j \pi/2 \theta_1\right) (\bv_j \cdot \bx)(\bu_j \cdot \bx)}{\max(\dist^{2-p}(\bx,\dot{L}),p\delta)} \right| \leq p\frac{\pi}{2}\left|\frac{\sum_{j=1}^d (\bv_j \cdot \bx)(\bu_j \cdot \bx)}{\max(\dist^{2-p}(\bx,\dot{L}),p\delta)} \right| \\ \nonumber
	&\leq p\frac{\pi}{2}\frac{\sum_{j=1}^d \left|\bu_j \cdot \bx \right|}{\max(\dist^{2-p}(\bx,\dot{L}),p\delta)} \leq p \sqrt{d} \frac{\pi}{2}  \min\left(\dist^{p-1}(\bx,\dot{L}), \frac{\dist(\bx,\dot{L})}{p\delta}  \right) .
	\end{align}
	We must split~\eqref{eq:Jbound1} into two cases:
	\begin{align} \label{eq:Jbound01}
        &|J_p(\bx)| \leq p \sqrt{d} \frac{\pi}{2} \frac{1}{(p\delta)^{(1-p)/(2-p)}}, & \ 0 < p \leq 1; \\ \label{eq:Jbound12}
        &|J_p(\bx)| \leq p \sqrt{d} \frac{\pi}{2}, & \ 1 < p < 2.
	\end{align}
	On the other hand, when $p=2$, we have a tighter bound for~\eqref{eq:J2x}
	\begin{align}\label{eq:Jboundpca}
	&|J_2(\bx)| \leq 2 \left|\sum_{j=1}^d \left(\theta_j \pi/2 \theta_1\right) (\bv_j \cdot \bx)(\bu_j \cdot \bx) \right| \leq 2\frac{\pi}{2}\left|\sum_{j=1}^d (\bv_j \cdot \bx)(\bu_j \cdot \bx) \right| \leq \pi  .
	\end{align}
	For $0<p\leq 1$ and a data set $\cX$ sampled i.i.d.~from~\eqref{mixtmeas1} with $K=1$, we can use~\eqref{eq:Jbound01} and apply Hoeffding's inequality to the random variable $J_p(\bx)$ to find
	\begin{align}
	\Pr \left(\left| \frac{\sum_{\bx_i \in \cX} J_p(\bx_i)}{N} - E_{\mu} J_p(\bx)\right| \leq \frac{C_\eta^*(p)}{2} \right) &\geq 1 - 2e^{-N \frac{C_\eta^*(p)^{2}}{2 \left(p \pi \sqrt{d} \frac{1}{(p\delta)^{(1-p)/(2-p)}} \right)^2}}.
	\label{eq:hoeffdiscrderiv}
	\end{align}
	For the case of $1 < p < 2$, we use~\eqref{eq:Jbound12} and again apply Hoeffding's inequality to find
	\begin{align}
	\Pr \left(\left| \frac{\sum_{\bx_i \in \cX} J_p(\bx_i)}{N} - E_{\mu} J_p(\bx)\right| \leq \frac{C_\eta^*(p)}{2} \right) &\geq 1 - 2e^{-N \frac{C_\eta^*(p)^{2} }{2 \left(p \pi \sqrt{d} \right)^2}}.
	\label{eq:hoeffdiscrderiv2}
	\end{align}
	Finally, when $p=2$ we use~\eqref{eq:Jboundpca} and apply Hoeffding's inequality to find
	\begin{align}
	\Pr \left(\left| \frac{\sum_{\bx_i \in \cX} J_2(\bx_i)}{N} - E_{\mu} J_2(\bx)\right| \leq \frac{C_\eta^*(2)}{2} \right) &\geq 1 - 2e^{-N \frac{C_\eta^*(2)^{2}}{8 \pi^2 }}.
	\label{eq:hoeffdiscrderivpca}
	\end{align}
	
	We observe that $E_\mu J_p(\bx) = \frac{d}{dt} F_{p,\delta}^*(L(t);\mu)\Big|_{t=0}$ and $\frac{\sum_{i=1}^N J_p(\bx_i)}{N} = \frac{1}{N}\frac{d}{dt} F_{p,\delta}(L(t);\cX)\Big|_{t=0}$. When $0<p<1$, from~\eqref{eq:asymderivbound} and \eqref{eq:hoeffdiscrderiv} we conclude
	\begin{align}\label{eq:discrderiv:probbound}
	&\left|\frac{d}{dt} \frac{F_{p,\delta}(L(t);\cX)}{N} \Bigg|_{t=0}\right| \geq \frac{C_\eta^*(p)}{4}, \ \text{w.p. $1 - 2e^{-N \frac{C_\eta^*(p)^{2} (p\delta)^{2(1-p)/(2-p)}}{2 \pi^2 p^2 d }}$}.
	\end{align}
	For $1<p<2$ we use~\eqref{eq:asymderivbound} and \eqref{eq:hoeffdiscrderiv2} to conclude
	\begin{align}\label{eq:discrderiv:probbound2}
	&\left|\frac{d}{dt} \frac{F_{p,\delta}(L(t);\cX)}{N} \Bigg|_{t=0}\right| \geq \frac{C_\eta^*(p)}{4}, \ \text{w.p. $1 - 2e^{-N \frac{C_\eta^*(p)^{2} }{2 \pi^2 p^2 d }}$}.
	\end{align}
	Finally, for $p=2$,~\eqref{eq:asymderivboundpca} and \eqref{eq:hoeffdiscrderivpca} imply that
	\begin{align}\label{eq:discrderiv:probboundpca}
	&\left|\frac{d}{dt} \frac{F_{2,\delta}(L(t);\cX)}{N} \Bigg|_{t=0}\right| \geq \frac{C_\eta^*(2)}{4}, \ \text{w.p. $1 - 2e^{-N \frac{C_\eta^*(2)^{2} }{8 \pi^2 }}$}.
	\end{align}
	In other words, for all $\dot{L} \in \mathcal{L}_\eta$, the derivative of $F_{p,\delta}$ is bounded away from zero w.o.p., which concludes the first part of the proof.
	
	We show that for any $\dot{L} \in \mathcal{L}_\eta$, there is a radius $R(\dot{L})$ such that for $\ddot{L} \in B(\dot{L},R(\dot{L}))$, there exists a directional derivative of $F_{p,\delta}$ at $\ddot{L}$ that is bounded away from zero. To show this, we must look at the derivative expression at $\dot{L}$ given by \eqref{eq:discrderiv} when $L(0) = \dot{L}$ for the extended geodesic through $L_1^*$. This derivative is continuous as a function of $\dot{L}$, $\theta_j$, $\bv_j$, and $\bu_j$. Thus, there is a $\gamma > 0$ such that $\|\theta_j - \theta_j'\| < \gamma$, $\|\bv_j - \bv_j'\| < \gamma$, $\|\bu_j - \bu_j'\| < \gamma$, and $\dist(\dot{L},\ddot{L}) < \gamma$ imply that
	\begin{align}	\label{eq:derivcont}
	\frac{1}{N}\Bigg| &\sum_{i=1}^N - p\frac{\sum_{j=1}^d \left(\theta_j \pi/2 \theta_1\right) (\bv_j \cdot \bx_i)(\bu_j \cdot \bx_i)}{\max(\dist^{2-p}(\bx_i,\dot{L}),p\delta)} - \\ \nonumber
    &\hspace{3cm} \sum_{i=1}^N - p\frac{\sum_{j=1}^d \left(\theta_j' \pi/2 \theta_1'\right) (\bv_j' \cdot \bx_i)(\bu_j' \cdot \bx_i)}{\max(\dist^{2-p}(\bx_i,\ddot{L}),p\delta)} \Bigg| < \frac{C_\eta^*(p)}{4}.
	\end{align}
	
	Now fix $\gamma_{\dot{L}}=\gamma$ with this property and fix a subspace $\ddot{L}$ such that $\dist(\dot{L},\ddot{L}) < \gamma_{\dot{L}}$. There is a minimal rotation $\mathcal{R}$ which takes vectors in $\dot{L}$ to $\ddot{L}$. Let $\hat{L}$ denote the subspace obtained from $\mathcal{R}(L_1^*)$. Then, we note that the principal angles between $\ddot{L}$ and $\hat{L}$ are identical to those between $\dot{L}$ and $L_1^*$. Further, we can define an orthonormal basis for $\ddot{L}$ as $\mathcal{R}(\bv_1),\dots,\mathcal{R}(\bv_d)$, and a complementary orthogonal basis for $\hat{L}$ by $\mathcal{R}(\bu_1),\dots,\mathcal{R}(\bu_d)$. Then, since $\dist(\dot{L},\ddot{L}) < \gamma_{\dot{L}}$, we get the inequalities $\angle(\bv_j,\mathcal{R}(\bv_j))<\gamma_{\dot{L}}$ and $\allowbreak \angle(\bu_j,\mathcal{R}(\bv_j))<\gamma_{\dot{L}}$. In turn, this then implies that $\|\bv_j - \mathcal{R}(\bv_j)\| < \gamma_{\dot{L}}$ and $\|\bu_j - \mathcal{R}(\bu_j))\|<\gamma_{\dot{L}}$. Putting this all together, for $0<p\leq 1$, from \eqref{eq:discrderiv:probbound} and \eqref{eq:derivcont} we get the bound
	\begin{align}\label{dF_cont}
	\frac{1}{N}&\left|\frac{d}{dt} F_{p,\delta}(L(t);\cX)\Big|_{t=0,L(0)=\ddot{L},L(1) = \hat{L}} \right| > \frac{C_\eta^*(p)}{4}, \ \text{w.p. $\geq 1 - 2e^{-N \frac{C_\eta^*(p)^{2} (p\delta)^{2(1-p)/(2-p)}}{2 \pi^2 p^2 d }}$}.
	\end{align}
	Repeating this argument for $1<p<2$ yields a bound similar to~\eqref{eq:derivcont} with a new $\gamma_{\dot{L}}$, which we combine with~\eqref{eq:discrderiv:probbound2} to find
	\begin{align}\label{dF_cont2}
	\frac{1}{N}&\left|\frac{d}{dt} F_{p,\delta}(L(t);\cX)\Big|_{t=0,L(0)=\ddot{L},L(1) = \hat{L}} \right| > \frac{C_\eta^*(p)}{4}, \ \text{w.p. $\geq 1 - 2e^{-N \frac{C_\eta^*(p)^{2} }{2 \pi^2 p^2 d }}$}.
	\end{align}
	Finally, by repeating the continuity argument for $p=2$, we again find a similar expression to~\eqref{eq:derivcont} with a new $\gamma_{\dot{L}}$, which we combine with~\eqref{eq:discrderiv:probboundpca} to find
	\begin{align}\label{dF_contpca}
	\frac{1}{N}&\left|\frac{d}{dt} F_{2,\delta}(L(t);\cX)\Big|_{t=0,L(0)=\ddot{L},L(1) = \hat{L}} \right| > \frac{C_\eta^*(2)}{4}, \ \text{w.p. $\geq 1 - 2e^{-N \frac{C_\eta^*(2)^{2}}{8 \pi^2 }}$}.
	\end{align}
	Finally, if we let $R_p(\dot{L}) = \gamma_{\dot{L}}$, we can find this relation for all $\dot{L} \in \cL_\eta$, and the existence of the function $R_p$ is concluded.
	\qed
\end{proof}

The set $\mathcal{L}_\eta$ can be covered by the set of open balls $\{B(\dot{L},R_p(\dot{L})) : L \in \mathcal{L}_\eta\}$ by Proposition~\ref{prop:derivative}. By compactness of the set $\mathcal{L}_\eta$, there is a finite sub-cover
\begin{equation}\label{eq:finitesubcov}
\{ B(\dot{L}_1,R(\dot{L}_1)),\dots,B(\dot{L}_{m_p},R(\dot{L}_{m_p})) \}.
\end{equation}
Within each of these balls, there are no stationary points w.o.p.~depending on the directional derivatives of $F_{p,\delta}^*$ at $\dot{L}_1,\dots,\dot{L}_{m_p}$ towards $L_1^*$. Using this observation and~\eqref{dF_cont}, for $0 < p \leq 1$, we get the desired result by the union bound
\begin{align}\label{eq:probnostat}
\Pr&\left(\mathcal{L}_\eta \text{ contains no stationary points} \right) \geq 1 - 2 m_p e^{-N \frac{C_\eta^*(p)^{2} (p\delta)^{2(1-p)/(2-p)}}{2 \pi^2 p^2 d }}.
\end{align}
For $1< p < 2$, from~\eqref{dF_cont2} we get
\begin{align}\label{eq:probnostat2}
\Pr&\left(\mathcal{L}_\eta \text{ contains no stationary points} \right) \geq 1 - 2 m_p e^{-N \frac{C_\eta^*(p)^{2} }{2 \pi^2 p^2 d }}.
\end{align}
For the case $p=2$, using~\eqref{dF_contpca}, the union bound gives
\begin{align}\label{eq:probnostatpca}
\Pr&\left(\mathcal{L}_\eta \text{ contains no stationary points} \right) \geq 1 - 2 m_2 e^{-N \frac{C_\eta^*(2)^{2}}{8 \pi^2 }}.
\end{align}
The final probability bounds in~\eqref{eq:wopFMS0p1},~\eqref{eq:wopFMS1p2}, and~\eqref{eq:woppca} follow from~\eqref{eq:probnostat},~\eqref{eq:probnostat2}, and~\eqref{eq:probnostatpca} using the bounds derived later for $C_\eta^*(p)$ in~\eqref{eq:fmsderivbd},~\eqref{eq:fmsderivbd2}, and~\eqref{eq:pcaderivbd}. For $0<p\leq 1$, the probability bound~\eqref{eq:probnostat} becomes
\begin{align}\label{eq:probnostatfin}
\Pr&\left(\mathcal{L}_\eta \text{ contains no stationary points} \right) \geq  \\ \nonumber &\hspace{3cm}
1 - 2 m_p e^{-N \left(\frac{ (p\delta)^{2(1-p)/(2-p)}}{2 \pi^2 p^2 d } \left(p\alpha_1 \frac{1}{d} \right)^2 \right) \min\left( \left(\frac{\pi}{6}\right) ^{2(p-1)},\frac{\eta^2}{(p\delta)^2}\right) }.
\end{align}
For $1< p < 2$, the probability bound~\eqref{eq:probnostat2} becomes
\begin{align}\label{eq:probnostatfin2}
\Pr&\left(\mathcal{L}_\eta \text{ contains no stationary points} \right) \geq \\ \nonumber & \hspace{3cm} 1 - 2 m_p e^{-N \left(\frac{1}{2 \pi^2 p^2 d } \left(p\alpha_1 \frac{1}{d} \right)^2 \right) \min\left( \eta^{2(p-1)},\frac{\eta^2}{(p\delta)^2}\right) }.
\end{align}
For the case $p=2$, the probability bound~\eqref{eq:probnostatpca} becomes
\begin{align}\label{eq:probnostatfinpca}
\Pr&\left(\mathcal{L}_\eta \text{ contains no stationary points} \right) \geq 1 - 2 m_2 e^{-N   \left(\frac{1}{8 \pi^2}\left(2\alpha_1 \frac{1}{d^2} \right)^2 \right)    \eta^2 }.
\end{align}

Since FMS$_p$ must converge to a stationary point, by Lemma~\ref{lemma:statpoint} and~\eqref{eq:probnostat} we conclude that $(L_k)_{k \in \nats}$ converges to $B(Z_{F_{p,\delta}^*},\eta)$.

\subsubsection{Conclusion of Theorem~\ref{thm:globconv}}
\label{sec:globconv:proof:fin}

It remains to show that the point recovered by FMS$_p$ or PCA lies in $B(L_1^*,\eta)$. We define a set
\begin{equation}
    \cB(\dot{L},c) = \{L \in G(D,d): \theta_1(\dot{L},L) < c\}.
\end{equation}
To show that the PCA solution is componentwise separated from $L_1^{*\perp}$, we use the following Lemma. With abuse of notation here, we also $B(L,\eta) = \{\bv \in S^{D-1} : \angle(\bv,L) < \eta\}$, and the context will make clear whether $B(L,\eta)$ is a set of vectors or subspaces.

\begin{lemma}\label{lemma:pcamaxmin}
	For all $\bv \in \overline{B(L_1^{*\perp},\eta)}$ and all $\bu \in \overline{B(L_1^*,\eta)}$, $F_{2,\delta}(\bv;\cX) > F_{2,\delta}(\bu;\cX)$ w.o.p.
\end{lemma}

\begin{proof}
To prove this lemma, we first look at the asymptotic PCA cost $F_{2,\delta}^*$ in each of these neighborhoods. We notice that we can separate this cost by measure:
\begin{equation}
	F_{2,\delta}^*(\bv;\alpha_0\mu_0 + \alpha_1\mu_1) = \alpha_0 F_{2,\delta}^*(\bv;\mu_0) + \alpha_1F_{2,\delta}^*(\bv;\mu_1)
\end{equation}
Define the function $\varphi_\eta^* : \overline{B(L_1^{*\perp},\eta)} \times \overline{B(L_1^*,\eta)} \to (0,\infty)$. Note that
\begin{align}\label{eq:pcamaxminphi}
	\varphi_\eta^*(\bv,\bu;\mu) &= F_{2,\delta}^*(\bv;\mu) - F_{2,\delta}^*(\bu;\mu) = \alpha_1 F_{2,\delta}^*(\bv;\mu_1) - \alpha_1 F_{2,\delta}^*(\bu;\mu_1) \\ \nonumber
	&= \alpha_1 \int_{L_1^*} \dist^2(\bx,\bv) - \dist^2(\bx,\bu) d\mu_1 \\ \nonumber
	&\geq \alpha_1\left(\sin^2(\pi/2-\eta) - \sin^2(\eta) \right).
\end{align}
For $\eta < \pi/6$, we have the bound
\begin{align}
\varphi(\bv,\bu;\mu) &\geq \alpha_1 \left(\frac{\pi}{2} \left(\frac{\pi}{2} - \eta \right) -  \frac{\pi}{2}\eta\right) = \alpha_1 \left(\frac{\pi^2}{4} - \pi \eta \right) .
\end{align}
Thus, for small enough $\eta$, $\varphi(\cdot,\cdot;\mu)$ is bounded above zero. Define the random variable $\varphi(\bx;\bv,\bu)$ for $\bx \in S^{D-1}$ by
\begin{equation}
	\varphi(\bx;\bv,\bu) = \dist^2(\bx,\bv) - \dist^2(\bx,\bu).
\end{equation}
Then, $\varphi(\bx;\bv,\bu) \in [-1,1]$ for all $\bx \in S^{D-1}$. We use Hoeffding's inequality to write
\begin{align}\label{eq:hoeffphipca}
	\Pr&\left(\left| \frac{1}{N}\sum_{i=1}^N \varphi(\bx_i;\bv,\bu) - E_{\alpha_1 \mu_1}\left( \varphi(\bx;\bv,\bu) \right)  \right| > \frac{1}{2} \alpha_1 \left(\frac{\pi^2}{4} - \pi \eta \right) \right) \geq \\ \nonumber
	&1 - 2e^{-N \left(\frac{1}{2} \alpha_1 \left(\frac{\pi^2}{4} - \pi \eta \right) \right)^2 \frac{1}{2}}.
\end{align}
By continuity of $\varphi(\bv,\bu;\bx)$, there exists a $\gamma$ such that $\angle(\bv,\bv') < \gamma$ and $\angle(\bu,\bu') < \gamma$ implies that
\begin{equation}
\frac{1}{N}\sum_{i=1}^N \varphi(\bx_i;\bv,\bu) - \frac{1}{N}\sum_{i=1}^N \varphi(\bx_i;\bv',\bu') < \frac{1}{4} \alpha_1 \left(\frac{\pi^2}{4} - \pi \eta \right).
\end{equation}
Thus, by a covering argument similar to the proof of Proposition~\ref{prop:derivative}, we get that $\varphi(\bv,\bu;\cX)>0$ for all $(\bv,\bu) \in \overline{B(L_1^{*\perp},\eta)} \times \overline{B(L_1^*,\eta)}$ w.o.p.~
\qed
\end{proof}
When $\eta < \pi/6$, the probability~\eqref{eq:hoeffphipca} dominates~\eqref{eq:probnostatfinpca}. Altogether, the fact that PCA can be defined sequentially, Lemma~\ref{lemma:pcamaxmin}, and~\eqref{eq:probnostatfinpca} imply that the PCA solution lies in $B(L_1^*,\eta)$ w.o.p.~stated in~\eqref{eq:probnostatfinpca}.

To extend to FMS$_p$, we use PCA initialization. With PCA initialization, the probability that $L_0$ (the initial FMS$_p$ iterate) is in $\cB(L_1^{*\perp},\pi/6)$ is
\begin{equation}
\Pr(L_0 \subset \cB(L_1^{*\perp},\eta) \neq \{\bzero\} ) \leq C_1'' e^{- C_2'' N (\pi/6)^2}.
\end{equation}
This comes from applying the PCA probability bound~\eqref{eq:woppca} for $\eta = \pi/6$ recovery. Thus, w.o.p.~that dominates the bound given in~\eqref{eq:wopFMS0p1} and~\eqref{eq:wopFMS1p2}, the initial FMS$_p$ iterate lies in $\cB(L_1^*,\pi/6)$.

\begin{lemma}\label{lemma:fmsmaxmin}
	The global minimum of $H_{p,\delta}(L,L_1;\cX)$ lies in $\overline{B(L_1^*,\pi/6)}$ for all $L_1 \in B(L_1^*,\pi/6)$ w.o.p.
\end{lemma}

By repeating the same argument used in~\S\ref{sec:globconv:proof:deriv}, one can show that w.o.p.there are no stationary points of $H_{p,\delta}(L,L_1;\cX)$ in $ G(D,d) \setminus \left(B(L_1^*,\pi/6) \cup B(L_1^{*\perp},\pi/6) \right)$ for all $L_1 \in B(L_1^*,\pi/6)$. Thus, to prove the lemma, it suffices to show that scaled version of~\eqref{eq:fmsmaxminphi} is positive w.o.p.Fix a point $L_1 \in B(L_1^*,\pi/6)$. Define the function $\psi_\eta^* : \overline{B(L_1^{*\perp},\eta)} \times \overline{B(L_1^*,\eta)} \to (0,\infty)$. By
\begin{align}\label{eq:fmsmaxminphi}
\psi_\eta^*(\bv,\bu;\mu) &= H_{p,\delta}^*(\bv,L_1;\mu) - H_{p,\delta}^*(\bu,L_1;\mu) \geq \alpha_1 H_{p,\delta}^*(\bv,L_1;\mu) - \alpha_1 H_{p,\delta}^*(\bu,L_1;\mu) \\ \nonumber
&\geq \alpha_1 \int_{L_1^*} \frac{\dist^2(\bx,\bv) - \dist^2(\bx,\bu)}{\max(\dist^{2-p}(\bx,L_1),p\delta)} d\mu_1 \\ \nonumber
&\geq \alpha_1\left(\sin^p(\pi/2-\eta) - \sin^p(\eta) \right).
\end{align}
Thus, by the same argument used in Lemma~\ref{lemma:pcamaxmin}, the discrete version of $\psi_\eta(\bv,\bu;\cX)$ is positive for all $\bv \in \overline{B(L_1^{*\perp},\pi/6)}$ and $\bu \in \overline{B(L_1^*,\pi/6)}$ w.o.p.Further, by a covering argument, this is true for all $L_1 \in B(L_1^*,\pi/6)$ w.o.p.
\qed

By Lemma~\ref{lemma:pcamaxmin}, we conclude that the initial FMS$_p$ iterate $L_0 \in B(L_1^*,\pi/6)$ w.o.p., and that this probability dominates the probabilities~\eqref{eq:probnostat} and~\eqref{eq:probnostat2} when $\eta < \pi/6$. By Lemma~\ref{lemma:fmsmaxmin}, the next FMS$_p$ iterate from any point in $\overline{B(L_1^*,\pi/6)}$ lies in $B(L_1^*,\eta)$ w.o.p., and this probability dominates the probabilities~\eqref{eq:probnostat} and~\eqref{eq:probnostat2} when $\eta < \pi/6$. Thus, the limiting probabilities of recovery are~\eqref{eq:probnostat} and~\eqref{eq:probnostat2}, and FMS$_p$ converges to $B(L_1^*,\eta)$ w.o.p.as stated in~\eqref{eq:probnostat} and~\eqref{eq:probnostat2}.
\qed

\subsubsection{Conclusion of Theorem~\ref{thm:globconv:stab}}
\label{sec:globconv:proof:stab}

In this section, we analyze the stability of the global convergence result when small noise is added to the data set. We now consider the more general mixture measure $\mu_\varepsilon$ given in~\eqref{mixtmeasnoise} and are able to get convergence of $L_k$ to a point in $B(L_1^*,\eta)$, provided that the noise is not too large.

\begin{proof}[Proof of Theorem~\ref{thm:globconv:stab}]
	%{\em Proof of Proposition~\ref{thm:globconv:stab}.}
	We require that the $p$th moment of $\nu_{1,\varepsilon}$ is less than $\varepsilon^p$. Let $L(t)$ be an extended geodesic from a point in $\cL_\eta$ through $L_1^*$. Then, for $0<p<2$, the difference between the derivatives asymptotic costs associated with $\mu$ and $\mu_{\varepsilon}$ can be written as
	\begin{align}\label{eq:noiseest}
	\left|\frac{d}{dt}F_{p,\delta}^*(L(t);\mu)\Big|_{t=0} - \frac{d}{dt}F_{p,\delta}^*(L(t);\mu_\varepsilon)\Big|_{t=0}\right| &= \E_{\alpha_1 \nu_{1,\varepsilon}} \left( p\frac{\sum_{j=1}^d \left(\theta_j \pi/2 \theta_1\right) (\bu_j \cdot \bx)(\bv_j \cdot \bx)}{\max(\dist^{2-p}(\bx,L(0)),p\delta)} \right) \\ \nonumber
	&\leq \frac{p \alpha_1 \pi \sqrt{d}}{2 \theta_1} \E_{\nu_{1,\varepsilon}}(\|\bx\|^p) \leq \frac{p \alpha_1 \pi \sqrt{d} \varepsilon^p}{2 \theta_1}.
	\end{align}
	On the other hand, for $p=2$, the difference between the asymptotic derivatives can be written as
	\begin{align}\label{eq:noiseestpca}
	\left|\frac{d}{dt}F_{2,\delta}^*(L(t);\mu)\Big|_{t=0} - \frac{d}{dt}F_{2,\delta}^*(L(t);\mu_\varepsilon)\Big|_{t=0}\right| &= \E_{\alpha_1 \nu_{1,\varepsilon}} \left( 2 \sum_{j=1}^d \left(\theta_j \pi/2 \theta_1\right) (\bu_j \cdot \bx)(\bv_j \cdot \bx) \right) \\ \nonumber
	&\leq \frac{\pi}{ \theta_1} \alpha_1 \E_{\nu_{1,\varepsilon}}(\bx^2) \leq \alpha_1 \frac{\pi}{ \theta_1} \varepsilon^2.
\end{align}
With these bounds, Proposition~\ref{prop:derivative} holds for $K=1$ with the noisy mixture measure~\eqref{mixtmeasnoise} for small enough $\varepsilon$. For example choosing $\varepsilon < \left(\frac{C^*_{\eta}(p)}{4} \cdot \frac{2\eta}{p\pi d \alpha_1}\right)^{1/p}$ for $0<p<2$ or $\varepsilon < \left(\frac{C^*_{\eta}(2)}{4} \cdot \frac{\eta}{\pi \sqrt{d} \alpha_1}\right)^{1/2}$ for $p=2$ allows the proof to still work. Combining this fact with the bounds on $C_{\eta}^*(p)$ given later in~\eqref{eq:pcaderivbd},~\eqref{eq:fmsderivbd}, and~\eqref{eq:fmsderivbd2},  we get the upper bounds on the magnitude of $\varepsilon$ for $\eta$-recovery given in Theorem~\ref{thm:globconv:stab}.

In fact, we can modify the proof of Proposition~\ref{prop:derivative} to hold for any $\pi/6 \geq \eta > 0$ provided that $\varepsilon < \left({C^*_{\eta}(p)} \cdot \frac{2 \eta}{p\pi d \alpha_1}\right)^{1/p}$ for $0<p<2$ or $\varepsilon < \left({C^*_{\eta}(2)} \cdot \frac{\eta}{\pi  \sqrt{d} \alpha_1}\right)^{1/2}$ for $p=2$.
\qed
\end{proof}

\subsubsection{Bounds on $C_\eta^*(p)$ and $m_p$}
%\subsubsection{Dependence of the Probabilities~\eqref{eq:probnostat} and~\eqref{eq:probnostatpca} on Parameters}
\label{sec:globconv:proof:parameters}

This section will seek to provide useful bounds on the constants $C_\eta^*(p)$ defined in~\eqref{eq:derivlowerbd} and $m_p$ defined in~\eqref{eq:finitesubcov}. While~\eqref{eq:probnostat},~\eqref{eq:probnostat2}, and~\eqref{eq:probnostatpca} give the overwhelming probability of recovery for both FMS$_p$ and PCA, we must examine the constants to see what kind of gain FMS$_p$ gives over PCA. In the following analysis, we restrict ourselves to the set of subspaces $\cB(L_1^*,\pi/6)$. It readily apparent that $\cB(L_1^*,\pi/6) \supseteq B(L_1^*,\pi/6)$. We restrict to this set to allow favorable estimates on $C_\eta^*(p)$: the derivative $\frac{d}{dt} F_{p,\delta}^*(L(t))\big|_{t=0} \to 0$ as $L(0) \to L_1^{*\perp}$. This restriction is reasonable if we take PCA initialization for FMS$_p$. From Theorem~\ref{thm:globconv} with $p=2$, taking $\eta = \pi/6$, the PCA solution lies in this set w.o.p.~Thus, FMS$_p$ with PCA initialization starts in $\cB(L_1^*,\pi/6)$ w.o.p., and so we focus on probabilistic recovery~\eqref{eq:probnostat} and~\eqref{eq:probnostat2} within this neighborhood of $L_1^*$.

We begin by noticing that
\begin{equation}\label{eq:intxi2}
\int_{S^{d-1}} x_1^2 d\sigma = \frac{1}{d},
\end{equation}
where $\bx = (x_1,\dots,x_d)$ and $\sigma$ is the uniform distribution on $S^{d-1}$. By symmetry, this integral is equal for all choices of $x_1,\dots,x_d$. Now consider the derivative of $F_{2,\delta}^*$ from a point $L(0) \in \cB(L_1^*,\pi/6)$ in the direction of $L_1^*$ using the extended geodesic $L(t)$
\begin{equation}\label{eq:asymPCAderiv}
\frac{d}{dt} F_{2,\delta}^* (L(t);\mu) \Big|_{t=0} = -2 \alpha_1 \sum_{j=1}^d \left(\theta_j \pi/2 \theta_1\right) \int_{L_1^*} (\bv_j \cdot \bx) (\bu_j \cdot \bx) d\mu_1.
\end{equation}
%For each $j$, the integral term in~\eqref{eq:asymPCAderiv} has the property
%\begin{equation}
%\int_{L_1^*} (\bv_j \cdot \bx) (\bu_j \cdot \bx) d\mu_1 \searrow 0 \  \mathrm{as } \ \dist(\bv_j,L_1^*) \to 0.
%\end{equation}	
%This should be expected because $L_1^*$ is a stationary point and the global minimum of the asymptotic PCA cost function $F_{2,\delta}^*$. Thus, the value $C_{\eta}^*(2)$ which lower bounds the derivative of $F_{2,\delta}^*$ is given by a point on the boundary of $\overline{B(L_1^*,\eta)}$.
Let $\bx^j$ be a unit vector spanning $L_1^* \cap \Sp(\bu_j,\bv_j)$ such that $\angle(\bx_j,\bu_j) < \pi/2$ and $\angle(\bx_j,\bv_j) < \pi/2$. We can rewrite~\eqref{eq:asymPCAderiv} using~\eqref{eq:intxi2}
\begin{align}
\frac{d}{dt} F_{2,\delta}^* (L(t);\mu) \Big|_{t=0} &= -2\alpha_1 \sum_{j=1}^d \left(\theta_j \pi/2 \theta_1\right)  (\bv_j \cdot \bx^j) (\bu_j \cdot \bx^j) \frac{1}{d} \\ \nonumber
&= -2\alpha_1 \frac{1}{d} \sum_{j=1}^d \left(\theta_j \pi/2 \theta_1\right) \cos(\theta_j)\sin(\theta_j).
\end{align}
From this formulation of the derivative, we obtain the inequality
\begin{equation}
- \left(2\alpha_1 \frac{\pi}{2} \frac{1}{d}  \sum_{j=1}^d \theta_j\right) \leq \frac{d}{dt} F_{2,\delta}^* (L(t)) \Big|_{t=0} \leq - \left(2\alpha_1 \frac{1}{d}  \theta_1\right).
\label{eq:p}
\end{equation}
In a similar fashion, we restate the derivative of the asymptotic FMS$_p$ cost derived in Lemma~\ref{lemma:deriv} using the extended geodesic parametrization
\begin{equation}\label{eq:proof:hoeffbd:pcaderivbd}
\frac{d}{dt} F_{p,\delta}^* (L(t);\mu) \Big|_{t=0} = - p \alpha_1 \int_{L_1^*}  \frac{\sum_{j=1}^d \left(\theta_j \pi/2 \theta_1\right) (\bv_j \cdot \bx)(\bu_j \cdot \bx)}{\max(\dist^{2-p}(\bx_i,L(0)),p\delta)} d\mu_1 .
\end{equation}
Let $c(L(0)) = \max(\max_{\bx \in L(0)}(\dist^{2-p}(\bx,L_1^*)),p\delta)$. Then, we have the following bound on the derivative of $F_{p,\delta}^* (L(t);\mu)$ over $\cB(L_1^*,\pi/6)$
%\begin{align}\label{eq:proof:hoeffbd:fmsderivupperbd}
%\frac{d}{dt} F_{p,\delta}^* (L(t);\mu) \Big|_{t=0} &\leq - p\alpha_1 \frac{1}{c(L(0))}  \int_{L_1^*}  \sum_{j=1}^d \left(\theta_j \pi/2 \theta_1\right) (\bv_j \cdot \bx)(\bu_j \cdot \bx) d\mu_1  \\ \nonumber
%& \leq - p\alpha_1 \frac{1}{c(L(0))} \frac{1}{d} \theta_1.
%\end{align}
%\begin{align}\label{eq:proof:hoeffbd:fmsderivlowerbd}
%\frac{d}{dt} F_{p,\delta}^* (L(t);\mu) \Big|_{t=0} &\geq - p\alpha_1 \frac{1}{p\delta}  \int_{L_1^*}  \sum_{j=1}^d \left(\theta_j \pi/2 \theta_1\right) (\bv_j \cdot \bx)(\bu_j \cdot \bx) d\mu_1  \\ \nonumber
%& \geq - p\alpha_1 \frac{1}{p\delta} \frac{\pi}{2} \frac{1}{d}  \sum_{j=1}^d \theta_j.
%\end{align}
\begin{align}\label{eq:proof:hoeffbd:fmsderivbd}
 \frac{d}{dt} F_{p,\delta}^* (L(t);\mu) \Big|_{t=0} &\leq - \frac{1}{c(L(0))} \left(p\alpha_1  \frac{1}{d} \theta_1\right).
\end{align}
We will now make the bounds~\eqref{eq:proof:hoeffbd:pcaderivbd} and~\eqref{eq:proof:hoeffbd:fmsderivbd} more clear. Using the fact that $\dist(L(0),L_1^*) = \sqrt{\sum_{j=1}^d \theta_j^2}$,~\eqref{eq:proof:hoeffbd:pcaderivbd} becomes
\begin{equation}
- \left(2\alpha_1 \frac{\pi}{2} \frac{1}{\sqrt{d}}  \dist(L(0),L_1^*) \right) \leq \frac{d}{dt} F_{2,\delta}^* (L(t)) \Big|_{t=0} \leq - \left(2\alpha_1 \frac{1}{d^2}  \dist(L(0),L_1^*) \right).
\label{eq:proof:hoeffbd:pcaderivbd2}
\end{equation}
Further, using the fact that $c(L(0)) \leq \max(\dist^{2-p}(L(0),L_1^*),p\delta)$,~\eqref{eq:proof:hoeffbd:fmsderivbd} becomes
\begin{align}\label{eq:proof:hoeffbd:fmsderivbd2}
\frac{d}{dt} F_{p,\delta}^* (L(t);\mu) \Big|_{t=0} &\leq - \left(p\alpha_1  \frac{1}{d} \frac{\theta_1}{\max(\theta_1^{2-p},p\delta)} \right) \\ \nonumber
& \leq - \left(p\alpha_1  \frac{1}{d} \min\left(\theta_1^{p-1},\frac{\theta_1}{p\delta} \right) \right).
\end{align}
From~\eqref{eq:proof:hoeffbd:pcaderivbd2} restricted to the set $\cB(L_1^*,\pi/6)\cap \cL_\eta = \cB(L_1^*,\pi/6) \setminus B(L_1^*,\eta)$, the bounds on the PCA constant $C_\eta^*(2)$ are given by
\begin{equation}\label{eq:pcaderivbd}
    \left(2\alpha_1 \frac{1}{d^2}  \eta \right)  \leq C_{\eta}^*(2) \leq \left(2\alpha_1 \frac{\pi}{2} \frac{1}{\sqrt{d}}  \eta \right).
\end{equation}
Over the set $\cB(L_1^*,\pi/6) \cap \cL_\eta$, we use~\eqref{eq:proof:hoeffbd:fmsderivbd2} to derive two bounds for the FMS$_p$ constant $C_\eta^*(p)$. If $0<p\leq 1$, then
\begin{equation}\label{eq:fmsderivbd}
\left(p\alpha_1 \frac{1}{d}  \min\left( \left(\frac{\pi}{6}\right) ^{p-1},\frac{\eta}{p\delta}\right) \right)  \leq C_{\eta}^*(p).
\end{equation}
On the other hand, if $1 < p < 2$,
\begin{equation}\label{eq:fmsderivbd2}
\left(p\alpha_1 \frac{1}{d}  \min\left(\eta^{p-1},\frac{\eta}{p\delta}\right) \right)  \leq C_{\eta}^*(p) .
\end{equation}
From this, we see the dependence of $C_\eta^{*}(p)$ on $p$, $d$, $\alpha_1$, and $\eta$.

While it is hard to come up with closed form expressions for the constants ${m_p}$ in the probabilities~\eqref{eq:probnostat} and~\eqref{eq:probnostatpca}, it is still important to see the dependence on $D$, $d$, $p$, and $\delta$. Proposition~\ref{prop:coveringfms} gives our bounds for the covering numbers for FMS$_p$.

\begin{proposition}\label{prop:coveringfms}
	At worst, the number of covering balls $m_p$ for $0<p\leq 1$ is
	\begin{align}
	 m_p = O\left( d^{4d(D-d)} \delta^{-\frac{3-2p}{2-p}} \max\left( \left(\frac{\pi}{6}\right)^{(1-p)d(D-d)},\left(\frac{\eta}{p\delta}\right)^{-d(D-d)} \right)  \right).
	\end{align}
	At worst, the number of covering balls $m_p$ for $1<p < 2$ is
	\begin{align}
	m_p = O\left( d^{4d(D-d)} \delta^{-\frac{3-2p}{2-p}} \max\left( \eta^{(1-p)d(D-d)},\left(\frac{\eta}{p\delta}\right)^{-d(D-d)} \right)  \right).
	\end{align}
	At worst, the number of covering balls $m_2$ for $p=2$ is
	\begin{align}
	m_2 &= O\left( \eta^{-d(D-d)} d^{2d(D-d)} \right).
	\end{align}
\end{proposition}

\begin{proof}
	
	We first remind ourselves of some details of the proof of Proposition~\ref{prop:derivative} used in the proof of Theorem~\ref{thm:globconv}. The covering argument relies on finding a function $R_p: \cL_\eta \to (0,\infty)$ such that for each $L \in \cL_\eta$, the derivative is non-stationary for all points in $B(L,R_p(L))$ w.o.p.~To prove this proposition, we will bound the radius below. The bound on the radius function then gives a bound on the necessary number of covering balls.
	
	Looking at the continuity statement on the derivative of $F_{p,\delta}$, we can rewrite the difference expressed in~\eqref{eq:derivcont}, when $\theta_j = \theta_j'$ and $\dist(\dot{L},\ddot{L}) \leq \gamma$  as
	\begin{align}
	&\frac{1}{N}\Bigg| \sum_{i=1}^N - p\frac{\sum_{j=1}^d \left(\theta_j \pi/2 \theta_1\right) (\bv_j \cdot \bx_i)(\bu_j \cdot \bx_i)}{\max(\dist^{2-p}(\bx_i,\dot{L}),p\delta)} - \sum_{i=1}^N - p\frac{\sum_{j=1}^d \left(\theta_j' \pi/2 \theta_1'\right) (\bv_j' \cdot \bx_i)(\bu_j' \cdot \bx_i)}{\max(\dist^{2-p}(\bx_i,\ddot{L}),p\delta)} \Bigg|\\ \nonumber
	&= \frac{p}{N}\Bigg| \sum_{i=1}^N -\left( \frac{1}{\max(\dist^{2-p}(\bx_i,\dot{L}),p\delta)} \cdot \frac{1}{\max(\dist^{2-p}(\bx_i,\ddot{L}),p\delta)} \right) \cdot  \\ \nonumber
	&\sum_{j=1}^d \left(\theta_j \pi/2 \theta_1\right) \Big(\max(\dist^{2-p}(\bx_i,\ddot{L}),p\delta)(\bv_j \cdot \bx_i)(\bu_j \cdot \bx_i) - \\ \nonumber
	&\ \ \ \ \ \   \max(\dist^{2-p}(\bx_i,\dot{L}),p\delta)(\bv_j' \cdot \bx_i)(\bu_j' \cdot \bx_i) \Big) \Bigg| \\ \nonumber
	&\leq \frac{p}{N} \frac{\pi}{2} \sum_{i = 1}^N  \Bigg|\sum_{j=1}^d \left( \frac{\max(\dist^{2-p}(\bx_i,\ddot{L}),p\delta) - \max(\dist^{2-p}(\bx_i,\dot{L}),p\delta)}{\max(\dist^{2-p}(\bx_i,\dot{L}),p\delta) \cdot \max(\dist^{2-p}(\bx_i,\ddot{L}),p\delta)} \right) (\bv_j \cdot \bx_i)(\bu_j \cdot \bx_i) - \\ \nonumber
	& \left( \frac{\max(\dist^{2-p}(\bx_i,\dot{L}),p\delta)}{\max(\dist^{2-p}(\bx_i,\dot{L}),p\delta) \cdot \max(\dist^{2-p}(\bx_i,\ddot{L}),p\delta)} \right) \left((\bv_j \cdot \bx_i)(\bu_j \cdot \bx_i) - (\bv_j' \cdot \bx_i)(\bu_j' \cdot \bx_i) \right) \Bigg| \\ \nonumber
	&\leq \frac{p}{N} \frac{\pi}{2}\sum_{i = 1}^N \sum_{j=1}^d \Bigg| \left( \frac{\gamma}{p\delta} \right) \frac{(\bv_j \cdot \bx_i)(\bu_j \cdot \bx_i)}{\max(\dist^{2-p}(\bx_i,\dot{L}),p\delta)}\Bigg| + \\ \nonumber
	&\ \ \ \ \ \  \Bigg|\sum_{j=1}^d \left( \frac{1}{\max(\dist^{2-p}(\bx_i,\ddot{L}),p\delta)} \right) \left((\bv_j \cdot \bx_i)(\bu_j \cdot \bx_i) - (\bv_j' \cdot \bx_i)(\bu_j' \cdot \bx_i) \right) \Bigg| \\ \nonumber
	&\leq \frac{p}{N} \frac{\pi}{2} \sum_{i = 1}^N  \sum_{j=1}^d  \left( \frac{\gamma}{p\delta}\right)\left( \frac{1}{(p\delta)^{\frac{1-p}{2-p}}} \right) + \sum_{j=1}^d  \left(\frac{1}{p\delta}\right) \left|(\bv_j \cdot \bx_i)(\bu_j \cdot \bx_i) - (\bv_j' \cdot \bx_i)(\bu_j' \cdot \bx_i) \right| \\ \nonumber
    &\leq \frac{p}{N} \frac{\pi}{2} \sum_{i = 1}^N \sum_{j=1}^d  \frac{\gamma}{p\delta} \left(\frac{1}{(p\delta)^{\frac{1-p}{2-p}}} +  2  \right) \leq  \frac{\pi}{2} \frac{\gamma}{\delta} d \left(\frac{1}{(p\delta)^{\frac{1-p}{2-p}}} +  2 \right).
	\end{align}
	Thus, if we choose
	\begin{equation}
	\gamma_1 = \frac{C_{\eta}^*(p) }{4} \frac{2\delta}{\pi d \left(\frac{1}{(p\delta)^{\frac{1-p}{2-p}}} +  2\right)},
	\end{equation}
	then the radius function $R$ is bounded below by $\gamma_1$. We can cover $G(D,d)$ by $(C_4)^{d(D-d)} / (\gamma_1)^{d(D-d)}$ balls of radius $\gamma_1$ using Remark 8.4 of~\cite{szarek83}, for a universal constant $C_4$. As a consequence, for $0<p\leq 1$ we can use the inequality~\eqref{eq:fmsderivbd} to bound the order of the covering number $m_p$ for $G(D,d)$
	\begin{align}
	m_p &= \left(\frac{ O\left(d \left((p\delta)^{-\frac{1-p}{2-p}} +  2\right)\right)}{C_\eta^* \delta}\right)^{d(D-d)} \leq \left(\frac{O\left(d (p\delta)^{-\frac{1-p}{2-p}}\right)}{O\left(d^{-2} \delta \frac{1}{d}  \min\left( \left(\frac{\pi}{6}\right)^{p-1},\frac{\eta}{p\delta}\right) \right) } \right)^{d(D-d)} \\ \nonumber
	&=  \left(O\left( d^{4} (p\delta)^{-\frac{1-p}{2-p}} \delta^{-1} \max\left( \left(\frac{\pi}{6}\right)^{1-p},\left(\frac{\eta}{p\delta}\right)^{-1} \right) \right) \right)^{d(D-d)}.
	\end{align}
	For $1 < p < 2$, we use the inequality~\eqref{eq:fmsderivbd2} to bound the order of the covering number $m_p$ for $G(D,d)$
	\begin{align}
	m_p &= \left(\frac{ O\left(d \left((p\delta)^{-\frac{1-p}{2-p}} +  2\right)\right)}{C_\eta^* \delta}\right)^{d(D-d)} \leq \left(\frac{O\left(d (p\delta)^{-\frac{1-p}{2-p}}\right)}{O\left(d^{-2} \delta \frac{1}{d}  \min\left(\eta^{p-1},\frac{\eta}{p\delta}\right) \right) } \right)^{d(D-d)} \\ \nonumber
	&=  \left(O\left( d^{4} (p\delta)^{-\frac{1-p}{2-p}} \delta^{-1} \max\left( \eta^{1-p},\left(\frac{\eta}{p\delta}\right)^{-1} \right) \right) \right)^{d(D-d)}. \\ \nonumber
	\end{align}
	For $p=2$, a simpler continuity argument shows that choosing
	\begin{equation}
	\gamma_1 = \frac{C_{\eta}^*(2) }{16}
	\end{equation}
	yields the desired continuity. We use the inequality~\eqref{eq:pcaderivbd} to bound the order of the covering number $m_2$ for $G(D,d)$
	\begin{align}
	m_2 &= O\left( \eta^{-d(D-d)} d^{2d(D-d)} \right).
	\end{align}
	\qed
\end{proof}

%%%%

\subsection{Proof of Theorem~\ref{thm:K2d1}}\label{sec:K2d1:proof}

The proof of Theorem~\ref{thm:K2d1} follows from the propositions in this section. We assume that $0 < p \leq 1$ in the following discussion. It is useful to define a set $\mathcal{M}_{\eta}$ that is used frequently in the following:
\begin{equation}
\mathcal{M}_{\eta} = \{L \in G(D,1) : \eta \leq\dist(L,L_1^*) \leq \min(\dist(L,L_2^*),\pi/6) \}.
\end{equation}
For a fixed $L_0$, we compare the global minimum of $H_{p,\delta}(L,L_0;\cX)$ to the global minimum of $H_{p,\delta}^*(L,L_0;\mu)$, which will characterize the FMS$_p$ sequence.
\begin{proposition}
	Let $\cX$ be sampled independently and identically from the mixture measure given in~\eqref{mixtmeas1} with $K=2$ and $d=1$.  Then, for any $\eta > 0$, the weighted PCA solution to $\argmin_{L \in G(D,d)} H_{p,\delta}(L,L_0;\cX)$ lies in an $\eta$-neighborhood of $\argmin_{L \in G(D,d)} H_{p,\delta}^*(L,L_0;\mu)$ w.o.p.~$1-C_1 e^{-C_2 N}$, for some constants $C_1$ and $C_2$.
	\label{prop:globconv:H}
\end{proposition}
We omit the proof of this proposition since it is essentially the proof of Theorem~\ref{thm:globconv} for $p=2$ with a reweighted measure. We continue with a proposition on the derivative of $F_{p,\delta}^*$. To simplify things, let $m(\eta,p\delta) = \max(\eta,\arcsin((p\delta)^{1/(2-p)}))$.

\begin{proposition}\label{prop:derivMdK2}
	The derivative of $F_{p,\delta}^*$ is negative towards $L_1^*$ for all $L_0 \in \mathcal{M}_{m(\eta,p\delta)}$.
\end{proposition}

\begin{proof}
	We examine the derivative of $F_{p,\delta}^*$ towards $L_1^*$ given a subspace $L(0) \in \mathcal{M}_{m(\eta,p\delta)}$, which will prove the proposition. Define a vector $\bv$ such that $L(0) = \Sp(\bv)$ and let $L(1) = L_1^*$. Also, let $\bx^1$ be a vector spanning $L_1^*$ and $\bx^2$ a vector spanning $L_2^*$ such that $\angle(\bx^1,\bx^2) \leq \pi/2$, and $\angle(\bx^1,\bv) \leq \pi/2$. The derivative of $F_{p,\delta}^*$ with this geodesic simplifies to
	\begin{align}
	\frac{d}{dt} F_{p,\delta}^* (L(t);\mu) \Big|_{t=0} &= -\theta \left[\alpha_1 \frac{(\bv \cdot \bx^1)(\bu \cdot \bx^1)}{(\bu \cdot \bx^1)^{2-p}}  + \alpha_2 \frac{(\bv \cdot \bx^2)(\bu \cdot \bx^2)}{\max(\dist(\bx^2,L(0))^{2-p},p\delta)} \right] \\ \nonumber
	&\leq \theta \left[- \alpha_1 \frac{(\bv \cdot \bx^1)}{(\bu \cdot \bx^1)^{1-p}}  + \left|\alpha_2 \frac{(\bv \cdot \bx^2)(\bu \cdot \bx^2)}{\max(\dist(\bx^2,L(0))^{2-p},p\delta)} \right| \right].
	\end{align}
	Also, let $\bx^3 \in \Sp(\bu,\bv)$ be a point such that $\angle(\bx^3,\bu) = \angle(\bx^2,\bu)$. Then,
	\begin{align}\label{eq:derivK2d1bd}
	\frac{d}{dt} F_{p,\delta}^* (L(t);\mu) \Big|_{t=0} &\leq \theta \left[- \alpha_1 \frac{(\bv \cdot \bx^1)}{(\bu \cdot \bx^1)^{1-p}}  + \left|\alpha_2 \frac{(\bv \cdot \bx^2)(\bu \cdot \bx^3)}{(\bu \cdot \bx^3)^{2-p}} \right| \right] \\ \nonumber
	&= \theta \left[- \alpha_1 \frac{(\bv \cdot \bx^1)}{(\bu \cdot \bx^1)^{1-p}}  + \left|\alpha_2 \frac{(\bv \cdot \bx^2)}{(\bu \cdot \bx^3)^{1-p}} \right| \right] \\ \nonumber
	&\leq \theta \left[- \alpha_1 \frac{(\bv \cdot \bx^1)}{(\bu \cdot \bx^1)^{1-p}}  + \left|\alpha_2 \frac{(\bv \cdot \bx^2)}{(\bu \cdot \bx^1)^{1-p}} \right| \right].
	\end{align}
	This implies that $\frac{d}{dt} F_{p,\delta}^* (L(t);\mu) \Big|_{t=0}$ is negative when $|\bv \cdot \bx^1| > \alpha_2 / \alpha_1 |\bv \cdot \bx^2| $. This is guaranteed since $\alpha_1 > \alpha_2$ in~\eqref{mixtmeas1}, and $|\bv \cdot \bx^1| \geq |\bv \cdot \bx^2| $ when $L(0) \in \cM_{m(\eta,p\delta)}$.
	\qed
\end{proof}

\begin{proposition}\label{prop:asymstepK2}
	For each $L_0$ lying on the geodesic between $L_2^*$ and $L_1^*$ such that $\dist(L_0,L_1^*) < \dist(L_0,L_2^*)$, $L_1 = FMS_p(L_0)$ (the next FMS$_p$ iterate from $L_0$) lies closer to $L_1^*$ than $L_0$ w.o.p.
\end{proposition}

\begin{proof}
	Let $L_0$ be a point on the geodesic between $L_2^*$ and $L_1^*$ such that $\dist(L_0,L_1^*) < \dist(L_0,L_2^*)$. We note that there are two geodesics between $L_2^*$ and $L_1^*$, one which has length less than or equal to $\pi/2$ and one that has length greater than or equal to $\pi/2$. For this proof, $L_0$ can lie on either of these geodesics. Let $\cU_{L_0,p,\delta} = \{ \bx \in S^{D-1} : \dist^{2-p}(\bx,L_0) < p\delta \}$. For two subspaces $L_0$ and $L$, we can write the asymptotic majorizing function $H_{p,\delta}^*$ (corresponding to $H_{p,\delta}$ given in~\eqref{majfunc:prop}) under the mixture measure as
	\begin{align}
	H_{p,\delta}^*(L,L_0;\mu)  &= \int_{S^{D-1} \setminus \cU_{L_0,p,\delta}} \left(\frac{p}{2} \frac{\dist(\boldsymbol{x},L)^2}{\dist(\boldsymbol{x},L_0)^{2-p}} + \left(1-\frac{p}{2}\right)\dist(\boldsymbol{x},L_0)^{p} \right) d\mu(\bx)+ \\ \nonumber
	&\int_{\cU_{L_0,p,\delta}}\left(\frac{\dist^2(\bx,L)}{2\delta}  + (p\delta)^{p/(2-p)} - \frac{(p\delta)^{2/(2-p)}}{2\delta}\right)d\mu(\bx).
	\label{}
	\end{align}
	Now define the point $L_1 = \argmin_{L \in G(D,1)} H_{p,\delta}^*(L,L_0;\mu)$. Let $\bx^0$ be a basis vector for $L_0$, $\bx^1$ be a basis vector for $L_1^*$, and $\bx^2$ a basis vector for $L_2^*$ such that $\angle(\bx^1,\bx^2) \leq \pi/2$, $\angle(\bx^0,\bx^1) \leq \pi/2$, and $\angle(\bx^0,\bx^2) \leq \pi/2$. Differentiating the function $H_{p,\delta}^*$ with respect to its first argument along the geodesic $L(t)$, with $L(0) = L_0$ and $L(1)=L_1^*$, we get
	\begin{align}\label{eq:proof:K2d1:Hder}
	\frac{d}{dt} H_{p,\delta}^* (L(t),L_0;\mu) \Big|_{t=0} &= -\alpha_1 \theta \int_{L_1^*} \frac{(\bx^0 \cdot \bx) (\bu \cdot \bx)}{\max(\dist(\bx,L_0)^{2-p},p\delta)} d\mu_1 - \\ \nonumber
	&\ \ \ \ \ \ \ \alpha_2 \theta \int_{L_2^*} \frac{(\bx^0 \cdot \bx) (\bu \cdot \bx)}{\max(\dist(\bx,L_0)^{2-p},p\delta)} d\mu_2 \\ \nonumber
	&= -\theta \left[\alpha_1 \frac{(\bx^0 \cdot \bx^1)(\bu \cdot \bx^1)}{\dist(\bx^1,L_0)^{2-p}}  + \alpha_2 \frac{(\bx^0 \cdot \bx^2)(\bu \cdot \bx^2)}{\dist(\bx^2,L_0)^{2-p}} \right] \\ \nonumber
	&= -\theta \left[\alpha_1 \frac{(\bx^0 \cdot \bx^1)}{(\bu \cdot \bx^1)^{1-p}}  + \alpha_2 \frac{(\bx^0 \cdot \bx^2)}{(\bu \cdot \bx^2)^{1-p}} \right] < 0.
	\end{align}
	Since the derivative~\eqref{eq:proof:K2d1:Hder} is negative, the stationary point along $L(t)$  must be closer to $L_1^*$ than $L_0$. It is apparent that this stationary point is also the global minimum of the function $H_{p,\delta}^*(L,L_0;\mu)$. By Proposition~\ref{prop:globconv:H}, the global minimum of $H_{p,\delta}(L,L_0;\cX)$ is arbitrarily close to the global minimum of $H_{p,\delta}^*(L,L_0;\mu)$ w.o.p., and therefore Proposition~\ref{prop:asymstepK2} is proved.
	\qed
\end{proof}

\begin{proposition}\label{prop:asymstepK2_2}
	For each $L_0 \in \mathcal{M}_{m(\eta,p\delta)}$, $L_1 = FMS_p(L_0)$ (the next FMS$_p$ iterate from $L_0$) is closer to $L_1^*$ than $L_0$ w.o.p.
\end{proposition}

\begin{proof}
	Let $L_0$ be a point in $\mathcal{M}_{m(\eta,p\delta)}$. We again consider the asymptotic majorization function $H_{p,\delta}^*(L,L_0;\mu)$. Let $L_0'$ be the point along the geodesic between $L_2^*$ and $L_1^*$ such that $\dist(L_0,L_1^*) = \dist(L_0',L_1^*)$ and $\dist(L_0',L_1^*) \leq \dist(L_0',L_2^*)$. This point always exists along one of the two geodesics between $L_2^*$ and $L_1^*$. By the previous proposition,
	\begin{equation}
	\dist\left(\argmin_{L\in G(D,d)} H_{p,\delta}^*(L,L_0';\mu),L_1^*\right) < \dist(L_0',L_1^*).
	\end{equation}
	Further, this implies that
	\begin{equation}
	\dist\left(\argmin_{L\in G(D,d)} H_{p,\delta}^*(L,L_0;\mu),L_1^*\right) < \dist\left(\argmin_{L\in G(D,d)} H_{p,\delta}^*(L,L_0';\mu),L_1^*\right).
	\end{equation}
	This is due to the fact that $\dist(L_0',L_1^*) = \dist(L_0,L_1^*)$, but $\dist(L_0,L_2^*)\geq \dist(L_0',L_2^*)$. Thus, for all $L_0 \in \mathcal{M}_{m(\eta,p\delta)}$,
    \begin{equation}\label{eq:distasymstep}
	\dist\left(\argmin_{L\in G(D,d)} H_{p,\delta}^*(L,L_0;\mu),L_1^*\right) < \dist(L_0,L_1^*).
	\end{equation}
    Again, by Proposition~\ref{prop:globconv:H} and~\eqref{eq:distasymstep}, we conclude Proposition~\ref{prop:asymstepK2_2}
	\begin{align}
	\argmin_{L\in G(D,d)} H_{p,\delta}(L,L_0;\cX) &\in B(\argmin_{L\in G(D,d)} H_{p,\delta}^*(L,L_0;\mu),\gamma), \\ \nonumber &\text{w.o.p.~for all $\gamma > 0$.}
	\end{align}
	\qed
\end{proof}

Based on Propositions~\ref{prop:derivMdK2},~\ref{prop:asymstepK2}, and~\ref{prop:asymstepK2_2}, we are able to construct the following functions: for each $L_0 \in \mathcal{M}_{m(\eta,p\delta)}$, let $L_0'$ be the subspace on a geodesic between $L_2^*$ and $L_1^*$ such that $\dist(L_0,L_1^*) = \dist(L_0',L_1^*)$ and $\dist(L_0',L_1^*) \leq \dist(L_0',L_2^*)$. Define $\phi^*: \mathcal{M}_{m(\eta,p\delta)} \to (0,\infty)$ and $\phi: \mathcal{M}_{m(\eta,p\delta)} \to (0,\infty)$ by
\begin{align}
\phi^*(L_0;\mu) = \dist(L_1^*,L_0) - \dist(L_1^*,\argmin_{L\in G(D,d)} H_{p,\delta}^*(L,L_0;\mu)), \\
\phi(L_0;\cX) = \dist(L_1^*,L_0) - \dist(L_1^*,\argmin_{L\in G(D,d)} H_{p,\delta}(L,L_0;\cX)).
\end{align}
By compactness of $\mathcal{M}_{m(\eta,p\delta)}$ and the previous propositions, $\min_{L \in \mathcal{M}_{m(\eta,p\delta)}}\phi^*(L;\mu) = c_\delta > 0$.

\begin{proposition}\label{prop:radiusK2}
	For all $L_0 \in \mathcal{M}_{m(\eta,p\delta)}$, there exists a function $R: \mathcal{M}_{m(\eta,p\delta)} \to (0,\infty)$ such that for all $L' \in B(L_0,R(L_0))$,
	\begin{equation}
        \dist(L_1^*,\argmin_{L\in G(D,d)} H_{p,\delta}(L,L';\cX)) < \dist(L_1^*,L'), \ \text{w.o.p.}
	\end{equation}
\end{proposition}

\begin{proof}
	First, by definition of the function $\phi(L_0;\cX)$, Hoeffding's inequality with the random variable $\phi(\bx;L_0)$ implies
	\begin{equation}\label{eq:K2d1phihoeff}
		\phi(L_0;\cX) \geq c_\delta / 2, \ \text{w.p. $1-C_1 e^{-NC_2}$}.
	\end{equation}
	Next, we note that the function $\phi(\cdot;\cX)$ is continuous with respect to its argument. Thus, for any given $L_0$, there exists a number $\zeta_{L_0}$ such that for all $L' \in B(L_0,\zeta_{L_0})$,
	\begin{equation}\label{eq:K2d1Hradius}
	\left| \phi(L_0;\cX) - \phi(L';\cX)\right| \leq c_\delta/4.
	\end{equation}
	Thus, we define the function $R$ to be $R(L_0) = \zeta_{L_0}$. Then, for all $L' \in B(L_0,\zeta_{L_0})$, combining~\eqref{eq:K2d1phihoeff} and~\eqref{eq:K2d1Hradius}, we can conclude that for any $L_0 \in \mathcal{M}_{m(\eta,p\delta)}$ and all $L' \in B(L_0,R(L_0))$, $\dist(L_1^*,\argmin_{L\in G(D,d)} H_{p,\delta}(L,L' );\cX) < \dist(L_1^*,L' )$ w.o.p.
	\qed
\end{proof}

Thus, we can put this all together and finish the proof of Theorem~\ref{thm:K2d1}. Assume that we are given a data set $\cX$ sampled i.i.d.~from the mixture measure~\eqref{mixtmeas1} with $K=2$ and $d=1$, and a number $\eta > 0$. By Proposition~\ref{prop:radiusK2}, we can cover $\mathcal{M}_{m(\eta,p\delta)}$ by $\{B(L_0,\zeta_{L_0}):L_0 \in \mathcal{M}_{m(\eta,p\delta)}\}$. The next FMS$_p$ iterate for each point in each ball is closer to $L_1^*$ w.o.p.~This cover has a finite sub-cover by compactness of $\mathcal{M}_{m(\eta,p\delta)}$, and thus there are no fixed points in $\mathcal{M}_{m(\eta,p\delta)}$ w.o.p.~Finally, due to the fact that the iterates get closer to $L_1^*$, we get that FMS$_p$ must converge to a point in $\overline{B(L_1^*,\max(\eta,\arcsin((p\delta)^{1/(2-p)})))}$ w.o.p.

\subsection{Proof of Theorem~\ref{thm:sublinconv}}
\label{sec:proof:sublinconv}

Denote the FMS$_p$ sequence by $(L_k)_{k \in \nats}$, and assume that $\cX$ is sampled i.i.d.~from the mixture measure~\eqref{mixtmeas1} with $K=1$. Let $L_k^*(t):[0,1] \to G(D,d)$ denote the extended geodesic from $L_k$ in the direction of $L_1^*$, and let $s_k^*$ be the length of this geodesic (i.e. $s_k^* = \dist(L_k^*(0),L_k^*(1))$). We begin by reminding ourselves that
\begin{align}
\frac{d}{dt} H_{p,\delta} (L_k^*(t),L_k;\cX) \Big|_{t=0} &= \frac{d}{dt} F_{p,\delta} (L_k^*(t);\cX) \Big|_{t=0} \\ \nonumber
&= \sum_{i=1}^N -p \frac{\sum_{j=1}^d \left(\theta_j \pi/2 \theta_1\right) (\bv_j \cdot \bx_i)(\bu_j \cdot \bx_i)}{\max(\dist^{2-p}(\bx_i,L_k),p\delta)}.
\end{align}
At a point $\hat{t} \in (0,1)$, we can instead write the derivative of $H_{p,\delta}(L_k^*(t),L_k;\cX)$ as
\begin{align}
\frac{d}{dt} H_{p,\delta} (L_k^*(t),L_k;\cX) \Big|_{t=\hat{t}} &= \sum_{i=1}^N -p \frac{\sum_{j=1}^d \left(\theta_j \pi/2 \theta_1\right) (\bv_j' \cdot \bx_i)(\bu_j' \cdot \bx_i)}{\max(\dist^{2-p}(\bx_i,L_k),p\delta)}
\end{align}
for a new set of basis vectors $\bv_j'$ and $\bu_j'$. Continuity of the derivative of $H_{p,\delta}$ with respect to $t$ implies
\begin{align}
&\left|\frac{d}{dt} H_{p,\delta} (L_k^*(t),L_k;\cX) \Big|_{t=0} - \frac{d}{dt} H_{p,\delta} (L_k^*(t),L_k;\cX) \Big|_{t=0} \right|  \\ \nonumber
&=\left| \sum_{i=1}^N -p \frac{\sum_{j=1}^d \left(\theta_j \pi/2 \theta_1\right) (\bv_j \cdot \bx_i)(\bu_j \cdot \bx_i)}{\max(\dist^{2-p}(\bx_i,L_k),p\delta)} -  \sum_{i=1}^N -p \frac{\sum_{j=1}^d \left(\theta_j \pi/2 \theta_1\right) (\bv_j' \cdot \bx_i)(\bu_j' \cdot \bx_i)}{\max(\dist^{2-p}(\bx_i,L_k),p\delta)} \right| \\ \nonumber
&=\left| -p \sum_{i=1}^N \sum_{j=1}^d \left(\theta_j \pi/2 \theta_1\right) \frac{  (\bv_j \cdot \bx_i)(\bu_j \cdot \bx_i) - (\bv_j' \cdot \bx_i)(\bu_j' \cdot \bx_i)}{\max(\dist^{2-p}(\bx_i,L_k),p\delta)} \right| \\ \nonumber
&= \frac{\pi}{2 \delta} \left|  \sum_{i=1}^N \sum_{j=1}^d  (\bv_j \cdot \bx_i)(\bu_j \cdot \bx_i) - (\bv_j' \cdot \bx_i)(\bu_j' \cdot \bx_i) \right|\\ \nonumber
&\leq \frac{\pi}{2 \delta}  \sum_{i=1}^N \sum_{j=1}^d  \left| (\bv_j \cdot \bx_i)(\bu_j \cdot \bx_i) - (\bv_j' \cdot \bx_i)(\bu_j' \cdot \bx_i) \right|.
\end{align}
If $\dist(L(0),L(\hat{t}) < C_\eta^*(p) p\delta/8$, then
\begin{equation}
\frac{1}{N}\left|\frac{d}{dt} H_{p,\delta} (L_k^*(t),L_k;\cX) \Big|_{t=0} - \frac{d}{dt} H_{p,\delta} (L_k^*(t),L_k;\cX) \Big|_{t=0} \right| < \frac{C_{\eta}^*(p)}{8}
\end{equation}

The first order Taylor expansion of $H_{p,\delta}(\cdot,L_k;\cX)$ at $L_k$ in the direction of $L_1^*$ is given by
\begin{equation}\label{eq:taylor1H}
H_{p,\delta}(L_k^*(t),L_k;\cX) = F_{p,\delta}(L_k;\cX) + t s_k^* \frac{d}{du} H_{p,\delta}(L_k^*(u),L_k;\cX) \Big|_{u=\hat{t}}
\end{equation}
for some $\hat{t} \in (0,t)$. Define the quantity $\lambda_k$ as
\begin{equation}
\lambda_k = \frac{C_\eta^*(p) p\delta}{8} \leq 1.
\end{equation}
The inequality $\lambda_k < 1$ follows from a simple estimate for $C_\eta^*(p)$. The first order Taylor expansion~\eqref{eq:taylor1H} evaluated at $t=\lambda_k$ is
\begin{align}\label{eq:1ordtaylorHlambdak}
H_{p,\delta}(L_k^*(\lambda_k),L_k;\cX) &= F_{p,\delta}(L_k;\cX) + \lambda_k s_k^*  \frac{d}{du} H_{p,\delta}(L_k^*(u),L_k;\cX) \Big|_{u=\hat{t}},
\end{align}
for some $\hat{t} \in (0,\lambda_k)$. Using~\eqref{prop:hyp1},~\eqref{iterates}, and~\eqref{eq:1ordtaylorHlambdak}, we conclude that
\begin{align} \label{eq:costdecrbound}
F_{p,\delta}(L_{k+1};\cX) &\leq H_{p,\delta}(L_{k+1},L_k;\cX) \leq H_{p,\delta}(L_k^*(\lambda_k),L_k;\cX) \\ \nonumber
&= F_{p,\delta}(L_k;\cX) + \lambda_k s_k^* \frac{d}{du} H_{p,\delta}(L_k^*(u),L_k;\cX) \Big|_{u=\hat{t}} \\ \nonumber
&\leq F_{p,\delta}(L_k;\cX) + \frac{C_\eta^*(p) p\delta}{8} \frac{d}{du} H_{p,\delta}(L_k^*(u),L_k;\cX) \Big|_{u=\hat{t}} \\ \nonumber
&= F_{p,\delta}(L_k;\cX) + \frac{C_\eta^*(p) p\delta}{8} \frac{C_\eta^*(p)}{8}.
\end{align}
We now split into different cases by $p$. For $0<p\leq 1$, $C_\eta^*(p) > O\left(\min\left( \left(\frac{\pi}{6}\right) ^{p-1},\frac{\eta}{p\delta}\right)\right)$ by~\eqref{eq:fmsderivbd}, which implies
\begin{align}\label{eq:sublinordF0p1}
F_{p,\delta}(L_{k};\cX) - F_{p,\delta}(L_{k+1};\cX) > O\left(\min\left( \left(\frac{\pi}{6}\right) ^{2(p-1)} p\delta ,\frac{\eta^2}{(p\delta)}\right)\right).
\end{align}
For $1 < p < 2$, $C_\eta^*(p) = O\left(\min\left(\eta^{p-1},\frac{\eta}{p\delta}\right)\right)$ by~\eqref{eq:fmsderivbd2}, which implies
\begin{align}\label{eq:sublinordF1p2}
F_{p,\delta}(L_{k};\cX) - F_{p,\delta}(L_{k+1};\cX) >O\left(\min\left(\eta^{2(p-1)} p\delta,\frac{\eta^2}{(p\delta)}\right)\right).
\end{align}

Thus, by~\eqref{eq:sublinordF0p1} for $0<p\leq1$
\begin{align}
T > O\left(\frac{1}{\min\left( \left(\frac{\pi}{6}\right) ^{2(p-1)} p\delta ,\frac{\eta^2}{(p\delta)}\right)}\right) \implies \dist(L_T,L_1^*) < \eta \ \text{(w.o.p.)}.
\end{align}
This follows from the fact that the cost cannot be negative. From this, the global convergence bound is concluded.
On the other hand, by~\eqref{eq:sublinordF1p2} for $1<p<2$
\begin{align}
T > O\left(\frac{1}{\min\left(\eta^{2(p-1)} p\delta,\frac{\eta^2}{(p\delta)}\right)}\right) \implies \dist(L_T,L_1^*) < \eta \ \text{(w.o.p.)}.
\end{align}
Again, the global convergence bound is concluded. 

A similar proof can be done for the case $K=2$ and $d=1$. The only difference now is that the constant $C_\eta^*(p)$ has a new bound. In this case, we bound the magnitude of the derivative of $F_{p,\delta}^*$ over the set $\mathcal{M}_{m(\eta,p\delta)}$. Let $L(t)$ be the extended geodesic between a point $L(0) \in \mathcal{M}_{m(\eta,p\delta)}$ and $L_1^*$. From~\eqref{eq:derivK2d1bd}, we get the following bound for $0<p\leq 1$:
\begin{equation}
\min_{L(0) \in \mathcal{M}_{m(\eta,p\delta)}} \left|\frac{d}{dt} F_{p,\delta}^* (L(t))\right| > \frac{\pi}{2} \frac{1}{2^{p/2}} (\alpha_1 - \alpha_2).
\end{equation}
In other words, for $0 < p \leq 1$, we now have $C_\eta^*(p) > \frac{\pi}{2} \frac{1}{2^{p/2}} (\alpha_1 - \alpha_2)$. This means that
\begin{align}\label{eq:sublinordF0p1K2d1}
F_{p,\delta}(L_{k};\cX) - F_{p,\delta}(L_{k+1};\cX) > O\left( (\alpha_1 - \alpha_2)^2 p\delta \right).
\end{align}
Thus, by~\eqref{eq:sublinordF0p1K2d1} with $0<p\leq1$,
\begin{align}
T > O\left(\frac{1}{(\alpha_1 - \alpha_2)^2 p\delta}\right) \implies \dist(L_T,L_1^*) < \eta \ \text{(w.o.p.)}.
\end{align}
%Repeating the argument used to obtain~\eqref{eq:taylor1Fu2}, we get
%\begin{align}\label{eq:taylor1FuK2d1}
%\dist(L_k,B(L_1^*&,\max(\eta,\arcsin((p\delta)^{1/(2-p)})))) = u_k s_k^*  \\ \nonumber
%&= \frac{F_{p,\delta}(L_k;\cX) - F_{p,\delta}(L_k^*(u_k);\cX)}{ \left|\frac{d}{du} F_{p,\delta}(L_k^*(u);\cX) \Big|_{u=\hat{t}} \right|} \\ \nonumber
%&\leq \frac{F_{p,\delta}(L_k;\cX) - F_{p,\delta}(L_k^*(u_k);\cX)}{C_\eta^*(p) / 4} \\ \nonumber
%&\leq \frac{F_{p,\delta}(L_k;\cX) - F_{p,\delta}(L_k^*(u_k);\cX)}{O\left(\frac{(\alpha_1 - \alpha_2)^2}{2^{p}} \right)}.
%\end{align}
%Finally, we see that the global convergence rates of $F_{p,\delta}(L_k)$ to $B(F_{p,\delta}(L_k),\eta)$ then translate to global convergence rates for the sequence $(L_k)_{k\in \nats}$.

\subsection{Proof of Theorem~\ref{thm:linconv}}
\label{sec:proof:linconv}

In order for a $r$-linear rate of convergence proof for the FMS$_p$ iterates, we need strong geodesic convexity in a neighborhood of the limit point $L^*$ (or for global convergence, geodesic convexity). The following theorem shows that under the mixture measure~\eqref{mixtmeas1} with $K=1$, the FMS$_p$ algorithm is strongly geodesically convex around the global minimum w.o.p.~under a condition on $\alpha_0$ and $\alpha_1$.  Another consequence of this theorem is that $H_{p,\delta}(L,L_1^*;\cX)$ is strongly geodesically convex at $L_1^*$.
\begin{proposition}\label{prop:secder}
	Let $\cX$ be a data set sampled i.i.d.~from the mixture measure~\ref{mixtmeas1} with $K=1$, or $K=2$, $d=1$, $\alpha_1 > (2-p) \alpha_2$, and $\dist(L_1^*,L_2^*) > 2 \arcsin(p\delta^{1/(2-p)})$. Then, the second derivative of $F_{p,\delta}$ is positive in all geodesic directions at $L_1^*$ w.o.p.
\end{proposition}

\begin{proof}[Proof of Proposition~\ref{prop:secder}]
A useful fact for the asymptotic FMS$_p$ theory comes in the separability of the cost function with respect to the mixture measure
\begin{equation}
F_{p,\delta}^*(L;\mu) = \sum_{i=0}^K \alpha_i F_{p,\delta}^*(L;\mu_i).
\end{equation}
We note that $F_{p,\delta}^*(L;\mu_0)$ is constant with respect to $L$ due to the spherical symmetry of $\mu_0$, and therefore any geodesic derivative of this term is zero. If we parametrize a geodesic $L(t)$, $t \in [0,1]$, and take the derivative of $F_{p,\delta}^*(L;\mu)$ with respect to $t$, we find that
\begin{equation}
\frac{d}{dt} F_{p,\delta}^*(L(t);\mu) = \sum_{i=1}^K \alpha_i \frac{d}{dt} F_{p,\delta}^*(L(t);\mu_i).
\end{equation}
Further, for each $i$, the derivative of the cost function is
\begin{equation}
\frac{d}{dt} F_{p,\delta}^*(L(t);\mu_i) =  \int_{L_i^*} -p\frac{\sum\limits_{j=1}^{d} \theta_j((\cos(t\theta_j)\bv_j + \sin(t\theta_j)\bu_j)\cdot \bx)((-\sin(t\theta_j)\bv_j+\cos(t\theta_j)\bu_j)\cdot \bx)}{\max(\dist^{2-p}(\bx,L(t)),p\delta)} d\mu_i.
\end{equation}
Taking a further derivative, we find that
\begin{align}
\frac{d^2}{dt^2} F_{p,\delta}^*(L(t);\mu_i) &=  \int_{{L_i^*} \setminus \cU_{L(t),p,\delta}} -\frac{p}{(\dist^{2-p}(\bx,L(t)))^2} \Bigg[ \\ \nonumber
&\dist^{2-p}(\bx,L(t))\left(\sum_{j=1}^{d}\theta_j^2(-\sin(t\theta_j)\bv_j+\cos(t\theta_j)\bu_j)\cdot \bx)^2\right)-\\ \nonumber
&\dist^{2-p}(\bx,L(t))\left(\sum_{j=1}^{d}\theta_j^2((\cos(t\theta_j)\bv_j + \sin(t\theta_j)\bu_j)\cdot \bx)^2\right)-\\ \nonumber
&\left(\sum\limits_{j=1}^{d} \theta_j((\cos(t\theta_j)\bv_j + \sin(t\theta_j)\bu_j)\cdot \bx)((-\sin(t\theta_j)\bv_j+\cos(t\theta_j)\bu_j)\cdot \bx)\right)\cdot \\ \nonumber
&(2-p)\dist^{1-p}(\bx,L(t))\frac{d}{dt}\dist(\bx,L(t)) \Bigg] d\mu_i + \\ \nonumber
& \int_{L_i^* \cap \cU_{L(t),p,\delta}} \frac{-p}{p\delta} \Bigg[ \left(\sum_{j=1}^{d}\theta_j^2(-\sin(t\theta_j)\bv_j+\cos(t\theta_j)\bu_j)\cdot \bx)^2\right)-\\ \nonumber
&\left(\sum_{j=1}^{d}\theta_j^2((\cos(t\theta_j)\bv_j + \sin(t\theta_j)\bu_j)\cdot \bx)^2\right) \Bigg] d\mu_i .
\end{align}
The second derivative at $t=0$ is then
\begin{align}\label{eq:secondderiv:term}
\frac{d^2}{dt^2} F_{p,\delta}^*(L(t);\mu_i) \Big|_{t=0}  &= p \int_{L_i^*}  \frac{\sum_{j=1}^{d}\theta_j^2 \left((\bv_j \cdot \bx)^2 - (\bu_j \cdot \bx)^2\right)}{\max(\dist^{2-p}(\bx,L(0)),p\delta)} d\mu_i - \\ \nonumber
&p \int_{L_i^* \setminus \cU_{L(0),p,\delta}} \frac{2-p}{\dist(\bx,L(0))} \cdot \left(\frac{\sum_{j=1}^d \theta_j (\bv_j \cdot \bx)(\bu_j \cdot \bx)}{\dist(\bx,L(0))}\right)^2 d\mu_i.
\end{align}
From~\eqref{eq:secondderiv:term}, we can find the second derivative of the cost function with respect to the full mixture measure
\begin{align}\label{eq:secondderiv}
\frac{d^2}{dt^2} F_{p,\delta}^*(L(t);\mu) \Big|_{t=0} &= p \sum_{i=1}^K \alpha_i \int_{L_i^*}  \frac{\sum_{j=1}^{d}\theta_j^2 \left((\bv_j \cdot \bx)^2 - (\bu_j \cdot \bx)^2\right)}{\max(\dist^{2-p}(\bx,L(0)),p\delta)} d\mu_i - \\ \nonumber
&p \sum_{i=1}^K \alpha_i \int_{L_i^* \setminus \cU_{L(0),p,\delta}} \frac{2-p}{\dist(\bx,L(0))} \cdot \left(\frac{\sum_{j=1}^d \theta_j (\bv_j \cdot \bx)(\bu_j \cdot \bx)}{\dist(\bx,L(0))}\right)^2 d\mu_i.
\end{align}
In the case of $K=1$,~\eqref{eq:secondderiv} simplifies to
\begin{align}\label{eq:secondderiv:K1}
\frac{d^2}{dt^2} F_{p,\delta}^*(L(t);\mu) \Big|_{t=0} &= p \alpha_1 \int_{L_1^*}  \frac{\sum_{j=1}^{d}\theta_j^2 \left((\bv_j \cdot \bx)^2 - (\bu_j \cdot \bx)^2\right)}{\max(\dist^{2-p}(\bx,L(0)),p\delta)} d\mu_1 - \\ \nonumber
&p \alpha_1 \int_{L_1^* \setminus \cU_{L(0),p,\delta}} \frac{2-p}{\dist(\bx,L(0))} \cdot \left(\frac{\sum_{j=1}^d \theta_j (\bv_j \cdot \bx)(\bu_j \cdot \bx)}{\dist(\bx,L(0))}\right)^2 d\mu_1.
\end{align}
When $L(0) = L_1^*$, the second derivative is strictly positive, because $(\bu_j \cdot \bx) = 0$ for all $\bx \in L_1^*$.

%It is desirable to know how big the neighborhood of strict convexity is. The first term is strictly positive when $L(0) \in B(L_1^*,\pi/6)$. Further, if we assume $L(0) \in B(L_1^*,\pi/8)$, we have the following inequality
%\begin{align}
%\frac{d^2}{dt^2} F_{p,\delta}^*(L(t);\mu) \Big|_{t=0} &= p \alpha_1 \int_{L_1^*}  \frac{\sum_{j=1}^{d}\theta_j^2 \left((\bv_j \cdot \bx)^2 - (\bu_j \cdot \bx)^2\right)}{\max(\dist^{2-p}(\bx,L(0)),p\delta)} d\mu_1 - \\ \nonumber
%&p \alpha_1 \int_{L_1^* \setminus \cU_{L(0),p,\delta}} \frac{2-p}{\dist(\bx,L(0))} \cdot \left(\frac{\sum_{j=1}^d \theta_j (\bv_j \cdot \bx)(\bu_j \cdot \bx)}{\dist(\bx,L(0))}\right)^2 d\mu_1.
%\end{align}
%Thus, we find that when $L(0) = L_1^*$, $\frac{d^2}{dt^2} F_{p,\delta}^*(L(t);\mu) \Big|_{t=0} >0$.
%\begin{align}
%\frac{d^2}{dt^2} F_{p,\delta}^*(L(t);\mu) \Big|_{t=0}&\geq \frac{p \alpha_1}{2} \int_{L_1^*}  \frac{\sum_{j=1}^{d}\theta_j^2 \left((\bv_j \cdot \bx)^2 \right)}{\max(\dist^{2-p}(\bx,L(0)),p\delta)} d\mu_1 - \\ \nonumber
%&p \alpha_1 (2-p) \int_{L_1^* \setminus \cU_{L(0),p,\delta}} \frac{\left(\sum_{j=1}^d \theta_j (\bv_j \cdot \bx)\right)^2}{\dist(\bx,L(0))}  d\mu_1.
%\end{align}

In the case of $K=2$ and $d=1$, we have
\begin{align}\label{eq:secondderiv:K2d1}
\frac{d^2}{dt^2} F_{p,\delta}^*(L(t);\mu) \Big|_{t=0} &= p \sum_{i=1}^2 \alpha_i \int_{L_i^*}  \frac{\theta^2 \left((\bv \cdot \bx)^2 - (\bu \cdot \bx)^2\right)}{\max(\dist^{2-p}(\bx,L(0)),p\delta)} d\mu_i - \\ \nonumber
&p \sum_{i=1}^2 \alpha_i \int_{L_i^* \setminus \cU_{L(0),p,\delta}} \frac{2-p}{\dist(\bx,L(0))} \cdot \left(\frac{ \theta (\bv \cdot \bx)(\bu \cdot \bx)}{\dist(\bx,L(0))}\right)^2 d\mu_i.
\end{align}
When $L(0) = L_1^*$, letting $\bx^1$ be a basis vector for $L_1^*$ and $\bx^2$ a basis vector for $L_2^*$ such that $\angle (\bx^1,\bx^2) \leq \pi/2$, we can bound~\eqref{eq:secondderiv:K2d1} as follows:
\begin{align}\label{eq:secondderiv:K2d1bd}
\frac{d^2}{dt^2} F_{p,\delta}^*(L(t);\mu) \Big|_{t=0}
%= p  \alpha_1   \frac{\theta^2 (\bv \cdot \bx^1)^2 }{p\delta}  + \\ \nonumber
%& p \alpha_2  \frac{\theta^2 \left((\bv \cdot \bx^2)^2 - (\bu \cdot \bx^2)^2\right)}{\max(\dist^{2-p}(\bx^2,L_1^*),p\delta)} -\\ \nonumber
%&p \alpha_2 \frac{2-p}{\dist(\bx^2,L_1^*)} \cdot \left(\frac{ \theta (\bv \cdot \bx^2)(\bu \cdot \bx^2)}{\dist(\bx^2,L_1^*)}\right)^2  \\ \nonumber
&\geq p  \alpha_1   \frac{\theta^2 (\bv \cdot \bx^1)^2 }{p\delta}  + \\ \nonumber
& p \alpha_2  \frac{\theta^2 \left((\bv \cdot \bx^2)^2 - (\bu \cdot \bx^2)^2\right)}{\max(\dist^{2-p}(\bx^2,L_1^*),p\delta)} -\\ \nonumber
&p(2-p) \alpha_2 \frac{\theta^2 (\bv \cdot \bx^2)^2}{\dist(\bx^2,L_1^*)}  .
\end{align}
Thus, we must have a condition on $\alpha_1$ and $\alpha_2$ in order to have strong convexity at $L_1^*$. If $2\arcsin((p\delta)^{1/(2-p)}) <\dist(L_1^*,L_2^*) \leq \pi / 4$, then a sufficient condition for strong convexity at $L_1^*$ is $\alpha_1 \geq (2-p) \alpha_2$, since the second term is positive in this case. On the other hand, if $\dist(L_1^*,L_2^*) > \pi / 4$, then a sufficient condition is $\alpha_1 \geq \alpha_2 p\delta 4 / \pi (3-p)$, which is true for all $\alpha_1 > \alpha_2$ when $\delta$ is sufficiently small ($\delta < \pi / (4p(3-p))$).

Finally, for $K=1$ in~\eqref{mixtmeas1} (or $K=2,d=1$), the second derivative of $F_{p,\delta}^*$ is continuous. To finish the proof the proposition, we let $b_{\alpha_1}$ be the minimum second derivative of $F_{p,\delta}^*$ across all directions. By~\eqref{eq:secondderiv:K1} (and~\eqref{eq:secondderiv:K2d1bd} with $\alpha_1 > \alpha_2 (2-p)$), $b_{\alpha_1} > 0$. For each directional derivative along a geodesic $L(t)$,
\begin{equation}
\frac{d^2}{dt^2} F_{p,\delta}(L(t);\cX) \Big|_{t=0} > \frac{b_{\alpha_1}}{2}, \text{w.p. $1 - e^{-N C_1} $,}
\end{equation}
for some constant $C_1$. Further, by continuity of the second derivative of $F_{p,\delta}$, there exists a number $\xi_{L(1)}$ such that for another geodesic $L'(t)$ with $L'(0) = L_1^*$ and $\dist(L(1),L'(1)) < \xi_{L(1)}$,
\begin{equation}
\left|\frac{d^2}{dt^2} F_{p,\delta}(L(t);\cX) \Big|_{t=0} - \frac{d^2}{dt^2} F_{p,\delta}(L'(t);\cX) \Big|_{t=0} \right| < \frac{b_{\alpha_1}}{2}.
\end{equation}
By another covering argument, for a data set sampled i.i.d.~from~\eqref{mixtmeas1} with $K=1$ (or $K=2,d=1$) the second derivative of $F_{p,\delta}$ is bounded away from zero w.o.p.
\qed
\end{proof}

By Proposition~\ref{prop:secder}, the second derivative of $F_{p,\delta}$ at $L_1^*$ is positive w.o.p.~for data sets sampled i.i.d from~\eqref{mixtmeas1} with $K=1$ (or $K=2$, $d=1$, $\alpha_1 > \alpha_2(2-p)$ and $\dist(L_1^*,L_2^*)> 2 \arcsin(p\delta^{1/(2-p)})$). Due to the fact that in the case $K=2$ and $d=1$, we cannot guarantee an $\eta$-approximation to $L_1^*$ for any $\eta$ (it is capped at $\arcsin((p\delta)^{1/(2-p)})$), we must be sure that Proposition~\eqref{prop:secder} can be extended to strong geodesic convexity at the limit point of the FMS sequence. This can be guaranteed if we set $\eta = \arcsin((p\delta)^{1/(2-p)})$, let $L^*$ be the limit point of FMS that is within $B(L_1^*,\eta)$, and notice that a modified version of~\eqref{eq:secondderiv:K2d1bd} is still positive at such an $L^*$ w.o.p.

By continuity of the second derivative, $F_{p,\delta}^*$ is strongly geodesically convex in a neighborhood of $L_1^*$ w.o.p.~Further, strong geodesic convexity of $F_{p,\delta}(L;\cX)$ at $L_k$ implies strong geodesic convexity of $H_{p,\delta}(L,L_k;\cX)$ at $L_k$ since $H_{p,\delta}$ majorizes $F_{p,\delta}$.

Let $L^*$ be the true limit point of the FMS$_p$ algorithm. We have strong geodesic convexity at $L^*$ w.o.p.~by the previous argument, and further there exists $\kappa>0$ such that for $k>\kappa$, all geodesics between $L^*$ and $L_k$ are strongly convex. Let $L_k^*(t)$ denote the geodesic from $L_k$ to $L^*$ for $t \in [0,1]$, and $s_k^* = \dist(L_k,L^*)$. For $k>\kappa$, by Taylor's Theorem we can write for some $\hat{t}_k \in (0,1)$
\begin{align}\label{eq:Htaylorexp}
H_{p,\delta}(L_k^*(t),L_k;\cX) &= F_{p,\delta}(L_k;\cX) + t s_k^* \frac{d}{dt} F_{p,\delta}(L_k^*(t);\cX) \Big|_{t=0} + \frac{1}{2} (ts_k^*)^2 \frac{d^2}{du^2} H_{p,\delta}(L_k^*(u),L_k;\cX) \Big|_{u=\hat{t}_k} \\ \nonumber
& = F_{p,\delta}(L_k;\cX) + t s_k^* \frac{d}{dt} F_{p,\delta}(L_k^*(t);\cX) \Big|_{t=0} + \frac{1}{2} (ts_k^*)^2 C(L_k^*(t)) ,
\end{align}
where $C(L_k^*(t))$ is strictly positive function depending on $L_k^*(t)$. We now follow the proof of~\citet{Chan99} for $r$-linear convergence of the generalized Weiszfeld method with some slight twists. We define a further majorization function for $H_{p,\delta}(L_k^*(t),L_k;\cX)$ as $\wH_k (t)$ for $t \in [0,1]$ by
\begin{equation}\label{eq:1ordtaylorHmaj}
\wH_k (t) = F_{p,\delta}(L_k;\cX) + t s_k^* \frac{d}{dt} F_{p,\delta}(L_k^*(t);\cX) \Big|_{t=0} + \frac{1}{2} (ts_k^*)^2 C(L_k),
\end{equation}
where $C(L_k) = \max_{t \in [0,1]} C(L_k^*(t))$, which is defined in~\eqref{eq:Htaylorexp}.

Define $\lambda_k$ as
\begin{equation}\label{eq:lambdak}
\lambda_k := \frac{\wH_k (1) - F_{p,\delta}(L^*;\cX)}{\frac{1}{2} (s_k^*)^2 C(L_k)}.
\end{equation}
Then, from~\eqref{eq:Htaylorexp} and~\eqref{eq:lambdak}, we find that
\begin{align}
F_{p,\delta}(L_{k+1};\cX) &\leq H_{p,\delta}(L_{k+1},L_k;\cX) \leq H_{p,\delta}(L_{k}^*(1-\lambda_k),L_k;\cX) \leq \wH(1-\lambda_k) \\ \nonumber
&= F_{p,\delta}(L_k;\cX) + (1-\lambda_k) s_k^* \frac{d}{dt} F_{p,\delta}(L_k^*(t);\cX) \Big|_{t=0} + \frac{1}{2} (1-\lambda_k)^2 s_k^{*2} C(L_k) \\ \nonumber
&= F_{p,\delta}(L_k;\cX) + (1-\lambda_k) \Big[s_k^{*} \frac{d}{dt} F_{p,\delta}(L_k^*(t);\cX) \Big|_{t=0} + \\ \nonumber
& \ \ \ \ \ \frac{1}{2} s_k^{*2} \left(1-\frac{\wH_k(1) - F_{p,\delta}(L^*;\cX)}{\frac{1}{2} s_k^{*2} C(L_k)}\right) C(L_k) \Big] \\ \nonumber
&= F_{p,\delta}(L_k;\cX) + (1-\lambda_k)\left[F_{p,\delta}(L^*;\cX) - F_{p,\delta}(L_k;\cX)\right].
\end{align}
Rearranging this equation then yields
\begin{equation}\label{eq:rateconvF}
F_{p,\delta}(L_{k+1};\cX) - F_{p,\delta}(L^*;\cX) \leq \lambda_k (F_{p,\delta}(L_k;\cX) - F_{p,\delta}(L^*;\cX)).
\end{equation}
Thus, if we can prove that the $\lambda_k$ are strictly bounded below 1, then~\eqref{eq:rateconvF} gives linear convergence of the cost iterates $(F_{p,\delta}(L_k;\cX))_{k\in \nats}$. First, we can write the first order Taylor expansion of $F_{p,\delta}$ at $L_k$ towards $L^*$ as
\begin{equation}\label{eq:1ordtaylorF}
F_{p,\delta}(L_k^*(t),L_k;\cX) = F_{p,\delta}(L_k;\cX) + t s_k^* \frac{d}{dt} F_{p,\delta}(L_k^*(t);\cX) \Big|_{t=0} + \frac{1}{2} (ts_k^*)^2 \frac{d^2}{du^2} F_{p,\delta}(L_k^*(u);\cX) \Big|_{u=\hat{t}_k}.
\end{equation}
Combining~\eqref{eq:1ordtaylorHmaj},~\eqref{eq:lambdak}, and~\eqref{eq:1ordtaylorF},
\begin{equation}
\lambda_k = \frac{\wH_k (1) - F_{p,\delta}(L^*;\cX)}{\frac{1}{2} s_k^{*2} C(L_k)} = 1 - C^{-1}(L_k) \frac{d^2}{dt^2} F_{p,\delta}(L_k^*(t);\cX) \Big|_{t=\hat{t}_k}.
\end{equation}
Here, by the strong convexity of $F_{p,\delta}$ along geodesics between $L^*$ and $L_k$, we can write
\begin{align}\label{eq:lambdakbound}
\lambda_k &=  1 - C^{-1}(L_k) \frac{d^2}{dt^2} F_{p,\delta}(L_k^*(t);\cX) \Big|_{t=\hat{t}_k} \\ \nonumber
&\leq \Lambda := 1 - \inf_k C^{-1}(L_k) \inf_k \frac{d^2}{dt^2} F_{p,\delta}(L_k^*(t);\cX) \Big|_{t=\hat{t}_k} <1.
\end{align}
The strict inequality in~\eqref{eq:lambdakbound} comes from compactness of the set $(L_k)_{k > \kappa} \cup \{L^*\}$ and strong geodesic convexity of $F_{p,\delta}$ at all $L \in \left((L_k)_{k>\kappa} \cup \{L^*\}\right)$. We also know that $\lambda_k > 0$ from~\eqref{eq:lambdak}, since $\wH_k(t) \geq H_{p,\delta}(L_k^*(t),L_k;\cX) \geq F_{p,\delta}(L_k^*(t);\cX) $ for all $t \in [0,1]$.

Finally, let $L_*^k(t)$ denote the geodesic from $L^*$ to $L_k$. The Taylor expansion of $F_{p,\delta}$ at $L^*$ towards $L_k$ is given by
\begin{equation}\label{eq:1ordtaylorFLstar}
F_{p,\delta}(L_*^k(t);\cX) = F_{p,\delta}(L^*;\cX) + \frac{1}{2} (ts_k^*)^2 \frac{d^2}{du^2} F_{p,\delta}(L_*^k(u);\cX) \Big|_{u=\tilde{t}_k}
\end{equation}
for some $\tilde{t}_k \in (0,t)$. Using~\eqref{eq:1ordtaylorFLstar} evaluated at $t=1$, with the corresponding $\tilde{t}_k \in (0,1)$, results in the estimate
\begin{align}\label{eq:taylorFdeltaLstar}
F_{p,\delta}(L_k;\cX) - F_{p,\delta}(L^*;\cX) &= \frac{1}{2} s_k^{*2} \frac{d^2}{dt^2} F_{p,\delta}(L_*^k(t);\cX) \Big|_{t=\tilde{t}_k} \\ \nonumber
& \geq \frac{1}{2} s_k^{*2} \inf_k \frac{d^2}{dt^2} F_{p,\delta}(L_*^k(t);\cX) \Big|_{t=\tilde{t}_k} > 0.
\end{align}
We rewrite~\eqref{eq:taylorFdeltaLstar} and define $y_k$ as the quantity
\begin{equation}\label{eq:skstarbound}
s_k^{*} \leq y_k := \sqrt{2 \frac{F_{p,\delta}(L_k;\cX) - F_{p,\delta}(L^*;\cX)}{\inf_k \frac{d^2}{dt^2} F_{p,\delta}(L_*^k(t);\cX) \Big|_{t=\tilde{t}_k} }}.
\end{equation}
Combining~\eqref{eq:rateconvF},~\eqref{eq:lambdakbound} and~\eqref{eq:skstarbound} then yields
\begin{equation}\label{eq:linearconv}
y_{k+1}^2 = 2 \frac{F_{p,\delta}(L_{k+1};\cX) - F_{p,\delta}(L^*;\cX)}{\inf_k \frac{d^2}{dt^2} F_{p,\delta}(L_k^*(t);\cX) \Big|_{t=\hat{t}_k} } \leq  2 \Lambda \frac{F_{p,\delta}(L_{k};\cX) - F_{p,\delta}(L^*;\cX)}{\inf_k \frac{d^2}{dt^2} F_{p,\delta}(L_k^*(t);\cX) \Big|_{t=\hat{t}_k} } = \Lambda y_k^2.
\end{equation}
Therefore, $y_{k+1} \leq \sqrt{\Lambda} y_k$, and so the sequence $(L_k)_{k \in \nats}$ is r-linearly convergent for $k$ sufficiently large w.o.p.~Further, the rate of convergence is at most $\sqrt{\Lambda}$ given in~\eqref{eq:lambdakbound}.

\section{Conclusions}
\label{sec:conclusions}
We have proposed the FMS$_p$ algorithm for fast, robust recovery of a low-dimensional subspace in the presence of outliers. The algorithm aims to solve a non-convex minimization, which has been studied before. Recent successful methods minimize convex relaxations of this problem. The main reason that we aimed to solve the non-convex problem was the ability of obtaining a truly fast algorithm for RSR. Indeed, the complexity of the FMS$_p$ algorithm is of order $O(T N D d)$, where the number of required iterations $T$ is empirically small. We also prove globally bounded and locally $r$-linear convergence for a special model of data. A side product of minimizing the non-convex problem is that its minimizer seems to be more robust to outliers than the minimizers of convex relaxations of the problem. Furthermore, it can even include non-convex energies when $p<1$ (on top of non-convex domain), which may yield faster convergence, although the theoretical results point to problems with $p<1$.
%Alternatively, we may replace e.g., $\dist^p(\bx_i,L)$ with $\log\dist(\bx_i,L)$ in \eqref{costf2} to obtain an analog of the Tyler M estimator energy~\cite{Teng_log_rpca}; though we have found that our current implementation is already competitive with Tyler M-estimator.
Empirically we see faster convergence for $p<1$ in Figure~\ref{fig:linconvergence_verification}, which is similar to the result of~\citet{Daubechies_iterativelyreweighted}, although it is not obvious from our theory why this is the case.

The non-convexity of the minimization makes it hard to theoretically guarantee the success of FMS$_p$. We were able to verify the convergence of the iterates to a stationary point. Further, in special cases when data is sampled from the mixture measure~\eqref{mixtmeas1}, the FMS$_p$ algorithm converges to the global minimum w.o.p.~There are a few interesting directions in which the theory of FMS$_p$ can be extended. First, we plan to extend the robustness to noise result of GMS and Reaper~\cite{Coudron_Lerman2012} to our setting. Also, we empirically find that FMS$_p$ converges to the correct solution in all cases of the most significant subspace model~\eqref{mixtmeas1} when $N$ is sufficiently large. We hope to extend our theorems to encompass all cases when $K>1$ and $d>1$. Finally, Figure~\ref{fig:linconvergence_verification} shows global linear convergence under~\eqref{mixtmeas1}, which should be theoretically justified.

It was interesting to notice that in both synthetic and real data that reflect our model, we never had problems with global convergence of the iterates $L_k$. In view of the current theory and strong experimental experience that we had, the FMS$_p$ algorithm seems very promising due to its potential for robustly reducing dimension in clustering and classification tasks. The denoising effect of dimensionality reduction by FMS$_p$ seems to have the potential to be better than PCA, as is demonstrated in Figure~\ref{UCIDSA}. While PCA is a standard technique for dimensionality reduction, FMS$_p$ does not add much complexity and thus can easily be tested anywhere PCA is used. We will make our implementation (including the randomized PCA implementation) available.

\section{Acknowledgments}

This work was supported by NSF awards DMS-09-56072 and DMS-14-18386 and the Feinberg Foundation Visiting Faculty Program Fellowship of the Weizmann Institute of Science. We thank Joshua Vogelstein for his recommendation of the astronomy data; Ching-Wa Yip for creating the astronomy dataset used in this paper; Tam\'{a}s Budav\'{a}ri and David Lawlor for their help with processing and interpreting the astronomy data; Teng Zhang for useful comments on earlier versions of this manuscript and helpful discussions; and Nati Srebro for encouraging us to write up and submit our results.

\bibliography{refs_1_20_15}

\appendix
\renewcommand*{\thesection}{\Alph{section}}

\end{document}